# A General Framework for the Representation of Function and Affordance:
# A Cognitive, Causal, and Grounded Approach, and a Step Toward AGI[1]

Seng-Beng Ho[2]

Institute of High Performance Computing, A*STAR, Singapore

**Abstract**

The function and affordance of natural or artificial objects and various physical constructs feature prominently in the functioning of an intelligent system or an intelligent autonomous system (IAS). In navigating an environment, an IAS needs to understand the function and affordance of roads, stairs, bridges, and various physical constructs that may assist with or obstruct its movement to achieve its goal(s). When encountering objects, understanding their function and affordance so that they may be manipulated or made use of to support the IAS in performing certain required tasks is a key aspect of the intelligent functioning of the IAS. In AI research, so far, the attention paid to the characterization and representation of function and affordance has been sporadic and sparse, and it has not received the same attention as, say, object categorization or natural language processing, even though this aspect features prominently in an intelligent system's functioning, and is more fundamental and important in many ways. In the sporadic and sparse, though commendable efforts so far devoted to the characterization and understanding of function and affordance, there has also been no general framework that could unify all the different use domains and situations related to functionality, and that could provide an intelligent system or an IAS the necessary computational, representational, processing, and reasoning constructs to effect recognition, generation, and application of functional concepts. This paper develops just such a general framework, with an approach that emphasizes the fact that the representations involved must be explicitly cognitive and conceptual, and they must also contain causal characterizations of the events and processes involved, as well as employ conceptual constructs that are grounded in the referents to which they refer, in order to achieve maximal generality for the framework to be applicable to a wide range of domains and situations. The basic general framework is described, along with a set of basic guiding principles with regards to representation of functionality. To properly and adequately characterize and represent functionality, a descriptive representation language is needed. This language is defined and developed, and many examples of its use are described. The general framework is developed based on an extension of the general language meaning representational framework called conceptual dependency. To support the general characterization and representation of functionality, the basic conceptual dependency framework is enhanced with representational devices called structure anchor and conceptual dependency elaboration, together with the definition of a set of ground level concepts. These novel representational constructs are defined, developed, and described. A general framework dealing with functionality would represent a major step toward achieving Artificial General Intelligence.

## 1. Introduction

Given various objects and physical constructs, whether they be natural or artificial, a large part of the concern currently in AI is the categories they belong to. Both in research and industry, massive amount of effort is devoted to categorization of objects and physical constructs, often accomplished using the various techniques of machine learning, specifically deep learning [1]. However, the function and affordance of an object lies at the core of its interaction with other objects, the environment, and human beings, and should be a major concern of AI. And this, this paper submits, has not been adequately addressed.

[1] This research did not receive any specific grant from funding agencies in the public, commercial, or not-for profit sectors. AGI = Artificial General Intelligence. Ironically, when the term AI (Artificial Intelligence) was first coined in 1956 by John McCarthy, it was meant to be *general* intelligence to start with. Due to a quirk of historical development, a more "specialized" kind of AI was developed to commercial success in recent years, and people have been identifying this "ASI" (Artificial Specialized Intelligence) with "AI", and "AGI" had to be concocted and now becomes the new goal to strive for.
[2] Email: hosb@ihpc.a-star.edu.sg, hosengbeng@gmail.com

The issue of functionality has been sporadically studied since the 1970's, but it has received scant attention compared to say, computer vision or natural language processing. There had been some earlier works in the 1970s – 1990s on functionality, represented by those of Freeman and Newell [2], Gibson 1979 [3], Ho [4], DiManzo [5], Stark [6], etc. There is a number of ways functionality can be addressed. From the point of view of trying to recognize if a certain object or physical construct can serve a certain function(s), one can rely on visual features. For example, if one sees an object that resembles a chair, one can infer that it can serve the chair function, and often the purpose of communicating to one other that certain visual shapes or constructs are seen is to convey the potential functions to solve one's need(s) at the moment. So, if one conveys to another that "I see a chair here," one is conveying that perhaps the recipient of the message can use this object to sit and rest. We term this visual feature approach the fast and shallow approach. However, this means of identification does not capture the fact that there could be a novel shape or construct that can serve the same function. For example, a Balans chair[3] is perhaps 90% dissimilar to that of a normal chair, and yet it is called a chair and can serve the chair function. Moreover, having identified that a certain object or construct belongs to a certain category, the next step is to know how to *use* it (i.e., in the case of chair, how to *sit* in it), so it is necessary to understand how it *functions*. There are also other objects and constructs that do not look like the normal chair or the Balans chair, and are not even remotely identifiable as "chairs," but can nevertheless be used to sit on (e.g., tables, stones, boxes), and being able to find something that can serve a certain function lies at the core of the intelligent functioning of an intelligent system or an intelligent autonomous system (IAS). (We define an autonomous system as one that has its own motivations or top level goals built in [7]–[10].) Hence, we term the ability to reason about the functionality of objects and constructs, not just categorization based on the visual features seen, the deep approach. Ho's work of 1987 [4] attempted just such an approach, in which objects that are visually as disparate as that of a normal chair and a Balans chair are reasoned and identified to serve the same chair function based on a certain functional definition. Research related to functionality both before ([2], [3]) and after ([5], [6], [11]–[13], etc.) Ho's effort had not quite addressed the representation of function at this fundamental level. For example, in Stark [6], an effort carried out after that of Ho [4], the so called "functional" definitions of ordinary chairs on the one hand, and that of the Balans chair on the other were stated and represented separately and disjunctively though the use of visual features, thus not capturing the "true" core functional definition of chairs. However, Ho's effort itself [4], with regard to the representation of functionality, also has other shortcomings which we shall discuss in Section 2, and which this paper targets to address along with other related issues.

Consider the intelligent tasks facing an IAS in its moment-to-moment operations. First, it has to navigate around in the environment, and the function and affordance of *physical constructs* such as roads, walls, bridges, stairs, elevators, are certainly its top concern. As mentioned above, one can rely on visual features to identify the correct objects or constructs to serve an IAS's, say, navigational purposes (so as to satisfy certain internal needs), but also as noted, even after knowing that a certain object or construct is of a certain category of thing, it has to know how to *use* the thing involved, and also often creatively as the things encountered may not always be exactly visually or physically identical. Second, when the IAS encounters particular *objects*, it also has to know not only what they are called, but also how to *manipulate* them, again to serve its purposes and needs. Therefore, understanding the function and affordance of objects and physical constructs lie at the core of the intelligent functioning of an IAS, or for that matter, any AI system.

Recently, a major AI research group (Center for Vision, Cognition, Learning, and Autonomy, UCLA – http://vcla.ucla.edu) has called for a major paradigm shift in AI, which is to shift from an emphasis on a "classification based" orientation to a "task based" orientation in addressing AI related issues [14]. They call for addressing the issues of Functionality, Physics, Intent, Causality, and Utility (FPICU), and they reckon that these represent the "dark matter and dark energy" issues of AI, while the categorization issues represent the "visible matter/energy" part of AI. As is well known, the ratio of dark matter/energy to visible matter/energy is 96% to 4% [15]. Hence, this underscores the utmost importance of the issues of functionality. Issues on Functionality is also closely connected to issues on Physics, Causality, Intent, and Utility, thus the various components of FPICU are to be considered together.

In this paper, we aim to address the following issues:

   a)  What are the basic concepts involved in characterizing functionality?
   b)  How to capture functionality in a general way with certain representational constructs?
   c)  Can these constructs be used to determine if certain object can support certain functionalities?
   d)  Can these constructs be used to demonstrate that the *understanding* of the functionality involved is achieved?

---

[3] The construction of a Balans chair is readily available on the Internet, or see Fig. 34.

e)  Are these constructs usable for *explaining* the functionality involved?[4]

f)  What does a general framework for functional representation look like?

What we are pursuing here is a computational level theory *a la* Marr [16]. As we will demonstrate in this paper, function-related representations can be implemented in a symbolic cum analogical form. However, we do not preclude other possibilities such as neural network/connectionist kind of representations or hybrid neural-symbolic implementations. As the characterization and representation of function and affordance is an important part of the operations of an intelligent system or an IAS, addressing it adequately and in a general manner is an important step toward Artificial General Intelligence.

This paper is structured as follows. As there has been no general framework defining what should constitute the basic ingredients for a satisfactory treatment of functionality, Section 2 first establishes the criteria for judging what constitute a satisfactory way of addressing the issues of functionality, and compares and contrasts the various efforts that have been carried out by others hitherto to what we plan to achieve in this paper. Hence, in Section 2, the motivation and background for this paper is described with respect to such a set of criteria. Following the establishment of the criteria for a general framework for functionality, Section 3 then drills down to the basic approach and the necessary conceptual and representational ingredients for characterizing functionality. This set of ingredients is meant to be the foundational material for a general framework for the representation of function and affordance. As will be highlighted in more detail, the representation of function and affordance necessitates a descriptive representation language. This descriptive representation language must have the generality to represent a wide and general variety of functionality in various conceptual domains. Section 4 thus describes such a descriptive representation language, building upon a powerful and general language representational framework for natural language understanding called conceptual dependency theory.

Even though the focus of this paper is not on learning, it is of course understood that a representational scheme that is not amenable to learning is not scalable and practical. Section 5 therefore discusses the possibility of learning for the descriptive representation language described in Section 4 and that is to be used throughout this paper, just to make sure that the scalability issue is addressable in future work. Armed with a powerful and general framework, Section 6 then delves into many illustrations of how several important functional concepts can be represented within this framework, and the representations involved are also checked against the criteria of satisfaction established in Section 2. Needless to say, the basic ingredients established in Section 3 for the general representation of functionality are all put to good use in the representation of these functional concepts in this section. In Section 7, further important concepts concerning the representation of function and affordance, namely structure causality, mental experiments, and functional segmentation, are described and illustrated with further examples that demonstrate the use of these conceptual constructs. Finally, Section 9 concludes the paper with a discussion of future work. The list of future work suggested is quite lengthy, and we believe that the principles and methodologies described herein represent just the beginning of the establishment and elaboration of a new and general paradigm for the understanding and representation of functionality.

## 2. Motivation and Background

An IAS, whether it is a biological system (a typical animal including human) or an artificial system, when exploring and interacting with the world, would come across many natural objects and artifacts, as well as physical constructs such as roads, ramps, stairs, bridges, walls in the environment. One kind of information that the IAS tries to make sense of is the category of an object seen and encountered. Another aspect it tries to make sense of is the function and affordance of the object. Understanding the function and affordance of the object or physical construct allows it to satisfy its internal needs and fulfil the requirements of its moment by moment motivations that drive its interaction with the world [7]–[9]. Also, understanding the function and affordance of the object or physical construct is more fundamental than understanding its category, for the purpose of understanding the category is ultimately to see if it can serve certain function and affordance, based on the IAS's knowledge of the function and affordance of other objects and artifacts in the same category. And for creatures or systems that can communicate with natural language, communicating the category of an object seen to another entity is a shorthand way of ultimately communicating the function and affordance involved (e.g., in another entity's absence at a certain location, communicating something like "I see a glass" to it would lead to the other entity's understanding that an artifact is seen that could be used to contain liquids to perhaps satisfy its thirst need).

Also, attempting to understand the functions of *objects* per se is a narrow way of looking at functionality. There is a generality to the idea of functionality: there are the functions of objects and physical constructs, and there is the *functioning* of systems and environments built upon these, and these together represent most, if not all, aspects of reality[5]. This includes the characterization of the function and functioning of the "mental processes" in the IAS themselves, whether they be natural or artificial. The ability of an IAS to "look at" and characterize its own mental processes constitutes the functioning of self-awareness or consciousness. Thus, activities in general that characterize the functioning and operations of reality, whether in the form of specific functioning of natural or artificial systems (e.g., geological processes or operations of human constructs such as restaurants or jet engines) or narratives ("stories") about physical processes, human social events and processes, or mental processes, all involve the characterization of functionality.

---

[4] This explainability requirement is fashionable in current AI research, arising partly from a quirk of machine learning, but not so much in the early days of AI research as such a requirement had always been understood to be essential and should emerge naturally from the underlying representational framework involved.

Despite the importance of functionality, there has been scant attention paid to it in AI research in general. In some cases where functionality is addressed, it is also not addressed in a deep way (e.g., [6], [11]–[13] ), "deep" in the sense as discussed in the Introduction section. Also, all along the way, there has been researchers and research groups that have attempted to address the issues of functionality, but no general framework or methodologies have been proposed that can cover a wide range of domains and situations in which functionality is the central issue of concern ([2], [4]–[6], [11]–[13], [17], [18], [19], etc.). Unlike classification, where the input to and output from the classification system are relatively simple and well-defined, concepts that are functional in nature are complex constructs in themselves, and they are linked to many aspects of the cognitive and affective operations within an IAS. Hence, probably due to this difficulty, there has been slow progress made in this area.

We believe there are two components that are important for the representation of function. One is the representational language. Unlike classification, which is to ultimately assign a label to an object, function is about describing componential activities in a system and how they operate and interact. Therefore, a descriptive representation language is necessary. In Section 3 we will describe what the basic necessary concepts are for the description of function, and in Section 4, we will propose a specific descriptive representation language to implement these concepts. The other component is an IAS architecture in which to elucidate how the activation, operation, and participation of various components within the architecture serve to represent various functional concepts. As mentioned above, the concepts of function and affordance, as well as concepts that are functional in nature, are linked to many aspects of the cognitive and affective operations within an IAS. This IAS architecture will be gradually developed, starting from Section 6, as we progress along the discussions.

The most important aspect of designing a function representational framework is that with the representations, the IAS can employ them to function adequately in various intelligent and cognitive tasks. Whether what is involved is a natural object, or a human artifact (say, a tool) or construct (say, a road), the functional representation involved must demonstrate its ability in the following aspects:

a) Object/tool/construct function recognition (**R**) – the representation must be able to assist in recognizing that an object/tool/construct can serve certain functions.

b) Cognitive and conceptual understanding (**U**) of the object/tool/construct involved – the representation must provide for the understanding and explanation of the purpose(s) of the object/tool/construct, and how its internal parts operate to support the purported function(s).

c) The representation must prescribe how to use (**U**) and operate the object/tool/construct involved in various situations, including novel situations never before encountered.

d) The representation must facilitate the invention (**I**) of new objects/tools/construct for various functions to satisfy various human needs – i.e., providing the IAS with some "intelligence" so that it does not have to carry out blind and extensive search in the problem-solving processes of "invention."

Therefore, being able to support function recognition, function understanding, use, and invention (**RUUI**) of objects/tools/constructs are the four most important aspects in connection with the functional characterization of objects/tools/constructs. Thus, it is of paramount importance that the representations for functional concepts must be *cognitive (Conceptual)*, *causal*, and *grounded*, so that an AI system is able to represent functionality in such a way that reflects a true understanding of the concepts involved, and so that in turn it may reason appropriately and correctly with these concepts and take appropriate actions as a consequence to satisfy the RUUI criteria. These requirements on the characteristics of the representation are explained below:

a) **Cognitive (Conceptual)** – there must be a means to explicitly define and represent the functional concepts involved, typically employing some kind of representation language (such as predicate logic, or some graphical methods), that allows for explicit reasoning processes to operate on the representations involved.

---

[5] The terms "function" and "functioning" refer to different aspects of an object or system. "Function" refers to the purpose or goal of an object or system. E.g., the function of a kettle is to generate boiled water. "Functioning" refers to how the components or parts of an object or system operate together to achieve the function – i.e., the functioning of a kettle includes the processes of how electricity flows into the kettle, and then into the heating elements to create heat to heat up the water, etc. It includes how the elements are structured to maximize delivery of heat, etc.)

    b) **Causal** – as specifying function involves specifying some mechanisms of object and system interactions and functioning, because of the first requirement of (a) above that there must be cognitive and conceptual representations for characterizing functions, there will naturally be a need for explicit descriptions of the *causalities* involved. These explicit descriptions would need to consist of explicitly stated causal steps.

    c) **Grounded** – when symbols are used to characterize (a) and (b), they may refer to other symbolic structures, but ultimately some of these symbols in the entire symbolic structural network must point to some external real-world referents, or certain indivisible or atomic internal states (such as various basic sensations and emotions) of the intelligent system or IAS involved [10], [20], [21]. These are known as ground referents, and they must be elucidated for a complete characterization of the functional concepts involved. Appendix A provides further discussions on this. (See p.26 of [10], and [21] for a discussion on "symbolic circularity" – a chain of definitions of symbols in terms of other symbols that ends up pointing back to the earlier symbols that does not elucidate anything - that is to be avoided for concepts that are not at the ground level. Also, see [22], [23] for the idea that AI has often treated symbols to mean more than its particular "ASCII code," just because the symbols themselves seem to resonate with "meanings" in the minds of the humans perceiving them, and these are "wishful mnemonics" which we believe should be purged from a fully grounded system.)

In the current paper, we focus on demonstrating the ability of the proposed representational framework, that has the properties of being cognitive, causal, and grounded, in satisfying three of the four items in RUUI – namely function recognition, function understanding, and use of the functional objects involved. We relegate "I" – Invention – to future work as the issues surrounding it are more involved. However, the representational framework we propose can easily support the process of invention with full explainability.

All along, there has been work done on some other issues related to functionality. One notable domain is qualitative physics and qualitative reasoning. To reason about functionality, some representational and processing means are needed to reason qualitatively and commonsensically about physical processes in particular and other processes in general. Many of the earlier as well as more recent representative works in this domain are collected in [24]–[27]. There have also been a number of monographs treating more in depth certain aspects of the related issues [28]–[31]. However, all these works, from the earlier to the more recent ones, have not resulted in any consensus on a general framework for qualitative reasoning and qualitative physics. Each representational and processing framework proposed is meant for a particular problem or a particular domain of concern.

As for function and affordance per se, Gibson's work [3] is usually recognized as the earliest, more thorough, and extensive study, from the psychological point of view, of issues pertaining to function and affordance. The earliest more extensive and detailed computational works in functionality are represented by those of Ho [4], DiManzo [5] and Stark [6], although it has been recognized as early as 1971 by Freeman and Newell [2] that functional reasoning is an important aspect of AI. After a hiatus, beginning in the 2000's, a slew of papers was published that addressed the issues of functionality in one aspect or another, such as [12], [13], [17]–[19], [32]–[38], etc. Despite this surge of interest, as discussed above, it is important to articulate a unified and general framework that incorporates representational and processing mechanisms that can capture a deep understanding of functionality at the cognitive and causal level in the AI system, as well as provide mechanisms to ground the functional concepts involved in real-world constructs as well as certain basic indivisible internal mental states [10], [20]. The various works cited above satisfy some, but not all, of the criteria listed: a general framework for cognitive, causal, and grounded representation and understanding. For example, the works of [12], [13], [17]–[19], [32], [36], [37] do capture important aspects of grounded understanding, and there is also characterization of functionality along the cognitive and causal dimensions, but a general framework has not been articulated. The works of Davis [33]–[35] represent the most sophisticated attempt so far at characterizing the behaviors of solids, liquids, and containers, capturing functionality at the cognitive level (in his case, logic representation is used for this level), and to some extent at the causal level, but the representational scheme does not provide systematic links between cognitive level representations and cross-domain real-world constructs – i.e., a general and systematic treatment of the grounding issues of representations. As for the earlier work of Ho [4], even though the methodology used captures functionality at a conceptual level - using functional definitions to recognize visually disjunctive but functionally similar concepts - the framework does not provide explicit conceptual and causal representations to represent the functional concepts involved (instead, implicit representations in the form of procedural representations were used). Hence, the framework is not general enough to be extendable to cover a wide range of functional concepts. The other earlier work on functionality in AI, i.e., those of Freeman and Newell [2], DiManzo [5], Stark [6], etc., also do not adequately address all the issues of conceptual, causal, and grounded representations.

Schank, Abelson, and a group of associates had developed a deep computational semantic framework for representing meaning for natural language understanding, question answering, and conversation, based on a theory called conceptual dependency (CD) [39]–[56]. Even though there have been various attempts in linguistics to address the issues of semantics [57]–[61] and also various attempts at computational linguistics [62], [63] to elucidate the computational nature of language, as we will see, CD still provides the most adequate framework for representing functionality. CD began with tackling the issues of basic representational constructs for the *meaning* of individual sentences, which often involve hidden causalities and functionalities [39], and continued on to develop the representational constructs for extended narratives for question answering and conversation based on the same basic CD constructs [41]. As mentioned above, at the fundamental level, the representational constructs needed for narratives and functionalities are basically the same, and whatever representational constructs that is suitable for representing narratives are potentially suitable for representing functionalities, especially if a narrative representational framework has the right kinds of ingredients as we shall spell out in detail in the ensuing discussions. Even though it was developed some time ago, and was originally meant for natural language understanding, CD contains the right kind of ingredients for us to build our framework on, in the form of an extended version we shall term CD+ (Sections 3.2.3 and 4). As will be seen, the fundamental constructs of CD provide the ingredients for cognitive or conceptual representations, and our extension, CD+, primarily enhances it with the facility for grounded representations, thus satisfying the triple requirements of possessing cognitive, causal, and grounded characteristics for functional representations.

There has been a relatively large effort in the past many years of AI and computational linguistics research that addresses narrative understanding and representation (see [64] and the various Workshops on Narrative Understanding organized by the Association for Computational Linguistics). The representation of narrative certainly involves some of the constructs we are exploring for the representation of function in our current study. As mentioned, we see narrative as a kind of "description of certain *functioning*, often involving human entities in a social context". One could also "tell a story" about the workings of a jet engine, even though no humans are involved inside the jet engine. Every happening in reality can be thought of as a narrative, or a story.

## 3. Basic Approach and Framework

In this section, we discuss the basic considerations that lie at the foundation of a general framework for representing functional concepts. There is a certain fundamental nature of functional concepts that has to be addressed, and that constrains the form of the representational framework for characterizing functional concepts. And, whatever representational framework is concocted, it must be able to be put to use in the functioning of an intelligent system or IAS. Hence, the RUUI criteria (Section 2) can be used to gauge if a certain representation is adequate or not within the total functioning of an IAS. Before embarking on these, we thought it important to clarify how we view the meanings of function and affordance which are sometimes treated as different. It is toward this we first turn.

### 3.1. Function vs Affordance

Both the terms "function" and "affordance" have been used and discussed quite extensively in the field concerned with issues related to functionality. In this section, we discuss our view on these terms, partly to clarify whether we think there is any distinction in these terms, and if not, why not. It is imperative to clarify our understanding on the use of these terms as they lie at the core of our current study.

There have been some discussions attempting to distinguish the differences between "function" and "affordance" [65]–[67]. However, we feel that in one way of looking at the matter, the differences seem to lie mainly in the lexical roles of these terms. For example, consider a sentence such as "the *function* of the jet engine is to provide thrust to move an airplane through the air" (3.1). The same fact can alternatively be stated as "the jet engine *affords* the movement of an airplane through the air" (3.2). Comparing (3.1) and (3.2), "afford" is a verb that allows the sentence to be stated in a different manner. Otherwise, suppose we use "function" in the verb role, the sentence would have to be "the jet engine *functions to provide thrust* to move an airplane through the air," (3.3) or, more compactly, "the jet engine *functions* to move an airplane through the air" (3.4). These sentences could sound a little unnatural linguistically.

The psychologist Gibson [3] coined a term "affordance" which functions more like the word *function*. Using that, the fact can be stated as "the affordance of the jet engine is to move the airplane through the air", or "the affordance of the jet engine is to provide thrust to move the airplane through the air." These sound natural linguistically.

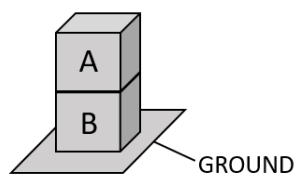

**Fig. 1.** A box supporting another box, at a bigger height compared to the ground level.

Also, depending on the nature of the object or artifact involved, *function* or *affordance* may be a more suitable description. Consider the case of a box supporting another box or other objects as shown in Fig. 1. This is a very common example used in AI, such as in connection with discussions on problem-solving or planning in many AI textbooks (e.g., Russell and Norvig [68]), and which will also be an example we will use extensively in subsequent sections in connection with our central discussion on function and affordance.

In the situation of Fig. 1, a visual or physical description would be something like "A is on B" (often represented in predicate logical form as On(A, B) [68]). This is a visual or physical description of the state of the world consisting of these two objects. From a functional point of view, A is supported by B at a bigger height than that of the ground level. This is done perhaps for the purpose of, say, placing something on A so that that thing can itself reach a further bigger height. Suppose we focus on the functional relationship between A and B, we would say something like, "B *functions* to support A at a bigger height than that of the ground level" (3.5). Alternatively, we could describe the situation as "B *affords* the support of A at a bigger height than that of the ground level" (3.6). In the current situation, if B were to be removed suddenly, A will fall to the ground level. Hence, counterfactually, B is the "cause" of A being supported at that level, or B "enables" the support of A at at that level (i.e., "Had B not been there, A would fall to or stay on the ground"). Between the sentences (3.5) and (3.6), (3.6) seems to be a more natural description, because unlike in the example of the jet engine above, we tend not to think of B as "functioning" in a certain way. "B affords something" seems more natural in this case.

However, on the whole, for the sake of uniformity, we will treat (3.5) and (3.6) as equally adequate descriptions of the situation in Fig. 1. I.e., we will treat "function to support" and "afford the support of" as equally good description of the situation. We will use them interchangeably in many situations. In some situations, using *afford* is more compact, like when we say, "what does a block, B, sitting on the ground afford?" we mean "what are the possible ways a block B sitting on the ground could function to serve certain purposes?" (The answers to this include more than just being used to support A. B could, for example, be pushed somewhere to be used as a blockage to other activities. This will be discussed in subsequent sections.)

On page 36 of Gibson [3], where he first introduced the concept of affordance, he quoted a number of examples. One of the examples is the phrase, "A *path* affords pedestrian locomotion from one place to another…" This can be stated using *function* as follows: "A *path provides the functional support for* pedestrian locomotion from one place to another…" On the other hand, another one of his examples, "A *brink*, the edge of a cliff, … affords injury," cannot quite easily be restated using *function*. One way to restate this sentence is "A *brink*, the edge of a cliff, … *enables* injury to happen" By this it means that if someone were to walk carelessly over the brink, the brink could be the "cause" of her falling and injuring herself, whereas walking carelessly somewhere else where a brink is not present may not lead to that kind of consequence. (This could also be stated as a counterfactual causal situation - "Had there been no brink, she would not have injured herself.") In our subsequent exposition though, causality and enablement feature centrally in our representation of function. Thus, due to perhaps the quirks of linguistic/semantic rules, *function* cannot be easily used to restate this phrase containing *afford*, but the situation involved nevertheless involves functionality of some kind.

There may yet be good reasons to make finer distinctions between function and affordance, as articulated by [65]–[67]. However, as our major concern in this paper is to work out a framework for the representation of function, or affordance for that matter, we are not concerned about this distinction and will instead use these terms in their sentential and conceptual contexts appropriately accordingly.

## 3.2. Basic Concepts for Characterizing Functionality

In this study, we seek a general set of fundamental constructs to represent functions and functioning of systems. These constructs include (i) goal and intention; (ii) cause, state, and state change; and (iii) structure anchor and conceptual elaboration. These are described in the following subsections.

### 3.2.1. Goals and Intention

The description or definition of a functional concept would always involve the specification of a goal(s) or intention(s). For example, in the physical domain, a jet engine "functions to provide thrust to move an airplane through the air". First, in our discussion, we treat goal and intention to be the same, but the distinction of different levels of goal or intention is necessary. In this simple example, the mechanisms in the jet engine first produce thrust, thus thrust can be considered as a goal of the jet engine. The purpose of generating thrust is so that it can cause the airplane involved to move through the air. This is the next level goal.

In the psychological and social sphere, humans may reach out, through social media, to other people to look for friendship and companionship. Thus, a person may take a sequence of steps of looking for a right place in the social media (Step 1), concocting an adequate message (Step 2), and posting the message in the place in the social media (Step 3), with the goal or intention of looking for companionship. One can think of the sequence of thoughts involved in the person in this process, including his psychological reasoning and problem-solving processes of Steps 1, 2, and 3 above, applying psychological rules he knows of other humans, as constituting a psychological "functioning" of some kind. Though there is no specific object such as something equivalent to a "jet engine" in this process, we can still characterize the psychological *function* involved in this process as attempting to achieve the goal of securing a companionship. (Though, unlike the relationship between thrust and movement of an airplane through the air, which is deterministic, the securing of companionship is probabilistic as a consequence of the posting of a supposedly adequate social media message.) The various steps of concocting an adequate companionship-seeking message, posting the message on the social media, and the "final" securing of companionship can be thought of as different levels of goals in the overall functioning of the psychological reasoning and problem-solving process.

In the above paragraph, we thought of the securing of companionship as a "final" goal because according to some psychological theories of motivation, humans have some built-in "ultimate goals" in life, which are the ultimate driving forces of an IAS [7]–[10], [69], [70], and Companionship is one of these "ultimate goals." We term these the Primary Goals of an autonomous system, and this constitutes part of the MOVIVATION CORE (MOTC) of the system as shown in Fig. 2. According to some motivation theories such as that of Maslow [71], there are Physiological needs, Safety needs, Companionship needs, Competence needs, etc. in the MOTC of a human. Other more recent psychological motivation theories have further modified and refine on this earlier theory but the basic ideas are the same [72]. Making finer distinctions, [69] lists 161 fundamental motivations driving the behaviors of a human.

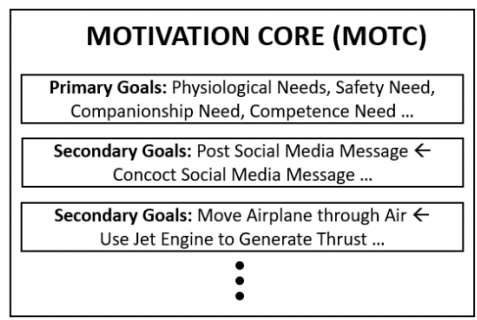

**Fig. 2.** The MOTIVATION CORE (MOTC) of an autonomous system.

In Fig. 2, MOTC contains the Primary Goals and the Secondary Goals [7]–[10], [69], [70], [72] . The Secondary Goals could be of multiple levels, ultimately leading to the satisfaction of the Primary Goals. In Fig. 2 we illustrate the Secondary Goals of the Companionship and jet engine examples above. The movement of the airplane through the air is not a Primary Goal, much like the seeking of money is not a Primary Goal, even though especially seeking of money occupies a large part of human activity in the current world, because humans are not born to have these goals. (The ultimate purpose of traveling through the air in an airplane could be to visit friends and relatives somewhere else, fulfilling the Primary Goal of Companionship.) The acquisition of money is a means, only available after humans have developed their civilizations over millennia to concoct this unique means of economic tool, to achieve other more Primary Goals, such as satisfying the Physiological needs of alleviating hunger by using money to exchange for food, using money to acquire shelter to in turn satisfy Safety and other Primary Goals, etc. In the current study, our concern is the representation of the functional processes leading to various Primary or Secondary Goals. What constitutes the ultimate Primary Goals for many of the Secondary Goals to be discussed, and how these Primary and Secondary Goals vie for priorities of being activated to drive the ultimate behavior of the IAS involved are left to separate and future expositions, even though they certainly constitute one of the most important aspects of an IAS and lie at the core of the characterization of functional concepts.

Since the motivation (or goal) aspect of function is key and fundamental, the basic representational constructs we propose and that will be described in subsequent sections (starting from Section 4) must be able to represent the goals involved in characterizing the corresponding functions as Goal Concepts or Motivation Concepts. We will illustrate how complex goals such as "move an airplane through the air" can be represented within our general representational framework.

### 3.2.2. States and Causes

Imagine a world that is static: there is no change of any properties associated with any object or the environment. There is no functioning of any individual objects or systems. In such a situation, the idea of "function" does not arise. Therefore, another concept that lies at the heart of function is the states of certain properties associated with physical or psychological objects and their attendant changes. For a physical object, a most common property change is the location. For a psychological object such as the human mind or consciousness, a possible change of state could be its emotion such as happiness, sadness, anger, etc.

The concept that is closely connected with state change is the *cause* of the change. There must be available causes that can effect state changes, physical or psychological, otherwise there will be no functioning or function of any kind. These causes could be hidden or visible, and visible causes could be effected by an animate object such as a human effecting certain *actions*, or by other inanimate objects *interacting* with the object whose state is changed, such as wind blowing on a tree, causing it to sway. If there are no visible causes, such as in the situation of a tree or human growing old, we conceptualize them as *temporal* changes without known causes. Of course, in these cases, certain causes could later be found when science such as biology explains certain underlying mechanisms associated with these changes.

The *concept* of cause can be distinguished from its *learning*, *discovery*, or *identification*. Causal learning and discovery is a domain that is being hotly pursued [10], [14], [70], [73]–[91]. Because of noise and other issues, often it is not easy to uncover the cause(s) involved in certain phenomena or changes. As mentioned above, if the causes are invisible or unknown, the changes can only be conceptualized as a temporal change without known cause(s). The framework and methods for the discovery of causes in the scientific domains, such as the medical or social domains, have been studied and established by statisticians for some time [73], [74]. Perceptual causality, which is the direct recovery of cause and effect relations from perceptual data, has been studied by AI researchers, which has relevance for AI or robotic systems' (IAS) learning of such knowledge from the physical environment and applying it to support their interaction with the environment [76]–[78], [90]. As articulated by Yang & Ho [86], whether a certain connection between two events is causal or merely correlational, they are both useful for an intelligent system's functioning. If a causal relation can be established, such as knowing that a certain action can create a certain effect, then the system could in future apply this causal knowledge in a backward reasoning problem-solving process to generate the correct action(s) when a certain effect is desired. If only a temporal relation can be established, then the system could still use it for prediction – say a certain event is likely to follow another event – which will benefit the system's functioning in the environment.

In this paper there will be extensive use of causal relations to characterize function and affordance, but we relegate the learning and discovery of such relations to other researchers' work or our own future work. Our focus in this paper is on the *representational* issues of function and affordance, with causality often in critical central positions in the representations involved.

Often, before a certain cause could create a certain effect on an object, there must be certain preconditions met. These preconditions could be certain states of the object involved, the mere presence of other objects, or the environment. Suppose the action of releasing a held object leads to its falling downward (a change of the object's state – location), the precondition could be that the object is situated in a gravitational field. To set up this precondition, which is a state of the object (i.e., situating it in a gravitational field), perhaps an action such as moving it from somewhere else without a gravitational field to the current location must be effected. Thus, in the functioning of a system comprising a number of objects, there is a web of causes creating certain states, which in turn become the preconditions for other causes to be effected, and so on. In [10], [92], these preconditions are known as *synchronic causal conditions*. The picture is shown in Fig. 3.

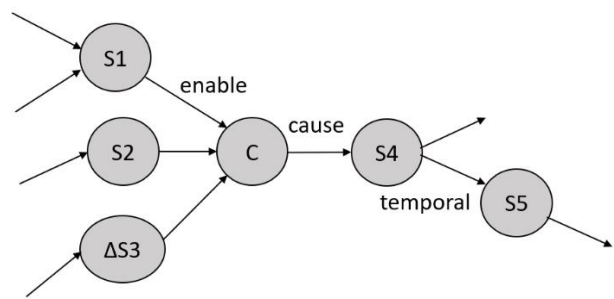

**Fig. 3.** A web of states (S1, S2…), ongoing state changes (ΔS3…) and causes (C) of state changes, including changes over time (temporal) with no identifiable causes – Causal-Temporal-State (CTS) graphs. The preconditions that enable certain cause-effects or the effects themselves could be states or state changes.

Often, the cause/action changes the states, and the final values of the states themselves are the preconditions of another cause. However, there could also be certain ongoing changes of certain states that are the preconditions of the cause. For example, a constant flow of air, which is a contain change of the state of air, is needed to enable an aircraft to stay afloat in the air. Similarly, the effect of a cause could be an ongoing state change. We term the graph in Fig. 3 a Causal-Temporal-State (CTS) graph. Spatial property changes such as location changes are a special case of state changes.

There has been a number of important works illustrating the use of CTS graphs to represent system functioning and narratives. Early representative works include Schank and Abelson's Restaurant Scripts [41] and recent representative works include the CST (Causal Spatio-temporal) graphs of the UCLA group [36], [77], [93]–[96]. These recent works on CST graphs include the illustration of how they could be learned from visual observation, whereas the earlier works of CTS graphs such as Schank and Abelson's were hand coded.

Abelson [40] incorporates *purpose* in his CTS graphs, and he calls graphs such as these "molecules," as shown in Fig. 4. Molecules are formed from atoms (the states and actions) and more complex molecules can be formed from simpler molecules. Incorporation of purpose in CTS graphs gels with what we discussed earlier about goal and intent being an integral part of the characterization of function.

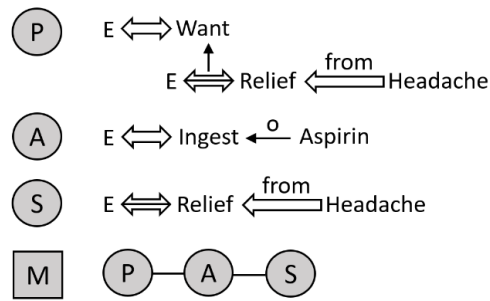

**Fig. 4.** An example of Abelson's belief system representation, in which a Molecule (M) is made up of "atoms" such as Purpose (P), Action (A), and State (S). Redrawn from [40]

In Fig. 4, the representation method is based on the conceptual dependency (CD) framework of Schank [39] which we will cover in more detail later in Section 4. Basically, it begins with a Purpose statement (P) that says that E, an agent, Wants Relief from Headache. E then takes the Action (A), in this case the Ingestion of Aspirin, that then leads to a State (S) in which E is Relieved from Headache. (The double arrow represents a *conceptualization* that links a subject with an action, and the double arrow with a line running down the center of its length represents a state reached.) The symbols used in this representation, E, Want, Relief, Headache, Ingest, and Aspirin, can be further elaborated and grounded to fully flesh out the meaning behind this representation, and the mechanisms to do so will be discussed in subsequent sections. The purpose of the Fig. 4 is mainly to demonstrate the connections between Purpose (goal), Action, and State.

In later sections we will see examples in which a functionality-related process could begin with an agent *wanting* to do something, such as being able to see farther which she cannot do at the ground level, and the *function* of some boxes (such as those depicted in Fig. 1) is such that they could support her in achieving the goal. In general, the P-A-S graph of Fig. 4 would be a complex web of states and actions such as shown in Fig. 3.

### 3.2.3. Structure Anchor and CD+ Elaboration

To adequately represent the full detail of function, we propose representational constructs called "Structure Anchor" (SA) and "CD+ Elaboration" (CD+E), to support the representation of function to the level where deep reasoning and understanding can be obtained. ("CD+" is Conceptual Dependency+, which will be explained shortly in Section 4.)

Fig. 5 illustrates this idea. Consider a simple event/narrative/functioning, "a person kicks a box and it causes the box to move a bit". Consider also a Conceptual Dependency+ (CD+) representation representing this event as shown in Fig. 5. (CD is based on Schank's early work [39]. In the bulk of this study, we will be using CD constructs extensively, and we will be using an enhanced version of CD called CD+. This will be described in detail in Section 4.) The basic CD representation of the "a person kicks a box" event is the double horizontal arrow in the figure that links Person to the action, Kick, and the object, Box. The vertical downward arrow is the "cause" link that links this event to the effect it causes, namely "the box moves a bit," represented as a change of the location (Loc) state of the box from one value to a slightly different value. So far, this part is how the basic CD represents the event "a person kicks a box and it causes the box to move a bit," except here we reverse the arrows' directions for the object link (o) and the cause link (cause) compared with that of the original CD to make the direction of causality and action to object clearer. (In the figure, ACT stands for action, and PP is Picture Producer which means it is something that can conjure up a picture – these are used in the CD framework [39].)

Now, to fully represent the functioning involved in this narrative, we introduce SA that elaborates and anchors the meaning of Person and Box using analogical representation, and CD+E that elaborates on the details of the Kicking action. In the original CD framework, Person and Box are called picture producers (PP) because a "picture" associated with it can be invoked [39], and Kick is an action (ACT), but there is no further elaboration other than the word-symbolic level descriptions themselves, i.e., "Person" and "Box." In the subsequent sections, we will demonstrate the utility of this analogical SA in reasoning (Sections 6.2.4 and 6.3). (The SA could conceivably be implemented, say, in the form of vector representation.) The SA could be an instance, a collection of instances, or a conceptual generalization of instances. If the SA is a generalization, then, other than the particular analogical structure shown in Fig. 5, there could be some statements stating, say, the typical range of values associated with some of its dimensions – e.g., a distribution associated with the height of Person or Box. In the figure, we show the SA of a person as a 3D articulated figure with various parts and joints, which is probably the simplest possible analogical representation that can be useful for reasoning about events that involve a human body (see later Section 6.3 in connection with the functional concept of Chair that requires a human body model to define its functionality). The SA for a human body could be a much more complicated analogical 3D model that includes skin texture, flesh elasticity, detailed bulging bones and muscles on the surfaces of the body and limbs, etc., which could be useful for reasoning with other objects such as clothing and other wearable objects.

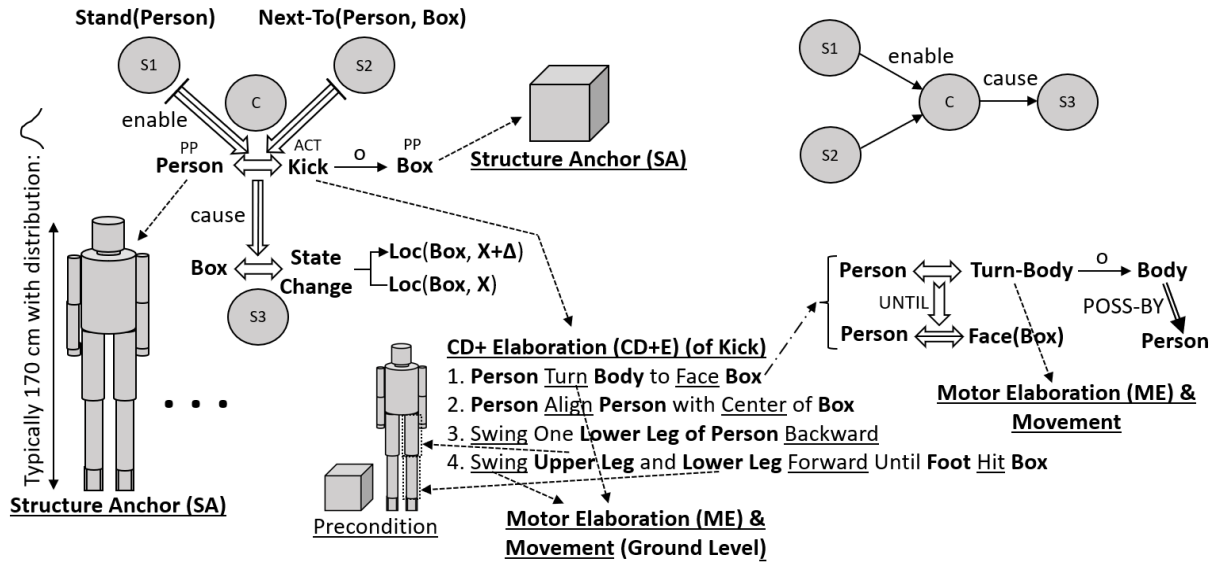

**Fig. 5.** CD+ representational framework: the idea of Structure Anchor (SA) and CD+ Elaboration (CD+E). There is also Motor Elaboration (ME) for the lower level actions. Together with the basic CD framework [39], these constitute the CD+ representational framework. SA could be a collection of instances.

In Fig. 5, in some places in the main body of the representation, we use predicate logic-like representation for some of the conditions and parameters, especially those associated with states. The details of the CD+E, except for one item, are stated in English for the compactness of illustration here. Each of the sentences in the CD+E shown in Fig. 5 can in turn be represented using the CD+ representational constructs to fully flesh out the corresponding causal, temporal, state, and state change descriptions. In the figure, the first CD+E item, "Turn to Face Box," is shown in its CD form. The Turning action is carried out UNTIL a certain condition is met, which in this case is that the State of Person is Face(Box). (The process associated with UNTIL usually involves carrying out a sequence of actions and this detail is omitted here but is shown in full in examples such as that in Fig. 25.) CD+E as well as the CD+ structures are hierarchical recursive structures that reach down to the ground level at some point. (See Appendix A for ground level constructs.) Actions such as Turning and Swinging require further Motor Elaboration (ME) in the form of CD+ representation ("Motor Concepts") together with Motor Movements which consists of ground level motor signals that are not further cognitively elaborated (see discussions associated with Figs. 19, 29, and 30). Motor Concepts and Motor Movements correspond roughly to the distinction between task and motor level in the TAMP (Task and Motor Planning) domain in AI and robotics [97], [98].

CD+E and ME may be reminiscent of *procedural attachments* used in some earlier work of AI [68]. However, unlike procedural attachments, which are usually implemented in implicit procedural representations in which the steps of operations are embedded in opaque procedures, and the connections and similarities among sub-operations are not visible to other processes operating alongside them [98], CD+E and ME are explicit CD+ representations which elucidate the steps of operations explicitly and clearly through grounded concepts, and any similarities across different concepts are clearly identifiable. (Note that as currently conceptualized, there are two parts to ME, Motor Concepts and Motor Movements. Motor Movements are meant to capture some low and atomic level motor operations, which is unique to each motor systems. Hence the transparency of their operations is not of concern even though they may be couched in opaque procedural forms.)

The major function of SA is to ground the corresponding *particular* concepts in word-symbolic forms. In Appendix A, we list a number of *general* ground level concepts that are applicable to a wide range of domains and situations. The basic CD representation, SA, CD+E, ME, together with the ground level concepts listed in Appendix A constitute the CD+ representational framework.

Fig. 5 also illustrates how a CTS graph can be mapped onto the CD+ representation, showing the connection between the states and causal links. For the key causal link to be effected, namely Person kicking Box causing Box to move, two preconditions must be present, namely Person must be in a standing position, Stand(Person), and next to the box, Next-To(Person, Box). (In the real world, it is conceivable that a person may kick a box while sitting, but what we are illustrating here is a particular kind of kicking involving a standing position of the kicker, i.e., "stand-kick.") The grounded meaning of spatial relations such as Next-To, Above, On, etc. are provided in [21], [99]. "Stand," as well as other human body postures are also given in Fig. 35.

## 4. Conceptual Dependency+ (CD+) as Core Representational Scheme for Function

CD was devised by Schank [39] that contains constructs that can be used to represent causal links, states, and state changes, ingredients that we deem necessary for the representation of function. The original idea had been explored in detail in a number of papers published by Schank and his associates [39]–[56]. CD was named as such because concepts are linked, depicting their "dependency" on each other. Though the representational scheme was originally concocted to provide deep semantic constructs for natural language understanding, and the work was carried out some time back and has not been expanded since, it had nevertheless demonstrated how complex state changes and causalities capturing various kinds of narratives and functioning can be represented. The representations of many complex concepts were demonstrated in a number of the group's works, notably a complex narrative of the functioning of a Restaurant in the form of a Script. [41]. Even though there had been other works that dealt with the representation of causality, states, and state changes [68], CD had gone much further and deeper in capturing complex concepts and narratives. We will describe the basic constructs of CD in this section and provide examples of how they are applied to represent some complex concepts.

Even though the bulk of our discussion in this paper is concerned with the more physical aspects of human activities and the environment, in this section we will use the examples more or less directly from Schank's original CD paper [39] which contain more social aspects of human activities. This is not only for the convenience of keeping with the original paper, but also to demonstrate that the representational scheme is general – whether they are social narratives or physical functioning, they all require the same constructs for their characterization as outlined in Section 2. Moreover, even what we normally deem to be "physical" functioning of inanimate objects and artifacts are often tied to the uses that they are put to, and needless to say, these are intricately linked with motivation, intention, and emotion, conceptual ingredients that are shared with the more psychological and social aspects of human activities.

Consider an utterance, "John hit his little dog yesterday." The CD representation of this "concept" or "conceptualization" is shown in Fig. 6(a). POSS-BY is "possessed by," a parameter associated with Dog, or, in general, any object or living thing. (There will be parameters such as this used in subsequent figures, such as STATE in Fig. 26, PART in Fig. 44, etc.)

|  | C-Rule | Example |
|---|---|---|
| 1. | PP ⟺ ACT | Mary ⟺ Hit [ —ᴼ→ John ] |
| 2. | PP ⟺ PA | Mary ⟺ tall |
| 3. | PP ⟺ PP | Mary ⟺ Doctor |
| 4. | PP ↓ PA | Mary ↓ tall |
| 5. | PP ↓ PP | woman ↓LOC N.Y.    dog ↓POSS-BY John |
| 6. | ACT —ᴼ→ PP | [ Mary ⟺ ] Hit —ᴼ→ John |

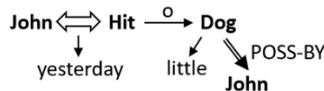

(a)                                    (b)

**Fig. 6.** (a) CD representation of "John hit his little dog yesterday." (b) Basic CD rules. All CD representations and rules are redrawn based on Schank [39].

In Fig. 6(a), as with subsequent figures illustrating CD representations, we reverse the arrow directions of some of the links compared with what was used in Schank [39]. This is so that they would be consistent with the arrow convention we will be using in the extension to CD, CD+, a convention which we believe is more in line with others' conventions, such as the direction of an arrow representing causality should point from the cause to the effect, instead of the other way in the original CD. The table in Fig. 6(b) illustrates the various rules used in Fig. 6(a) as well as rules that will be used in subsequent representations. The three elemental kinds of concepts are *nominal*, *action*, and *modifier*. As mentioned earlier, PP stands for "Picture Producer" because a word that is a nominal is something that can conjure up a picture, such as a person. PA is "Picture Aider" which is a modifier of a nominal such as "tall" modifying "person." ACT is an action like "hit". In CD, in the sentence "Bill likes Mary," "like" is not an ACT because it is not a direct action, but it implies there are other actions that exist that cause Bill to have positive emotions and sentiments toward Mary (see Fig. 11(a) for the representation of "like"). However, we apply CD slightly differently in our paper – we will still use symbols like "like" in the ACT position to capture the surface structure of the sentence involved, but we will set a pointer to point the symbol in the ACT position to a separate structure that elaborates on the causalities involved.

In Fig. 6(b), the first rule is a CD between a PP and an ACT, the second rule is a CD between a PP and a PA and the third rule, a PP and another PP. These are called *conceptualizations*. The difference between Rule 2 and Rule 4 is that while Rule 2 is what Schank calls an "attributive conceptualization" which encodes the dependency between two items on equal footing (e.g., "Mary is tall"), in a phrase such as "the tall woman", "tall" is used to differentiate two women with different heights, and the dependency is shown in Rule 4. Similarly, while Rule 3 is a set membership relationship between two PPs (e.g., "Mary is a doctor"), Rule 5 encodes attributive differentiation like in Rule 4. Rule 6 indicates the dependency between an ACT and the object of the action (objective dependency).

Fig. 7 illustrates the CD representations for the sentences, "The woman took a book" and "I gave the woman a book." At the surface level, it would appear that Fig. 7(a) is sufficient to encode "The woman took a book." ("p" stands for a past event.) However, at a deeper level, there is an implied transfer of possession from someone to the woman. Therefore Fig. 7(b) is a more complete representation of the situation – that the possession was transferred from Someone to Woman. "Someone" is not specified in the original sentence, but it is implied that such a person exists.

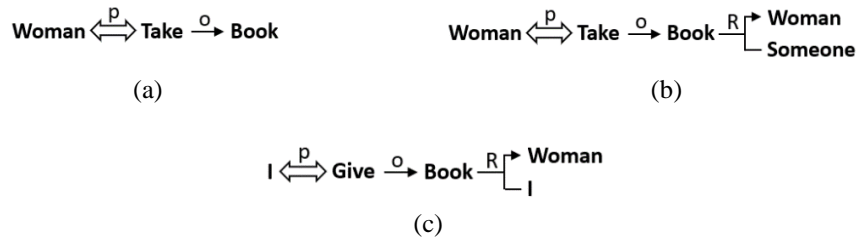

(a)                    (b)

(c)

**Fig. 7.** (a) Representation for "The woman took a book." (b) Implied deeper representation for "The woman took a book." (c) The representation for "I gave the woman a book." [39]

For the sentence "I gave the woman a book" (Fig. 7(c)), the Give CD representation is very similar to that which involves "Take," except that the actor and originator are the same. This inspires the more general representation in Fig. 8 in which Give and Take are replaced by ATRANS (Abstract TRANSfer), which is a transfer of possession, and the difference between Figs. 8(a) and (b) is that in Fig. 8(a) the actor and the recipient are the same, while in Fig. 8(b) the actor and the originator are the same.

Hence this inspires a new CD rule that requires a two-part recipient, unlike Rule 6 of Fig. 6(b) that has a one-part recipient. Rule 6 is shown in the table in Fig. 8(c) along with other new rules. Rule 6 is called an OBJECTIVE case, and Rule 7 is the case we just discussed above, which is called a RECIPIENT case. Rule 8 is an INSTRUMENTAL case in which the instrument (I) used for the action (ACT) is specified. The INSTRUMENT can be just an item or a conceptualization, shown as a double arrow. Rule 9 is a DIRECTIVE case which specifies the physical direction of transfer.

We now consider the example of "John grew the plants with fertilizer" to illustrate the use of the DIRECTIVE case as well as the use of causality, as shown in Fig. 9.

In Fig. 9(a) we can see that the idea of "grew" or "grow" is not a direct action. The sentence implies that John transfers fertilizer physically (PTRANS – Physical TRANSfer) from, say, a bag to the ground around the plant, and that *causes* the physical size of the plant to change. (The downward arrow with a line down its middle and an "i" next to it represents the causal link. "i" means that the causality is intended.)

In our CD+ scheme, we sometimes use a slightly different form of representing State Changes as shown in Fig. 9(b). Also, as mentioned above, in our CD+ scheme, we will still have situations in which we put non-direct actions such as Grow in the position of ACT, but have it CD+ Elaborated using the representations of Figs. 9(a) or (b), as shown.

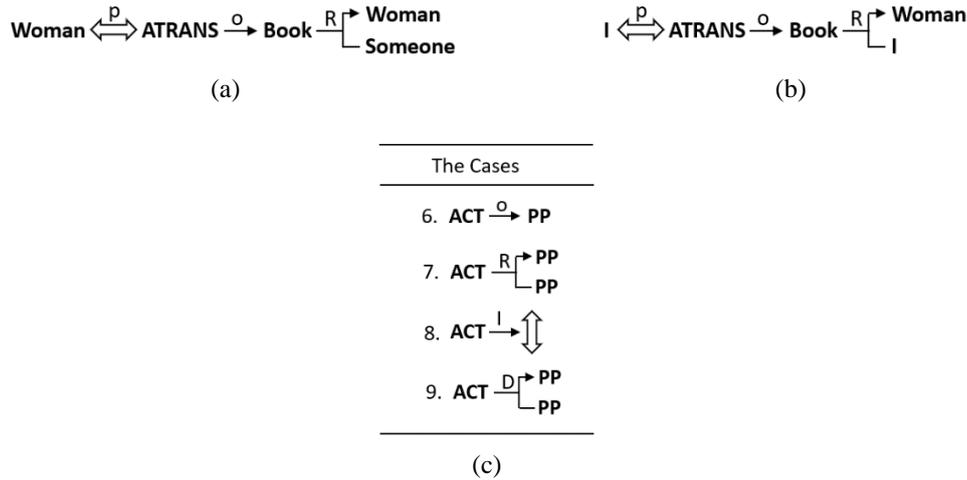

(a)

(b)

(c)

**Fig. 8.** (a) Representation for "The woman took a book" or "someone gave the woman a book" – using ATRANS to unify Take and Give - same actor and recipient. (b) Using ATRANS for "I gave the woman a book" - same actor and originator. (c) CD rules related to cases. [39]

Fig. 9(c) shows two forms of causalities in Schank's scheme: one is a conceptualization causing another conceptualization, and another is a conceptualization causing a State Change. Fig. 9(a) would correspond to the second kind of causality (Rule 10b), but Fig. 9(b) would correspond to the first kind (Rule 10a), even though originally in Schank's conceptualization Rule 10a is meant for situations other than that of State Changes. To us, considering a State Change being also a conceptualization unifies seemingly different kinds of causalities into the same kind.

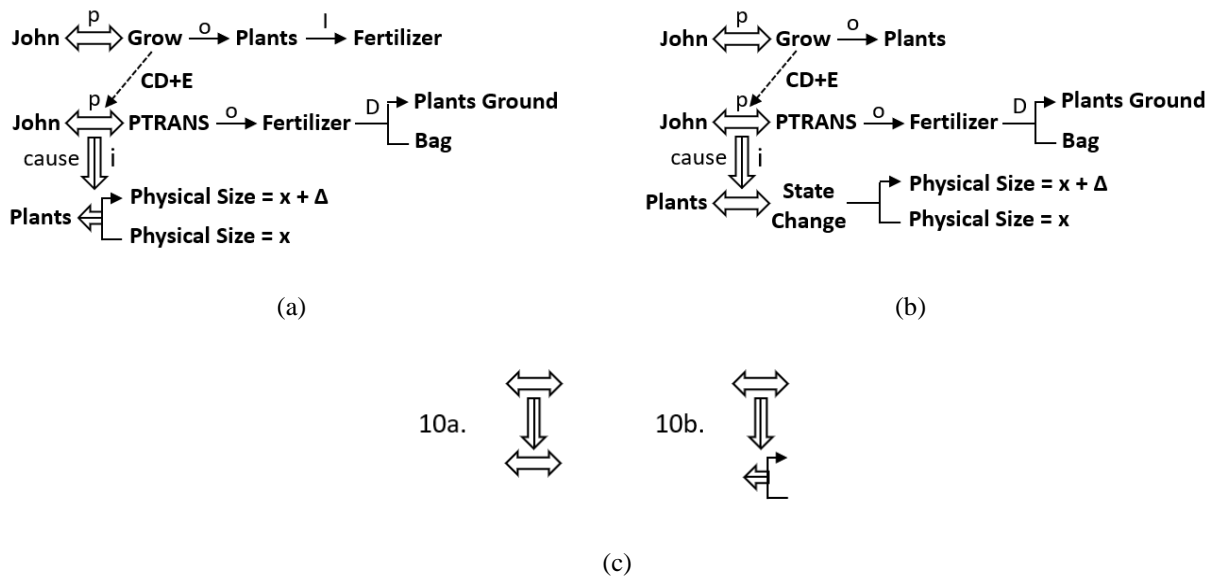

(a)

(b)

(c)

**Fig. 9.** (a) The representation for "John grew the plants with fertilizer." (b) As with (a), but with a different way of showing state change. (c) General rules for representations of causality. [39]

Figs. 9(a) and (b) also show that even though the original sentence "John grew the plant with fertilizer" does not have causalities explicitly stated, it indeed plays a critical role in this conceptualization.

Fig. 10 illustrates the use of the INSTRUMENTAL case using the example "Mary ate the ice cream with a spoon."

In this case, Schank noted in his paper that "If every ACT requires an INSTRUMENT case which itself contains an ACT," then there could be an *endless* sequence of INSTRUMENT cases such as "John ingested the ice cream on a spoon to his mouth, by transferring the spoon to the ice cream, by grasping the spoon, by moving his hand to the spoon, by moving his hand muscles, by thinking about moving his hand muscles," and so on [39]. Schank went on to note that these INSTRUMENT cases exist in the speaker's mind when using natural language to communicate, even though they may not always be mentioned explicitly. (This is what we usually understand as "common sense.") However, in our CD+ representational scheme, we actually promote the idea that firstly there is indeed a hierarchical structure of these conceptualizations that represent further elaborations of the higher-level symbols used to represent the concepts involved, and secondly, it is possible to drill down to a ground level where there is no need to drill down any further (see ground level concepts listed in Appendix A). Thus, it is true that there may be a long sequence of conceptualizations that elaborate the higher-level conceptualizations further and further, but it is not endless – there is a final ground level at which the elaborations terminate. This final level includes what we term SAs (Structure Anchors), an analogical representation of the concept involved, as discussed in Section 3.2.3.

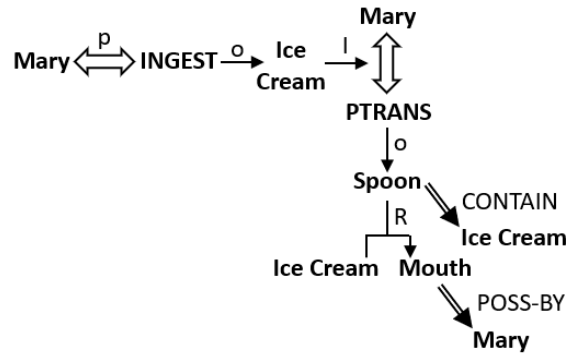

**Fig. 10.** Representation of "Mary ate the ice cream with a spoon." (Note that in this figure as well as in some other figures, the verb in the ACT position – INGEST - as with PTRANS in earlier figures, is in uppercase because in Schank's framework, this is deemed a "primitive" [39], [41]. We will have more to say about this in the concluding portion of this section.)

Fig. 11 shows more examples of implicit causalities. Fig. 11(a) shows the representation for "I like books." "do" is an unspecified action. In this particular case, however, it is likely that "do" is "read," as common sense tells us that is a typical thing we do with books. However, in general, unless the sentence states specifically what the reason for the liking is, "do" is a generic action that causes the "like". "Pleased" is a basic/ground level emotion as listed in Appendix A and it has to do with the satisfaction in the fulfilment of fundamental needs and motivations, in this case, a certain satisfaction from doing something with Books.

Figs. 11(b) and (c) show the representations of "x prevents y" and "x instigates y" respectively. In Fig. 11(b) the first conceptualization causes the second one *not* to happen.

Fig. 11(d) is a representation of "x threatens y" (e.g., Mary threatens John). In this case, there is a chain of multiple causalities involved. x (an agent) threatens y (another agent) means x communicates to y a conceptualization (unspecified) that causes y to believe the if y does something ("cf" stands for "conditional & future"), it will cause x to do something, which in turn will cause y to be hurt. This representation also prescribes problem-solving solutions for both x and y (shown as a PS vertical arrow in Fig. 11). If x intends to warn y about the fact that she could cause y to be hurt and that is her final goal, she could begin with communicating a conceptualization which is a negative consequence for y (a backward chained problem-solving process). If y wants to prevent experiencing hurt, he can break any of the links of causal chain leading to the "hurt" – i.e., he could "not do" the thing that x specifies that if he were to do it, it would lead, after two causal links, to him being hurt. He could also break the last causal link by not allowing what x said she would do to cause hurt to him to happen – e.g., counter-attack x before x could hurt him.

Fig. 11(e) is an elaboration of Fig. 11(d) with the specification of how the hurt is carried out – "I threatened him with a broken nose."

In Fig. 11(a) we had considered the concept of "like." We shall now consider the concept of "want." Schank thought that the difference between *want* and *like* is that, unlike the tenselessness of *like* (i.e., if someone say, "I like ice cream," presumably she has tasted it in the past and it caused her to be Pleased by it and this will be so in the future as well), *want* is more about what Pleases someone "now." So, the basic representation of *want* is shown in Fig.12(a). But again, this not sufficient, because holding a *want* concept is holding something in the IAS's conceptual memory. So, the correct representation should be Fig. 12(b) – the system "conceptualizes" (CONCP) that if it does something (c – conditional) in the future (f) it will be caused to be Pleased. This correspond to some kind of self-awareness on the part of the intelligent system – "I want ice cream" is equivalent to something like "I know I want ice cream."

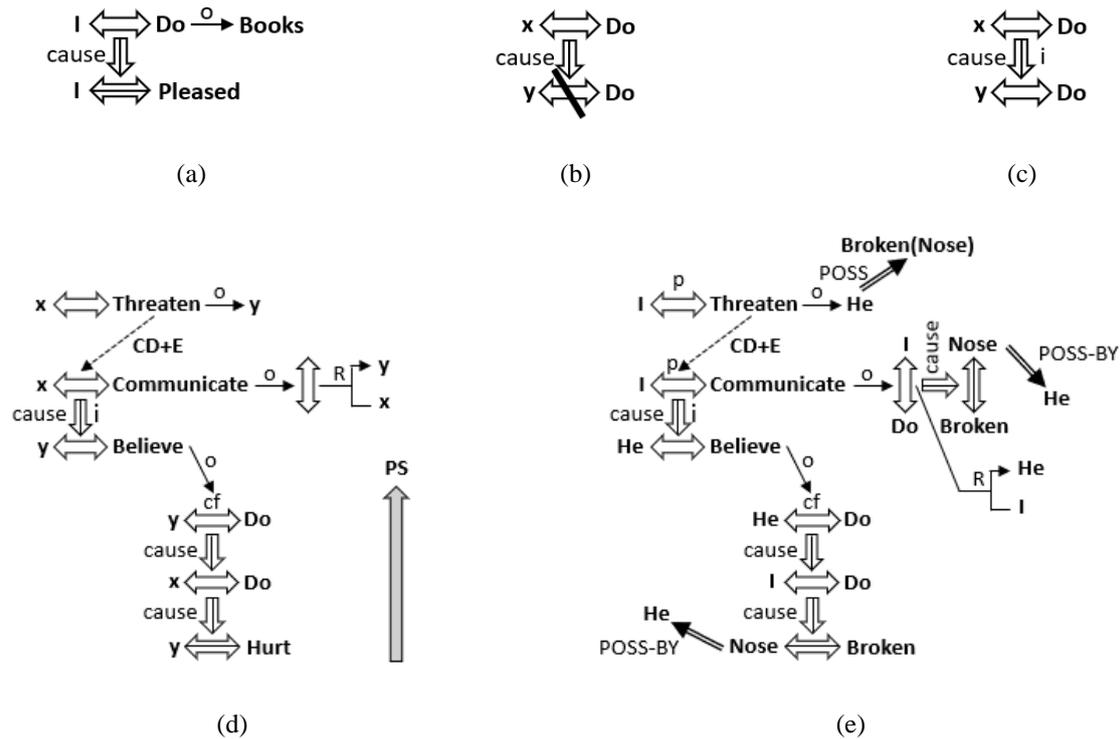

**Fig. 11**. Implicit causalities: (a) I like books. (b) x prevents y. (c) x instigates y. (d) x threatens y. [39] PS = Problem-Solving direction. (e) I threatened him with a broken nose.

This leads to a powerful construct that is found in CD, which is the idea that certain words connotate operations in the "mind" of the cognitive system involved. There are called "verbs of thought" [39]. The corresponding actions move concepts around in different parts of the memory of the IAS. Four kinds of memory are distinguished in CD, and they are short term memory (STM) (the usual "working" memory), immediate memory (IM) (something that holds the immediate context), conscious processor (CP) (where thoughts are currently processed), and long-term memory (LTM) (where earlier learned concepts are stored).

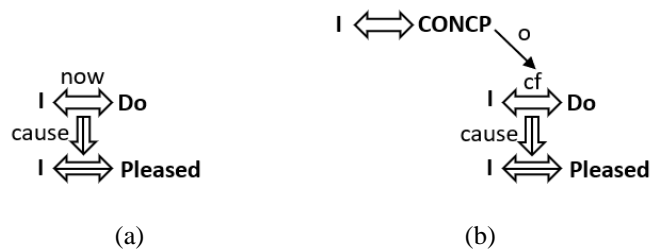

**Fig. 12.** (a) Superficial representation of "I want it." (b) Deeper representation of "I want it." [39] "Pleased" is a basic emotion as listed in Appendix A.

To illustrate the uses of these mental constructs, in Fig. 13(a), the concept of "I saw Mary eating soup" is shown. Basically, the CD representation says "I" conceptualize (CONCP) a conceptualization of Mary INGESTing soup, and the INSTRUMENT of that conceptualization is the mental transfer (MTRANS) of that conceptualization from the eyes to the CP, which in turn is INSTRUMENTalized by the fact that I am LOOKing AT (i.e., the focus of my eyeballs is directed at) Mary and the soup. Other examples of MTRANS – learn, remember, forget – are shown in Fig. 13(b). To Learn is to MTRANS conceptualizations from CP to IM, to Remember is to MTRANS conceptualizations from LTM to CP, and to Forget is the *failure* to MTRANS conceptualizations from LTM to CP. With conceptual structures like those in Fig. 13(b), one can represent many recursive layers of concepts, such as "I remember that I remember that I remember that I went to the store."

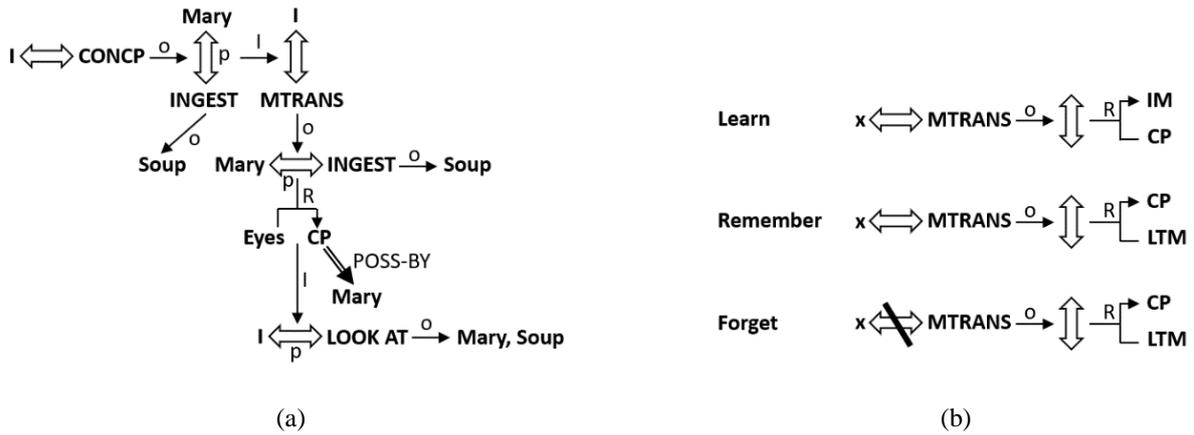

**Fig. 13.** (a) Representation of "I saw Mary eating soup." MTRANS is used. (b) Other MTRANS. [39]

Another important mental operation is MBUILD. An illustrative example is "Fred believes John." Now, what is implicit in this statement is the fact that John must have stated a conceptualization that Fred believes in. In Fig. 14(a), it is shown that this conceptualization has been MTRANSed from John to the CP of Fred, because Fred has presumably heard it. This causes Fred to also CONCePtualize it (the same conceptualizations are indicated with the same gray box around them). Next, there is a difference between Fred merely having heard John uttering a conceptualization, i.e., something like "Fred believes he heard John said x," and Fred really believing what John said is true, i.e., "Fred believes John." So, there is a need for another mental ACT, called MBUILD, that *concludes* that the conceptualization uttered by John is true, and that is transferred from Fred's CP to his IM. This causes the conceptualization uttered by John to be now located in the IM of Fred.

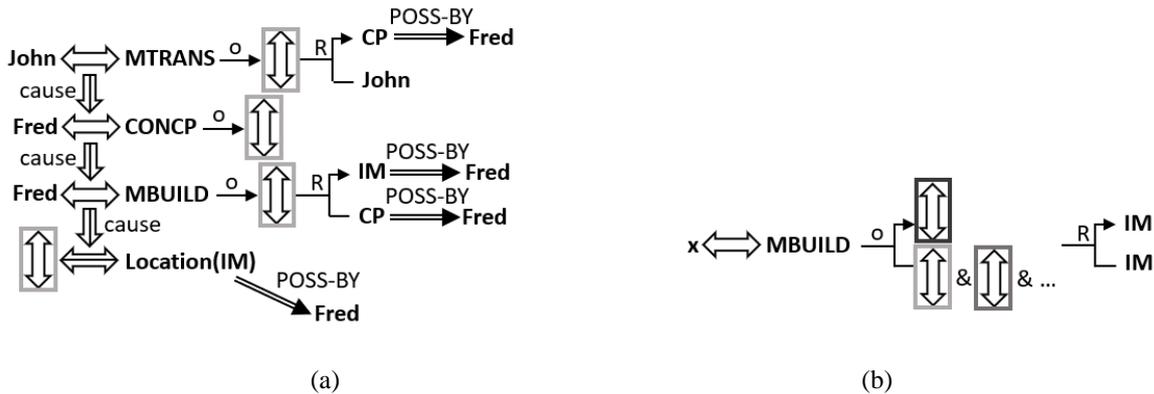

**Fig. 14.** (a) Fred believes John. (b) MBUILD builds conclusions from other information. [39]

In general, MBUILD takes a number of conceptualizations, as shown in Fig. 14(b), and *concludes* another conceptualization out of them (these different conceptualizations are indicated with boxes of differently shaded gray). The action of MBUILD is taking old information from memory (IM) and creating new information and putting it back in memory (IM).

In other related work of CD, it has been shown that it is able to represent very complex concepts very clearly, elucidating the causal and cognitive relationships between the various conceptual components, showing a "true understanding" of the concepts involved. One of the most complex statements and its attendant representation illustrated within the CD framework, as discussed in [52], is "I know you believe you understand what you think I said, but I am not sure you realize that what you heard is not what I meant."

One of the most useful constructs concocted under the CD paradigm, initially meant for natural language question answering and conversation, was the idea of a Script. In their 1977 exposition [41], Schank and Abelson laid out the complex structures needed for deep and complex language understanding, which include Script, Plan, Theme, and their associated representational constructs (Plans and Themes are higher level constructs built on Scripts). We will illustrate the key ideas with the Restaurant Script from their book as shown in Fig. 15.

Schank and Abelson deemed that in order for an AI system to be able to understand and answer or respond to various questions posed or statements mentioned to it around the concept of, say, "restaurant," it is not sufficient to, say, encode a simple dictionary-type (functional) definition of restaurant, like "a business establishment where meals or refreshments may be purchased" (Merriam-Webster definition – www.m-w.com). Instead, a full experiential representation of visiting and eating at a restaurant is needed, such as that shown in Fig. 15. The rationale is simple. If a typical human were to have experienced the entirety of visiting and eating at a restaurant, and have understood the experience, and she communicates about it with another human who has a similar experience, the questions and answers exchanged are bound to involve this knowledge, which would contain a lot more detail than the simple dictionary definition of (the basic function of) a restaurant above.

In Fig. 15 there is a portion at the top in which the Props and Roles involved are stated (much like in a script for a play or a movie). Following that is a specification of the "Entry Conditions" and "Results." These are like the "START State" and "OUTCOME and GOAL" of a typical problem-solving process in AI. Then, the main body of the script follows, which consists of four Scenes: Entering, Ordering, Eating, and Exiting. Within each Scene, a detailed specification of the events that take place within that stage of restaurant experience is given, using CD representations. So, for the Entering Scene, "S PTRANS S into restaurant," "S ATTEND eyes to tables," "S MBUILD where to sit," etc. capture the physical and mental processes involved in the Entering process. Each of these events can be thought of as a node in a graph and they are chained together causally and temporally. There are cases where more than one edge is found leading to a node, because there is more than one type of events that might lead to that event, such as depending on whether the menu is already on the table or it has to be asked for. Also, there could be more than one edge that leads from a node, as often in the real world more than one consequence could lead from one event. Therefore, the graph in Fig. 15 is effectively a CTS or CST AND-OR graph, as discussed in Section 3.2.2.

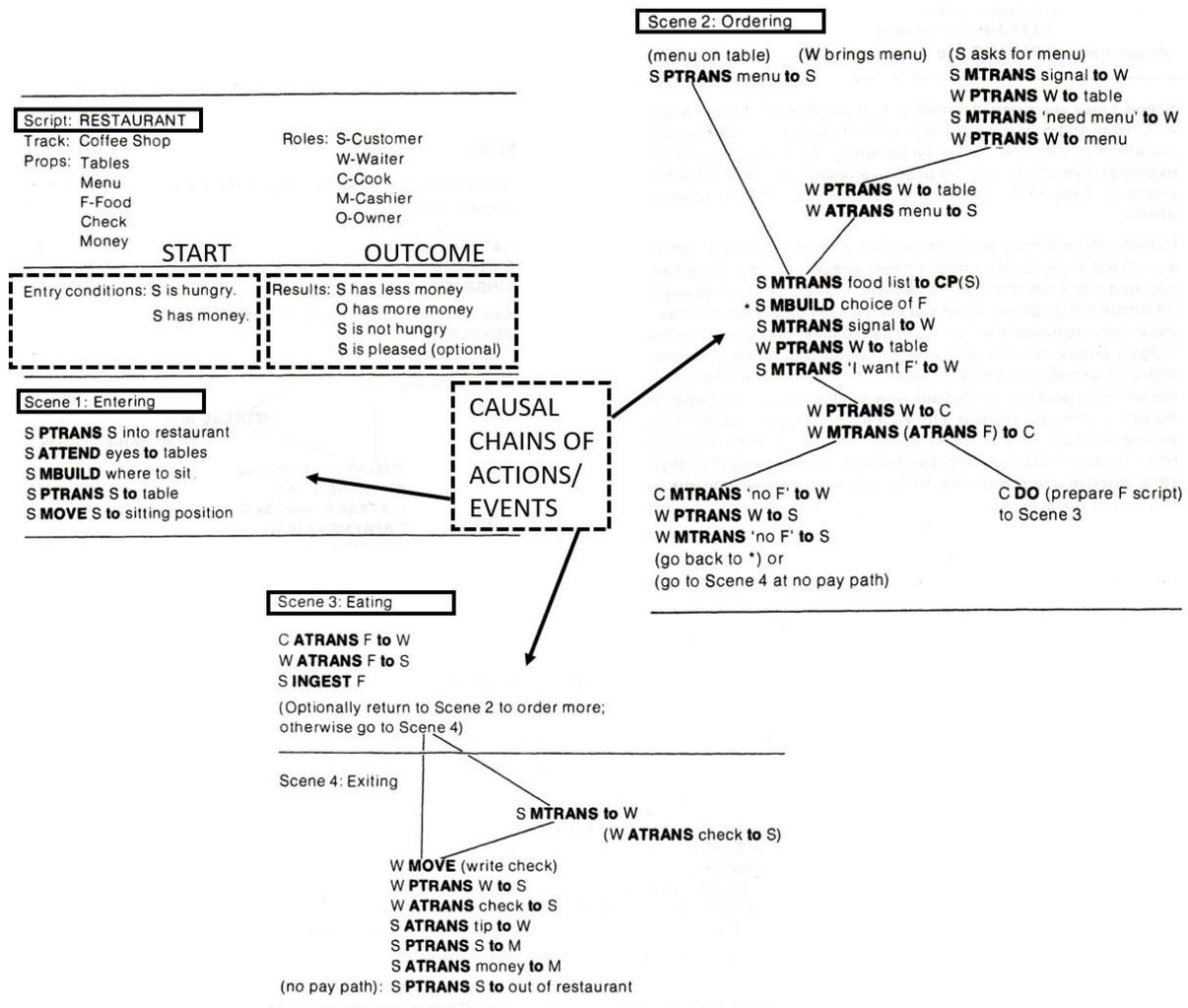

**Fig. 15.** The Restaurant Script. (See text for explanation.) [41] Copyright ©1977 From Scripts, Plans, Goals, and Understanding: An Inquiry into Human Knowledge Structure by R. C. Schank and R. P. Abelson. Reproduced by permission of Taylor and Francis Group, LLC, a division of Informa plc.

The Restaurant Script enables a human-like complex question-answering and conversation process to take place. For example, imagine the following conversational session:

    a) Human: "I went to a restaurant yesterday."
    b) AI: "How was your experience?"
    c) Human: "I did not pay and I left"
    d) AI: "Was it because they ran out of food?"

The AI is able to ask the relevant question because of information gleaned from the script. The question-answering session is hence human-like.

Scripts, in a narrow sense, can be thought of as a narrative of events pertaining to certain contexts. However, the form of knowledge encoded using CD representation as conceived by Schank [39] has a kind of generality that can be applied to any situation involving events and their attendant causal and temporal processes. It has the full power to represent functionality in general.

The basic original CD representation of Schank [39] may yet be functionally equivalent to other kinds of representations such as logic representation. However, the main reason we have decided to build on the original CD representation is that Schank [39] had explored a number of very interesting and useful constructs within the CD framework such as the MTRANS and MBUILD conceptual building blocks that allow certain important "mental" constructs to be represented. These mental constructs correspond to processes taking place *within* the reasoning engine in the intelligent system involved, thus representing and modeling the reasoning processes themselves. Complex knowledge structures such as Scripts, Plans, and Themes have also been proposed and described [41]. It is also easier to extend CD, a graphical construct, to CD+ with the SA and CD+E constructs (Section 3.2.3). Some of the earlier CD work contains further complex conceptual constructs not explored here [39]–[56], and using CD as the core representational construct allow our future work along this direction to dovetail better with those earlier valuable, unexplored, and yet to be but should be extended work.

As it has been demonstrated that the CD representation can be used to represent a large variety of, if not all, natural language expressions [39]–[56], and natural language expressions can be used to describe various functionalities, it follows that CD representations, and its extension, CD+ representations, can be used to represent most, if not all, functional concepts. Therefore, the CD+ representational framework is a general framework for the representation of function and affordance.

Used in a reversed direction, the functional representations that are in CD or CD+ forms can also be easily converted back to natural language descriptions, facilitating the explanations of the functional concepts involved.

The CD representational scheme had also been successfully implemented computationally in a number of question-answering and conversation systems [100]. This implies that with some extensions, notably processes that deal with the SA and CD+E portions, the CD+ representation could also be implemented within a computational AI system to perform the various tasks associated with the recognition, understanding, and use of functional concepts to be discussed in the following sections.

As we pointed out above, that even though CD has its roots in the 1970's, it still has relevance as we will demonstrate that an extension that we propose, CD+, can provide the necessary representational constructs that satisfy the criteria laid out in Section 2 with regards to functional representation. Ironically, many subsequent works on linguistic semantic and computational linguistics that are seemingly more "current" do not provide the necessary representational constructs [57]–[63]. And, it is interesting to note also that the current most popular machine learning method, deep learning and its associated convolutional neural network, also has its roots in the Neocognitron of Fukushima that was concocted in 1980 [101]. It was through the effort of many people who continued to push on to further develop on the original convolutional neural network of Fukushima, as embodied in his 1980 Neocognitron, that we finally arrived at the viable deep learning paradigm of today [1]. We believe a similar developmental trajectory can take place for CD and CD+.

One technical detail to note here again is that while in the original CD framework, the causality arrow points from the effect to the cause, in our proposed CD+ framework, we invert this instead, and have the causality arrow points from the cause to the effect (as has already been depicted in Fig. 5). Similarly, the other arrows' directions are also reversed – e.g., for the ACT to object arrow, we will have it point from ACT to the object involved. We reckon that this is more intuitive and is more in consonant with other various work in graphical representations. At the same time, it does not alter the fundamental ideas behind CD, on which CD+ is based.

Also, in the original CD framework, 11 ACT "primitives" were stated - ATRANS, PTRANS, PROPEL, MOVE, GRASP, INGEST, EXPEL, MTRANS, MBUILD, SPEAK, ATTEND [41]. However, among the various papers on CD [39]–[56], sometimes conceptual symbols other than these primitives are placed in the ACT position in the CD graph, such as "communicate," "believe," etc. [39] In addition, some of these primitives, such as INGEST, are not primitive in the framework of CD+ - they can be further elaborated and grounded with other concepts. Therefore, in CD+, we allow all kinds of conceptual symbols to be placed in the ACT position, but we require that they be elaborated and grounded through SA and CD+E, and ultimately there is a set of grounded concepts that these elaboration and grounding constructs should reach. Appendix A contains a list of grounded concepts that we are using in this paper. There may be more grounded concepts to add to the list in future work.

## 5. The Learning of CD+ Representation

Even though we are not planning to address the issue of the learning of CD+ representation in detail in this paper, it is important to have an inkling of how that may be achieved, as a framework of representation that cannot accommodate the learning of the representations involved would not be useful as a practical method.

We begin by considering the complex representation of a Restaurant Script shown in Fig. 15. If we compare that with the kind of graphs used in a number of more recent papers called Causal-Spatio-Temporal (CST) AND-OR graphs for representing the causal and temporal connections between events as discussed in Section 3.2.2 [36], [76], [77], [93]–[96], the Restaurant Script can also be seen as a kind of CST graph, even though it has not been explicitly named as such, and the research work surrounding it took place in the 1970s while the research surrounding CST graphs originated in the 2010s, a full four decades later. More importantly, it has been shown that CST graphs are learnable directly from video input [36], [77], [93]–[96]. Even though there are some differences between the CD+ or the Restaurant Script kind of graph and the CST graphs, the key parts are similar, notably the causal and temporal links, and hence the learning techniques used in CST graphs are transferrable to the learning of CD+ graphs.

The process of CST graph learning as described in papers such as Si [93] begins with applying deep learning to recognize a number of "leaf" level objects, which are objects that are found in the environment (in their case, a room) such as telephone, mouse, water-cooler, etc. As humans enter the environment and interact with the objects, the activities such as "pick-up telephone," "move mouse," "turn on water-cooler," etc. are tracked. The causal and temporal connections between the events are parsed into an AND-OR graph. Though as of now no one has applied this technique to construct the Restaurant Script, in principle this is possible.

Fig. 16 depicts a simplified restaurant setting, in which there are Chairs, Tables, Waiter, etc. Suppose a little robot (an IAS) with computer vision capability follows a human into the restaurant and observes all the activities taking place in it, in principle it can construct the Restaurant Script of Fig. 15, with the aid of some background knowledge.

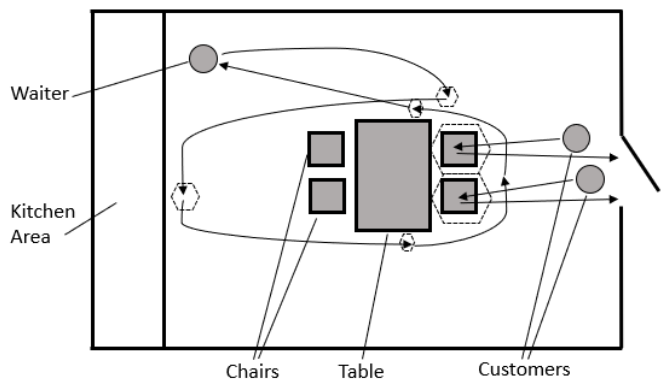

**Fig. 16.** A simplified "restaurant" environment illustrating how the learning of a Restaurant Script could take place.

There are other ways by which something like a Restaurant Script or any CD+ representation can be constructed. The system, instead of moving around in a real environment, could watch videos and movies and learn something. Using humans to provide the knowledge through a crowd sourcing process, such as the Amazon's Mechanical Turk process, could also provide commonsense knowledge such as this. The current efforts in the learning of causality will also contribute to the learning of CST graphs [10], [14], [70], [73], [76]–[92].

Though learning is important, in this paper, as mentioned above, we focus on understanding the representation and representational framework necessary for the encoding of functional knowledge. We believe it is important to understand first what it is that is to be learned that can be useful in intelligent processes, in this case functional concepts and their attendant representations, and then the research on the learning mechanisms to learn these functional concepts and representations could ensue.

## 6. The Representation of Functional Concepts

In this section, we consider primarily two kinds of functional concepts. In CD or CD+ representations, often there are tenses (such as future or past), conditionals (such as "if"), modals (such as "may" and "can") that modify the conceptualizations involved. There are also what we term "control concepts" that act on conceptualizations, such as "enable," "cause", and "until," and these are also among some concepts that participate in the reasoning processes themselves, namely the internal "mental processes" of an IAS. We term these Meta-Functional concepts which are key to defining general functional concepts, and they are functional in nature themselves and can be represented in the CD+ framework. The other kind of functional concepts are those that are related to objects and functioning of configurations and systems. We begin with a discussion on Meta-Functional concepts.

### 6.1. Functional Representation of Meta-Functional Concepts

In general, the description of function or the content of narrative must involve hypothetical conditions, such as "if," or prospective actions, such as "will," "may," and "can." In this section, we will consider the meaning representations of a number of important and related Meta-Functional concepts: *will, may, if, want, can,* and *enable*. These concepts constitute parts of the representations for many other functional concepts. These concepts are mental operations, and are also functional in nature, therefore the same CD+ framework can be used to represent them.

As we will see, the functioning of these concepts involves operations within the memory and reasoning constructs of an IAS. These constitutes "mental processes" as they are internal to the system. We sometimes use "reasoning process" and "mental process" interchangeably, though not all mental processes are reasoning processes – e.g., transferring a conceptual item from one memory location to another, such as that shown in Fig. 13(b), is a mental process but not a reasoning process.

First, consider a simple event, "Person goes (or went) from location L1 to L2." Using CD+, it is represented as shown in Fig. 17.

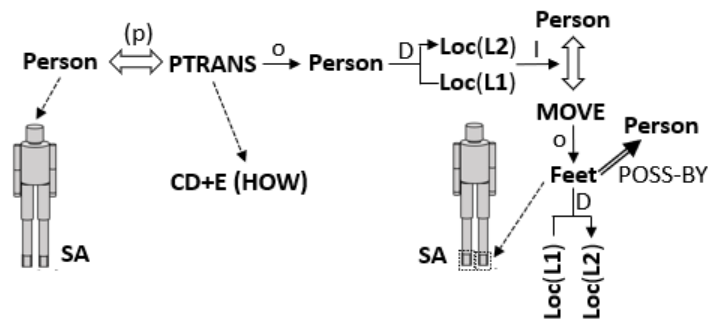

**Fig. 17.** The CD+ representation of "Person goes (went) from location L1 to L2." (Loc = Location. "p" signifies that it is a past event for the representation of the past tense version of the sentence.)

In Fig. 17, we employ a fundamental concept, PTRANS (Physical TRANSfer – see Fig. 9(a)), from CD+, to represent the event. The representation shows that the location parameter of the object of the PTRANS event, Person, changes from L1 to L2, over time. There is a predicate logical equivalent of this representation as follows:

$$\text{PTRANS(A, L1, L2)} \rightarrow \text{Location(A, L1, T1) \& Location(A, L2} \neq \text{L1, T1+}\Delta)$$

which means that if A is PTRANS from L1 to L2, then the Location of A is L1 at time T1 and L2 at time T1+$\Delta$. Physical transfer or PTRANS is a spatio-temporal concept involving the change of the location parameter over time. Therefore, representing it in terms of Location, Change of Location, and Time provides its grounded representation, whether it is in the predicate logical or the CD+ representational form. In the CD+ representation in Fig. 17, time is implicit. (See Appendix A for grounded concepts and see [10], [20] for a discussion of symbolic circularity such as defining a concept like *move* in terms of *go*, and then defining *go* in terms of *move*, which is found in dictionaries such as the Merriam-Webster – www.m-w.com. This ungroundedness is undesirable within the functional concept representational framework because functional representations ultimately prescribe actions that could potentially be taken and these must be taken in the predicate logical or the CD+ representational form.)

In Fig. 17, an INSTRUMENAL (I) case is also used to specify the means by which the PTRANS is effected, which in this case is the use of Person's Feet. Now, in our CD+ representational scheme, we provide for a further specification of the exact means of achieving the requisite PTRANS in a certain context, just like in the case of the Kick event in Fig. 5. This is encoded in the CD+E as shown, pointing from the PTRANS symbol. Suppose Person is now in a room. The CD+E could specify *how* he might choose a path out of the room (say L1 is a location in the room and L2 is a location outside the room), including perhaps first orienting his body toward the door, walking toward the door, opening the door, etc. Suppose the door is locked, and there are no other means to get out of the room (i.e., the problem-solving process of the intelligent system involved – Person, or an IAS attempting to recommend a solution - cannot return a solution), then the original intention of Person's wanting to PTRANS from L1 to L2 is thwarted. Hence, the event "Person goes (went) from L1 to L2" would not have happened. Thus, CD+ provides for representing situations such as these and these will be addressed in the ensuing discussion.

Therefore, whether the event description is in the present tense "Person goes from L1 to L2" or past tense "Person went from L1 to L2," there is an implication that there exists or existed a means for Person to get from L1 to L2. As such, a CD+E of the concept symbol PTRANS is specified for both of these events, except that a "p" would be placed above the conceptualization link for the case of the past tense to signify that it is a past event.

Now, in the original CD scheme, there is no further elaboration on the difference between the present tense and the past tense, other than to place a "p" on the conceptualization link if the event is in the past. Similarly, an "f" is placed there in the event that it is future tense, like "Person *will* go from L1 to L2." As mentioned earlier, the full representation of functionality also involves the participation of the memory and processing modules of an IAS. In Fig. 18, we introduce an important module of an IAS, the EXPERIENTIAL CORE (EXPC), sometimes known as the "episodic memory," [102] in which the experience of an IAS is stored. EXPC is part of the LTM (long-term memory) [102]. Ideally, every bit of the experience (in the modalities of visual, auditory, somatosensory, etc.) of the IAS should be stored, but due to memory capacity limitation, typically only a subset of this is stored, as storing everything would be formidably expensive and impractical. Within our framework, we also incorporate *prospective* experiential memory, PROSP(EXPC) as part of EXPC - what *will*, *may*, or *can* happen in the *future* of the IAS's experience. This is contrasted with the *retrospective* portion of EXPC - RETRO(EXPC). Within our representational framework, the contents in EXPC are represented in the CD+ forms (Fig. 5) as well as the CD+ forms of Scripts, Plans, and Themes (Fig. 15) for extended and complex experiential narratives [41].

In Fig. 18, we illustrate the grounding of the "p" and "f" labels for the conceptualization involved – they point to RETRO(EXPC) and PROSP(EXPC) respectively. For the present tense version of the sentence, it points to the PRESENT(EXPC) location in EXPC.

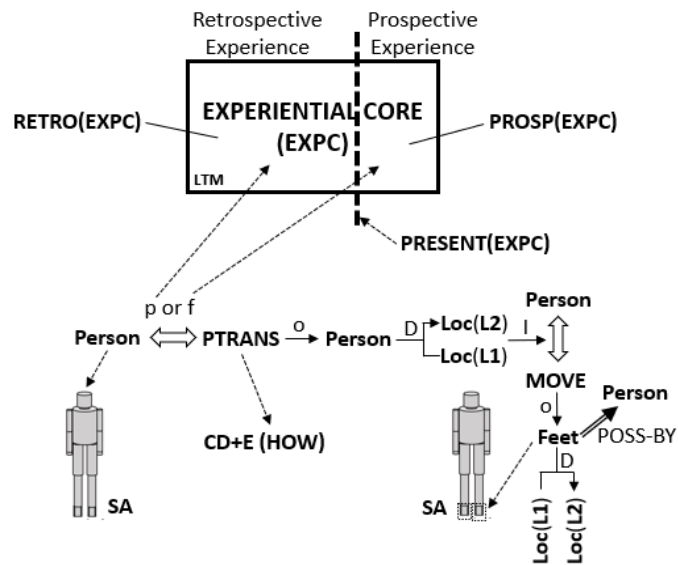

**Fig. 18.** The EXPERIENTIAL CORE (EXPC) of an IAS and the grounding of the "p" and "f" labels of a conceptualization in the CD+ form. Past ("p") event: "Person *went* from L1 to L2." Future ("f") event: "Person *will* go from L1 to L2." The concept of *may* would be a future event (an "f") but with a certain probability associated with it. The contents in EXPC are represented in the CD+ forms (Fig. 5) as well as the CD+ forms of Scripts, Plans, and Themes (Fig. 15) [41].

An interesting question can be raised here. Why do we conceptualize "p" and "f" to be pointing to somewhere in EXPC, which is somewhere internal to the IAS, and not somewhere "out in the real-world"? Say, somewhere along a "time-line" that exists "out there"? First, let us consider the "f" label. As it is more or less the consensus of most people that the future has yet to exist, "f" cannot be pointing to somewhere or sometime in the world outside the IAS. The "future" event can only be a conceptualization that originates somewhere in the IAS, and finally positioned in the PROSP(EXPC) portion of EXPC. Second, let us consider the "p" label. It refers to "sometime in the past." Again, as far as we can tell from the properties of the physical universe as we know it, at any given time, the "past" no longer really exists. For the convenience of cognition and conceptual processing, we often spatialize time by drawing a line with past, present, and future all laid out in space along the line, but this is just a representation and metaphor. Therefore, it is more reasonable to have "p" point to somewhere in the RETRO(EXPC) portion of EXPC. Now, lastly, we come to the "present." To some extent, we can conceptualize the present as existing somewhere "out there." But what really is the "present"? How "wide" is it temporally? Or is what we usually identify as the "present" a mere infinitesimally small point in time? If so, the present "out there" is very fleeting. As soon as that present can be sensed or conceptualized, it becomes the past. Therefore, it would be better to point the present to a point in EXPC - PRESENT(EXPC) - that sits at the boundary of the past and the future, as at least it will always be there in the form of a memory content for the IAS to conceptualize about and brood over. Of course, an IAS can choose to conceptualize these time points as something *outside* in the real world and draw a time line *inside* the IAS (say in CONC and EXPC) to represent this "real world out there" and have "p", "f" and the present point to the corresponding points on that line. Alternatively, one can also think of the EXPC itself as embodying this "real world time line," and "p" and "f" are positioned in it. These are alternative conceptualization of the nature of time, and the CD+ framework, as we shall see as we proceed with our discussions, allows multiple representations for the same concept.

For the presentation of the concept "may," as in "Person *may* go from L1 to L2," it could be represented similarly to the concept of "will," but with a probability associated with the conceptualization, indicating that it is an event that "may or may not happen." If it happens, it is something that resides firmly in PROSP(EXPC).

In Fig. 19, we illustrate the representation of the concept of "*if* Person goes from L1 to L2," a hypothetical event. The convention for representing "if" in CD is "c" – "conditional." We use "IF" and "c" interchangeably.

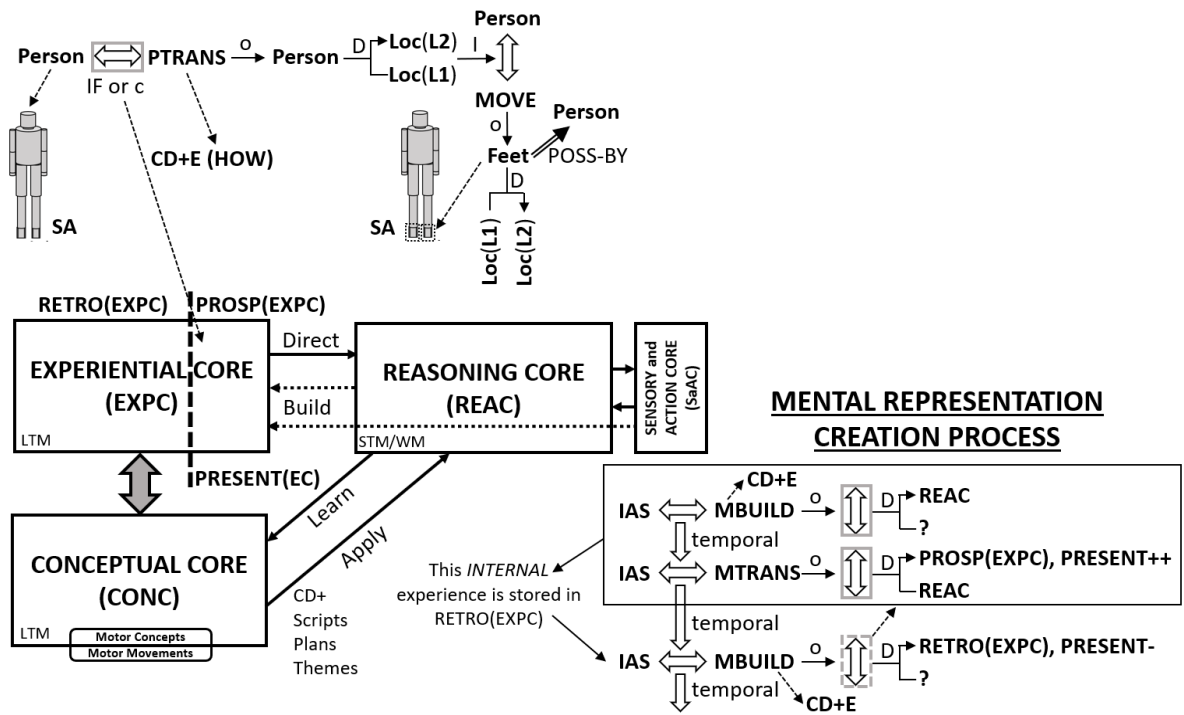

**Fig. 19.** The representation of the hypothetical or conditional, IF ("If Person goes from L1 to L2"). The double arrow with a gray box around it represents the entire "Person goes from X1 to X2" conceptualization. Two more modules of the IAS are added compared to that in Fig. 18 – The REASONING CORE (REAC) and the CONCEPTUAL CORE (CONC). The process of "mental" representation *creation* is itself represented in the CD+ form. The single, thick downward arrow is a "temporal" link – no causality is implied between the linked conceptualizations. The MBUILDs themselves have further CD+E which we will omit in subsequent figures.

Since the description "if Person goes from L1 to L2" involves a hypothetical event, the event has not happened yet in the real world. Therefore, the representation resides in PROSP(EXPC). The difference between this representation and that for "Person will go from L1 to L2," even though both reside in the same PROSP(EXPC) space, is the context or narrative under which they are created. Presumably, the IAS concocts the representation of these conceptualizations inside its memory, and the source of its concoction could be the result of some other thinking or reasoning processes such as a problem-solving (say, backward chaining) process or a (forward) simulation process.

To illustrate the creation of these "mental" representations, Fig. 19 shows, in addition to EXPC, two other main modules in a typical IAS's internal processes (there are other modules and other details that we will discuss later). The REASONING CORE (REAC) is the module in which reasoning takes place, which resides in the short-term or working memory (STM or WM) [102]. This is the module in which problem-solving (say, backward chaining) processes or (forward) simulation processes take place. The reasoning process could derive information from a CONCEPTUAL CORE (CONC) in which a repository of knowledge such as Scripts, Plans, Themes, etc. in CD+ form resides [41] (Plans and Themes are higher level constructs built on Scripts). CONC is part of the LTM and part of CONC consists of Motor Concepts and Motor Movements (both of which are part of ME - Motor Elaboration - refer to the discussion in connection with Fig. 5). Strictly speaking Motor Movements is outside CONC as it is not cognitive/conceptual. The reasoning process could also derive information from the sensory input, the SENSORY AND ACTION CORE (SAAC). (Sensory information includes proprioception input from the motor system.) The EXPERIENTIAL CORE (EXPC) not only records experiences sensed from the external world but also those that take place internally as well, such as the reasoning processes. As shown in Fig. 18, EXPC has a retrospective part and a prospective part – RETRO(EXPC) and PROSP(EXPC) respectively. RETRO(EXPC) contains past external as well as *internal* experiences, and PROSP(EXPC) contains imagined or simulated (external or internal) events in the future. EXPC provides the context for the reasoning processes to take place in REAC.

Using the constructs of CD+, the IAS involved in concocting the hypothetical or future event first MBUILDs (mentally builds - see Section 4, Fig. 13) the PTRANS representation of Fig. 17 (i.e., the representation of "Person goes from X1 to X2") in REAC (perhaps as a result of some other reasoning processes, as mentioned above,). (In order to reduce congestion, the entire representation around PTRANS is represented by a gray box enclosing the double arrow and the gray box with the double arrow inside linked to the first MBUILD is meant to represent the entire PTRANS representation for "Person goes from L1 to L2.") This representation is then MTRANSed (mentally transferred – see Section 4, Fig. 13) to PROSP(EXPC) from REAC, and situated in PROSP(EXPC) in some future time, PRESENT(EXPC)++. The single, thick downward arrow is a "temporal" link, which means an event follows another with no specification of the causalities involved. Immediately after these processes, these internal events (i.e., the MBUILD and MTRANS events – grouped together in a box in the figure) are stored (MBUILDed) in RETRO(EXPC) as events that have been experienced (at time PRESENT-). The is shown as the bottommost MBUILD that takes a conceptualization, which is the dotted gray box with a double arrow inside, which points to the entire box containing the first MBUILD and the MTRANS, and stores that in RETRO(EXPC). Thus, this demonstrates that the CD+ representation can be used to represent the mental processes that create other representations for reasoning, and these mental processes are also functional in nature. Internal mental experiences are stored along with external experiences in EXPC.

Even though "if Person goes from L1 to L2" is a hypothetical event, it still implies that there exists a means to do so, as illustrated in Fig. 19. Now, consider the description "Person *wants* to go from L1 to L2." Fig. 20 shows the CD+ representation of this concept derived from Schank's CD framework discussed in Section 4 (Fig. 12(b)) [39]. In this case, there may or may not exist a means to go from L1 to L2 ("HOW? - IRRELEVANT" – the existence of a means is irrelevant to the conceptualization of "want"). But *want* implies that Person desires to PTRANS from L1 to L2, and will be caused to be in the state of Pleased if PTRANS from L1 to L2 is successful. Sometimes, a situation such as "I want to get there but I can't" exists.

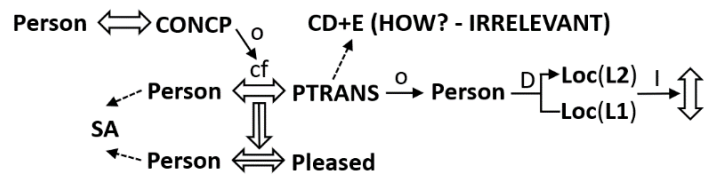

**Fig. 20.** The representation of "Person *wants to* go from L1 to L2." "HOW? - IRRELEVANT" indicates that there may or may not be a means for Person to really go from L1 to L2, and it is irrelevant to the conceptualization of "want." In this figure we removed the details of the SA and the INTRUMENTAL case for clarity.

Fig. 21(a) shows a related concept, "CAN," in a related description, "Person *can* go from L1 to L2." CAN is known as a modal verb which connotes something in the future. It implies something like "If Person *wants* to go from L1 to L2, there exists a means to do so." Fig. 21 illustrates that CAN consists of a few parts:

a) There is a *want* – Person *wants* to go from L1 to L2
b) There is a request for a problem-solving process to find a means to go from L1 to L2
c) There exists a means to do so (a means is found in the problem-solving process)

So, in contrast to *want*, CAN implies there is a means to do what is "wanted." Hence, the "HOW?" part in Fig. 20 is relevant in this case. So, mentally, the IAS calls forth REAC to see if a solution can be found for the "wanted task," which, in this case, is "going from L1 to L2."

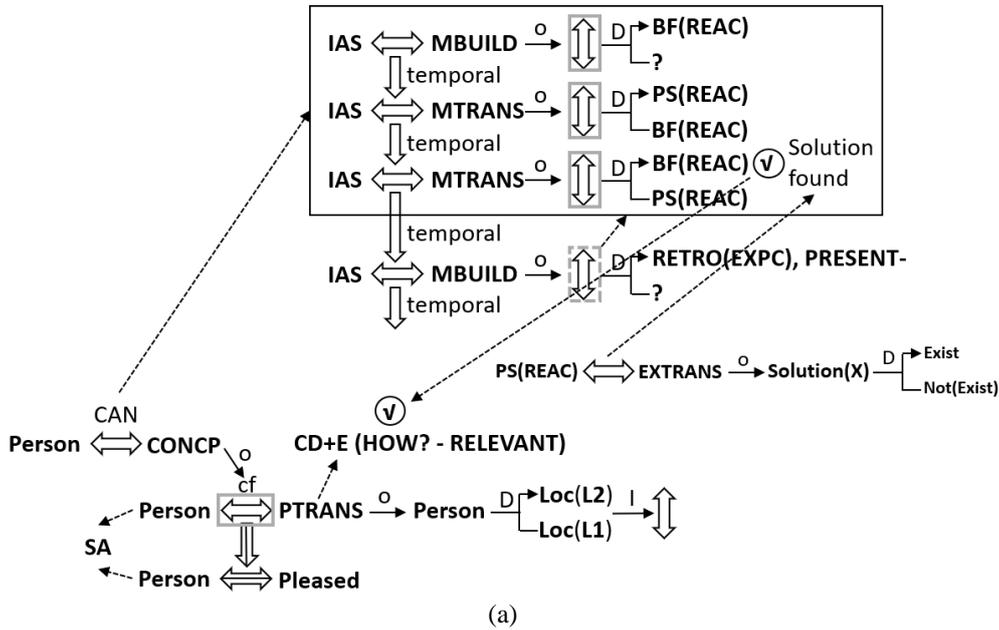

(a)

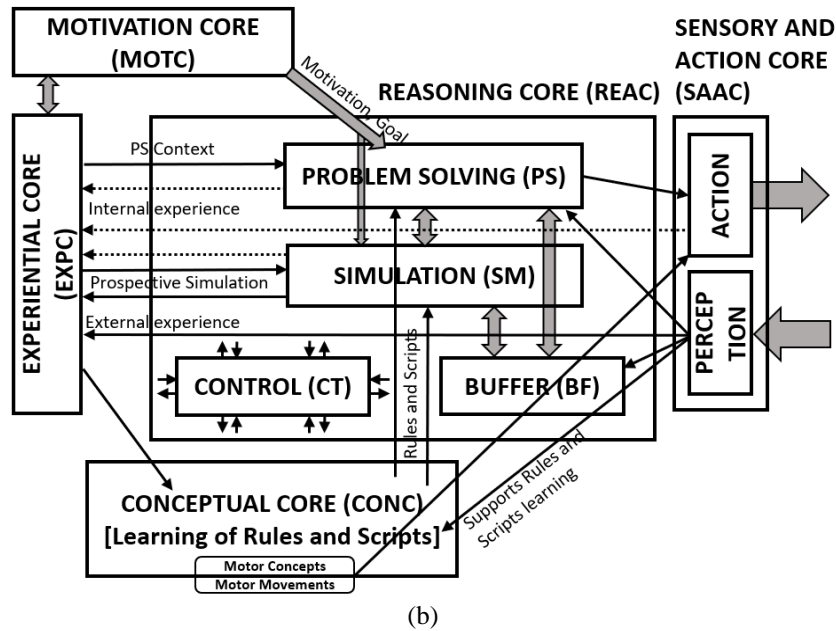

(b)

**Fig. 21.** (a) The representation of "Person *can* go from L1 to L2." (b) The submodules of REAC: PROBLEM-SOLVING (PS), SIMULATION (SM), BUFFER (BF), and CONTROL (CT), and the interactions between MOTC, EXPC, CONC, SAAC, and these submodules.

Hence, the concept of CAN involves operations in certain submodules within REAC. In Fig. 21(b) we expand on the module REAC which consists of the PROBLEM-SOLVING (PS), SIMULATION (SM), BUFFER (BF), and CONTROL (CT) submodules. In an ongoing functioning of an IAS, the process may begin with the PS module receiving a request(s) from MOTC (Fig. 2) to find a solution(s) to satisfy a certain need(s) or to achieve a certain motivational goal(s). (In principle, signals from MOTC are modulated by personality, culture, emotion, and situation before they are finally presented as goals to be tackled by REAC, but for simplicity, we omit these here.) These processes are controlled from and coordinated through CT. PS queries CONC to obtain the necessary Rules and Scripts to try to solve the problem. SM can also be involved in the reasoning process. The request for problem-solving and simulation could also come from BF, which may contain subgoals of earlier problem-solving processes. Information from EXPC and the PERCEPTION part of SAAC may also be involved in the problem-solving process. Normally, when PS or SM obtains a solution, it will be stored in CONC (in a form of a new Rule or Script) for future use (this arrow is not shown in Fig. 21(b) to avoid clutter). PERCEPTION input together with information from EXPC can participate in the formulation of rules in CONC (e.g., new observations allow new generalizations). EXPC also provides the context for PS. The internal experiences of all these reasoning and mental processes are also stored in EXPC, along with external experiences coming from the PERCEPTION module of SAAC. Any generation of ACTION will have proprioceptive information which is considered an internal experience to be stored in EXPC as well. CT contains the reasoning and mental processes to be discussed in later sections, such as the reasoning process of Fig. 37 and the Mental Experiments of Section 7.2. These reasoning processes are represented in CD+ forms, similar to those in Figs. 19 and 21(a).

CT is connected to all the modules – MOTC, EXPC, CONC, SAAC, PS, SM, and BF, not shown explicitly to avoid clutter. As will be discussed, in our framework the reasoning processes are learnable as a result of experience, with modifications made to the reasoning processes represented in CD+ forms, and this information for learning comes from EXPC.

In the problem-solving process, there may be a need to consult the particular situation in the external world the "want" description is referring to. I.e., to ascertain that Person CAN indeed go from L1 to L2, PERCEPTION in SAAC inputs the situation in the real world and sends it to PS(REAC) to find a solution.

Therefore, in Fig. 21(a), it is shown that the meaning of CAN involves the IAS first MBUILDing the "Person wants to go from L1 to L2" conceptualization in BF(REAC). Then, this representation is MTRANSed to PS(REAC) to find a solution (this process employs an EXTRANS (EXistential TRANSformation) for its representation that will be explained later in connection with Fig. 26). A solution is found, and it is MTRANSed back to BF(REAC), and the idea of CAN is established. It is possible that there could be a situation such as "Person thinks that she can go from L1 to L2, but actually she cannot" – there could be some facts unknown to the IAS such that even though the system returns a solution, in actuality the solution does not exist.

In Fig. 3, it is shown that in a general situation, certain preconditions and states *enable* a certain cause to be effected. In the causal framework discussed in [10], [92], these states are called "synchronic causal conditions." A state or a collection of states need not just be enabling one cause to take place, as depicted in Fig. 3. It could be enabling a sequence of causes/actions/events to take place. This is shown in Fig. 22 in which *enable* is allowing a certain solution to be available, which could consist of many steps of actions and state changes, to take place. The example is "The door between L1 and L2 being unlocked *enables* Person to PTRANS from L1 to L2." (We use an English statement – "door between L1 and L2 is unlocked" - to describe the conceptualization involved for clarity.) The symbol for *enable* is the usual symbol for *cause* with a short horizontal bar placed on top. *Enabling* can also bring about a state change, not just effecting an action. *Enabling* can be thought of in counterfactual terms: "Had certain action or state not been present, this action or state would not have been possible." In the usual causal scenario of an action causing another action, or an action causing the change of state, it can also be thought of in counterfactual terms, "Had that action not been taken, there would not have been the subsequent action or state." Therefore, *enable* and *cause* connect at the counterfactual level.

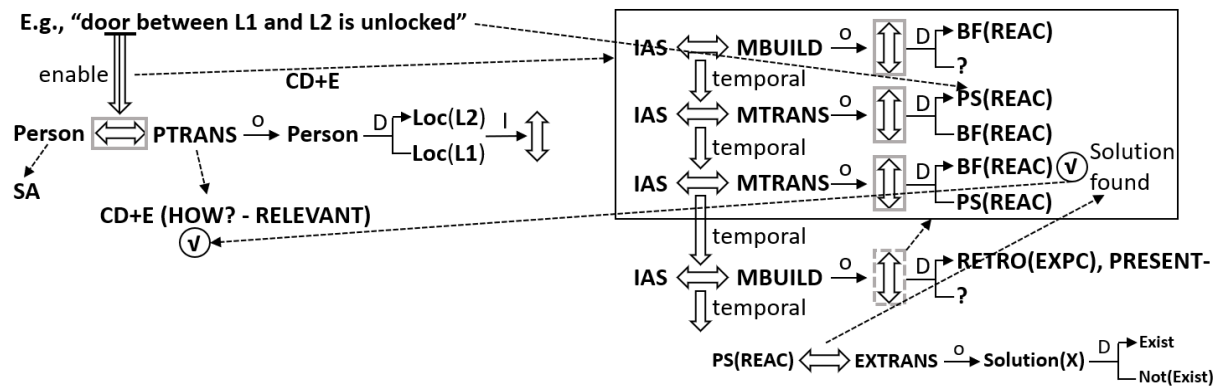

**Fig. 22.** CD+E of the concept of *enable*. The symbol for *enable* is the symbol for *cause* with a short horizontal bar on top. "The door between L1 and L2 being unlocked *enables* Person to PTRANS from L1 to L2."

As for the representation of the concept of *cause* itself, Appendix A lists it as a near ground level concept and Ho [10] has provided a CD representation, capturing a reasonable aspect of it, though it could be further improved in future work.

The various Meta-Functional Concepts represented and discussed in this section – *will, may, if, want, can, enable* and *cause* – will form parts of the representations for functional concepts in subsequent discussions. We will now begin the discussion of some of these functional concepts.

## 6.2. The Functions of a Box

We begin our discussion on functional concepts using the CD+ representation by considering the simple example of a box, an artifact that has a regular shape as shown in Fig. 23(a). This discussion is also applicable to a natural object such as a stone that has an irregular shape. One of the most often used configurations for problem-solving in the early stage of research in AI is the stacking of two boxes shown in Fig. 23(b) [68]. Typically, the purpose of using that configuration is to see how an AI system, in a problem-solving process, can achieve something like that as an end state. A typical predicate logical representation is often used to represent the state as "On(A, B)," which connotes certain spatial relationship. We will use this configuration as a good example of the function of a box. What is the function of box B in this situation? It is used to "enable box A to stay at a certain height." Our focus will be on how to represent this function as well as the function of a box in other situations.

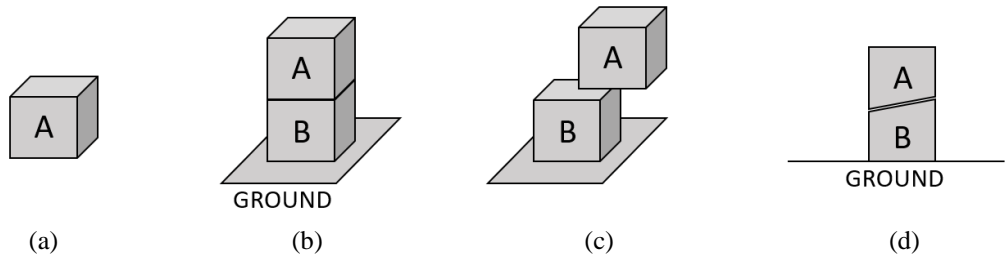

(a)          (b)          (c)          (d)

**Fig. 23.** (a) A box. (b) Box A is on box B. (c) An off-centered or misaligned box A above box B. (d) Box A and box B both have slanted bottom and top surfaces respectively.

It is also instructive to distinguish the representation of a certain function on the one hand, and on the other hand, the conditions for the function to be satisfied given a certain object or a certain arrangement of objects. E.g., suppose in Fig. 23(b), box B is made of paper and box A is made of a stone-like material. B may fail to achieve the function of "enabling box A to stay at a certain height" as box A may cause box B to collapse and fall to the ground itself. Thus, one condition for B to be able to provide that function is that it must be made of a strong material. Box B also must have a significant thickness before it can raise the height of A to a significant level.

Similar situations are shown in Figs. 23(c) and (d). In Fig. 23(c), box A is placed quite off-centered from (and quite "misaligned" with) box B. Box A may fall off and thus box B cannot fulfil the said function. In Fig. 23(d), box B has a slanted top surface (shown in a side view for clarity). In subsequent discussions, we will switch between 3D and 2D depictions). If the surface is not too slanted or the friction of the material involved is high, then perhaps box A can be supported for a prolonged period of time on top of box B, otherwise it may slide off the top of box B. Therefore, "On(A,B)" may not be On(A,B) forever. The precondition(s) for a stable positioning of box A on top of box B could be stated such as in the example of Fig. 5.

Other than the function of "enabling another appropriately placed box or other kinds of object to stay at a certain height" like in the situation of Fig. 23(b), there is a number of other possible functions of a box. For example, a box can

a) be used to increase the height of a person so that the person can see far
b) support a person in a sitting position (if the box has suitable dimensions and hardness)
c) be used to block a path
d) be pushed to hit another object
e) be used to crush things (by swinging and crushing, like a hammer, or just pressing it downward on things)
f) be thrown at something to damage it or someone to hurt her
g) be used to contain things (if empty inside)

The above is not exhaustive and we will discuss some of these in detail in subsequent discussions.

### 6.2.1. The Support Function of a Box

In this section we will discuss the first function of the box mentioned in the previous section, "enabling another appropriately placed box or other kinds of object to stay at a certain height" (Fig. 23(b)), which is what is usually understood as "a box *supporting* another object at a certain height." (In Ho's early work [4], the functional definition of *support* was also concocted and employed in a function recognition scenario, but it was couched in implicit procedural representations. In the subsequent discussion, we will demonstrate how this can be done in an explicit conceptual and cognitive manner.) Before we delve into the discussion concerning function, it will be instructive to consider some fundamental grounded concepts as shown in Fig. 24.

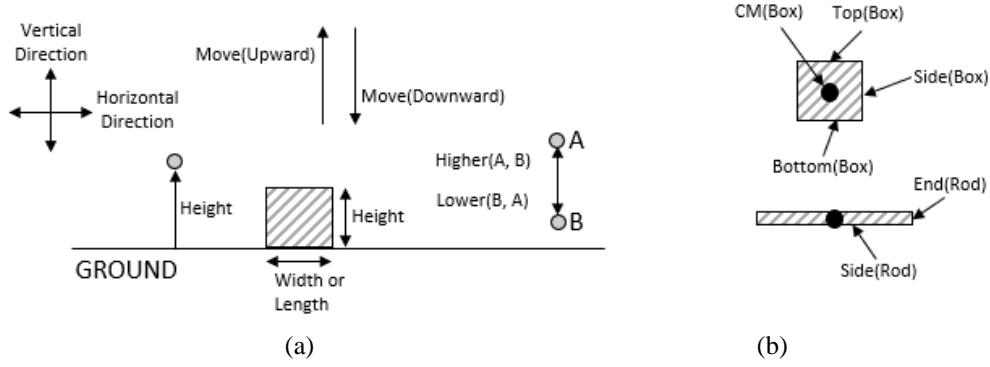

(a)                                                                (b)

**Fig. 24.** (a) The various ground level concepts as a result of a preferred direction brought about by gravity. An object with a hatched interior represents "any" object, not necessary having the exact shape shown, using the convention of[21]. (b) Ground level names for different parts of an object.

In our earthly environment, one important feature is gravity and it defines a direction for many human and object activities. Gravity is listed in Appendix A as a ground level concept. Gravity engenders the concept of GROUND as a spatial reference plane, which is of course a "ground" level concept itself. In the 2D side view representations of Fig. 23(d) and Fig. 24(a) we depict GROUND as a horizontal line. With GROUND comes the idea of Vertical Direction, and hence the Upward and Downward movement directions. Horizontal Direction is a direction that is Parallel to the GROUND (Parallel is a ground level concept listed in Appendix A and defined in [21]). The location of a point or an object, situated away from GROUND in the Upward direction, is also referred to as the Height (as in "how high" it is from the ground). A point or object can be Higher or Lower than another point or object depending on its Height. The dimension of an object, especially one that is sitting on GROUND, measured in Vertical Direction, is usually called the Height of the object (as opposed to Length and Width, which measure its dimensions in the horizontal directions). An object with a hatched interior represents "any" object, not necessary having the exact shape shown, using the convention of [21]. All these above concepts are considered ground level concepts (see Appendix A).

In Fig. 24(b) we show a number of ground level concepts related to labeling the different parts of an object, specifically for a box-like object (though similar concepts are applicable to other-shaped objects as well). These include Top, Bottom, Side, and CM (Center of Mass). For an object with a significant Long-Axis, like a rod-shaped object, (see Fig. 27, Appendix A, and [21]), the surface that is parallel to that axis tends to be labeled Side, and that which is parallel to the short-axis tends to be labeled End.

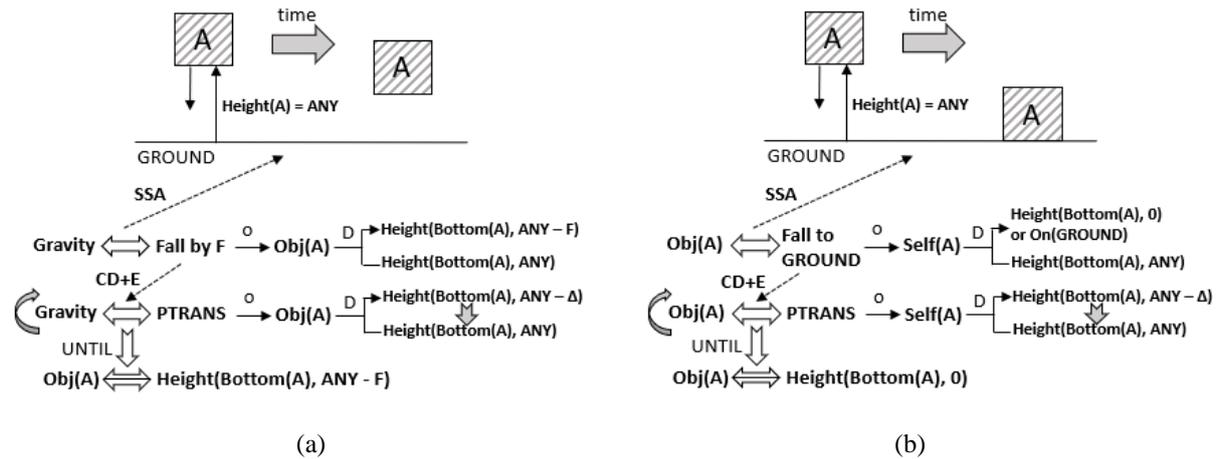

(a)                                                                (b)

**Fig. 25.** (a) An object, Obj(A), falls by distance F. (b) Obj(A) falls to GROUND. In (a), we assume the agent of the fall, Gravity, is known. In (b), an alternative way of representing the event is depicted, in the event that the agent of the fall is unknown or of no concern, and "Obj(A) falls by itself." In both (a) and (b), the SSA (Situation Structure Anchor) contains a temporal change, indicated by the thick horizontal arrow between the two configurations – the initial and final configurations.

Objects in a gravity environment tend to fall toward GROUND. Fall is not a ground level concept and can be represented in terms of the above ground level concepts as shown in Fig. 25.

Fig. 25(a) illustrates the concept of "Object A falls by a distance F". In this CD+ representation, we assume that the agent of the fall is known, which is Gravity. The ground level concept of PTRANS is used to elaborate the CD+E on the detailed process of falling. The curved arrow on the left indicates an iteration of the little PTRANS process, and the downward pointing arrow between the Heights on the right indicates that the value of the Height on top (the end of one step of the PTRANS process) would replace the value of the Height below for the next iteration. The concept of UNTIL, listed in Appendix A as a near ground level concept, is used to terminate the process. There is a variant of SA we introduce here, which is a Situation SA (SSA), which indicates a situation or configuration involved. The SSA in the figure includes an indication of temporal change, with the thick horizontal arrow linking the initial and final configurations.

Fig. 25(b) illustrates the concept of "Object A falls to GROUND." The terminating condition is when Obj(A) reaches GROUND. In this representation, we illustrate how, in the event that the real agent of the fall is unknown (or is not of any concern to be specified), the object can "fall by itself." This also illustrates a basic principle of our framework, which is that the same situation or concept can have more than one representation, depending on the viewpoint or information available to characterize it.

### FACTUAL (FACT) Representation

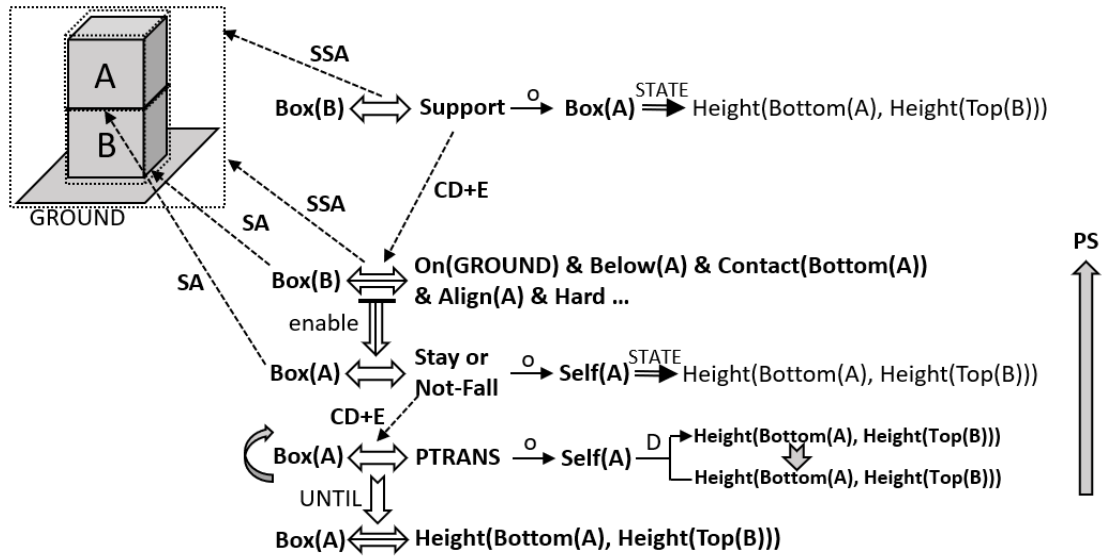

### COUNTERFACTUAL (C-FACT) Representation

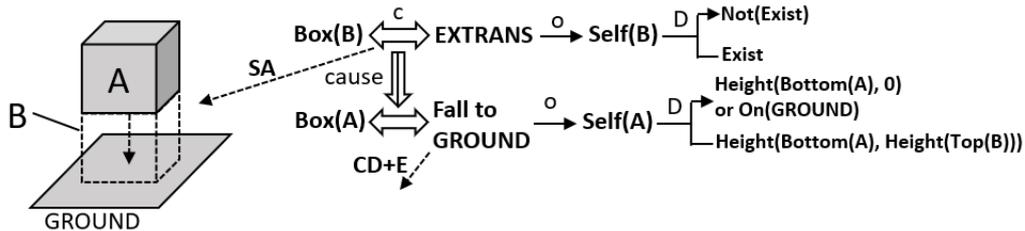

**Fig. 26.** The representation of the supporting function of a box B for a box A. SSA is Situation SA, as distinct from the usual SA that describes particular objects. The C-FACT representation corresponds to "if Box(B) does not exist (in this configuration), it *causes* Box(A) to fall to GROUND." STATE is a parameter – the state - associated with an object, such as POSS-BY. The predicate Height(Bottom(A), Height(Top(B))) means the Height of the Bottom of A is the same as the Height of the Top of B. PS = Problem-Solving process.

Fig. 26 illustrates the representation of the function of a box, B, involved in the physical configuration of Fig. 23(b), in which a box A is placed on top of the box B. We can conceive of this configuration as having been obtained from RETRO(EXPC). This would be a particular instance, before any generalization. Therefore, the boxes have their specific dimensions and arrangements with respect to each other. This configuration in RETRO(EXPC) could have come from an encounter and observation in the real world, or a conceptualization generated by CONC in certain circumstances (such as the current circumstance in which we concoct the configuration for the purpose of discussion and analysis). Both boxes have the same dimensions. First, there is an SSA that depicts the configuration involved. The configuration could likewise be described using predicate logic as shown. However, because there are many aspects of the configuration and the objects involved in the configuration, it is difficult to obtain a complete description of the physical configuration using predicate logic alone (e.g., the three dimensions of the objects, how the objects are aligned with each other, etc.). Therefore, an analogical SSA representation in the form shown is the best way to capture the Situation involved, and the predicate logical form is shown to be incomplete with a series of periods, "…"

One thing to note is that there is a scalar property of Box(B) that has to be specified, which is that it is hard enough to support A, represented here as Hard(B). Interestingly, hardness can also be defined functionally, which is that property that enables the object that is hard to defy compression, which in this case, defy compression from an object placed above it under the force of gravity. (Hard is listed in Appendix A as a near ground level concept.) The Hard predicate used in specifying the hardness of B does not describe the degree of hardness needed, so its presence is just to alert the reasoning system that there is such a scalar property that has a role to play in the support function of a box.

However, as mentioned earlier in the introduction portion of Section 6.2, as we discuss function, we are investigating in the event that a certain function is present, how that function is represented. There are conditions that must be present before a certain function is possible, such as in this case, box B being hard and has certain spatial relationship with the object being supported, and in the case of Fig. 5, the conditions stated for the situation involved. Therefore, the specification of the conditions for box B to really realize the support function is separate from the representation of the function of support per se.

Also, note that in our predicate logic-like representation for the specification of various states, in this figure and in other figures, we omit the agent involved, in this case, Box(B), as it is already present on the left side of the conceptualization, which is a conceptualization that means "has State." So, in Fig. 26, On(GROUND) is taken to be On(Box(B), GROUND) in the usual predicate logic form, as Box(B) is the "agent" already stated on the left side of the conceptualization.

So, the SA, namely the "On(A, B)" structure in Fig. 26, represents a specific instance of the configuration that allows the support function to be present. If there is a need to represent a more general situation which includes many possible configurations (e.g., box A with smaller dimensions and box B with bigger dimensions, or vice versa to certain extents, or certain hardness range of box B relative to the weight of box A, etc.), something like the range constructs in [21] with regards to spatial relations, in which a range of values is specified for certain parameters, could be used. In general, if more complex specifications of the general characteristics of the objects involved that can participate in the support relation of Fig. 26 are needed, they can also be stated in CD+ representational form. We relegate the further exploration of this to future work.

The main CD+ expression in Fig. 26 states that the State of Box(B) *enables* a certain action that obtains on Box(A), and this action turns out to be a Stay action in which Box(A) stays at a certain height from GROUND. We term this the FACTUAL (FACT) representation (as opposed to a COUNTERFACTUAL – C-FACT - representation to be discussed shortly). As mentioned above, the State of Box(B) described by the predicate logic expression is incomplete, and it points to the SSA. "Stay" is in turn elaborated through the ground level concept of PTRANS (i.e., "moving" from one place to the same place, which implies Stay.)

In Fig. 26, the C-FACT representation of the function of Box(B) is also illustrated for the current configuration. An EXistential TRANSformation (EXTRANS) action is used, which converts something that exists (e.g., a material point, or light) to something that does not (e.g., the material point *dematerializes*, or the light is turned off) or vice versa (see [10], [103] for a discussion on the representation of materialization and dematerialization). The C-FACT description is couched in the description of a future possibility, namely "if Box(B) does not exist (in this configuration), it *causes* Box(A) to fall to the ground." One can think of this as a "real" dematerialization, or a mentally imagined one, MBUILDed within the simulation and reasoning system, REAC, of the IAS involved. In the figure, we have omitted the detailed CD+E of the Fall to GROUND process as it is the same as that in Fig. 25.

The fact that we could have both a FACT and a C-FACT representation in Fig. 26 illustrates the principle within our framework that there could be more than one way to represent a certain concept, as in the situation of Fig. 25.

In Fig. 26, we show that the entire CD+ expression of Box(A) on top of Box(B) is also usually encapsulated in a one-word functional concept Support – "Box(B) Supports Box(A)." The CD+E of Support would be the entire CD+ expression we discussed above. We also show a PS (Problem-Solving) arrow going upward, meaning that the CD+ representation can be used in a backward chained manner to provide a solution given the desired effect.

The representation of the Support concept as illustrated in Fig. 26 possesses the RUUI property required of functional representation as discussed in Section 2. First, it allows for the *recognition* (R) of a Support instance as stipulated by the representation. The functional representation also allows one to reason about the functioning of support – e.g., Box(B) *enables* Box(A) to stay at a certain height, and the C-FACT representation stipulates that had B not existed, it would *cause* A to fall to GROUND, which reflects the *understanding* (U) of the function involved. It also provides for problem-solving solutions – i.e., if one desires something like A to be sustained at a certain height, one can implement the configuration depicted through the backward chained PS process indicated, which is being able to put the concept to *use* (U). The representation is thus problem-solving actionable. Being able to facilitate the generation of other unseen instances of Support would constitute the *invention* (I) requirement for functional representation. As mentioned in Section 2, we leave the further exploration of this invention aspect to future work, but the generation of novel instances of Support certainly could be facilitated by representational constructs such as that depicted here.

One may raise an issue on the utility of the SA representation in general, and in the configuration in Fig. 26 in particular. Suppose one wishes to know what happens if Box(B) is removed. In this simple scenario, this consequent can be derived from the FACT or C-FACT representation without querying the SA, but shortly we shall see that for complex objects such as a chair, the SA is indispensable for physical reasoning and the derivation of physical consequences. Even in the current situation, we can use the SA to derive, say, the range of possible locations of Box(A) on Box(B) in which the support function is realizable. We will discuss this further in Section 6.2.4.

### 6.2.2. The Support Function of a Pole on a Slanted Object

In this section, we present a variant of the support concept discussed above. In Fig. 27 we show a rod that is used to prop up or support another slanted rod. In our physical reality, objects can also "fall sideways," in addition to falling downward, as in the examples we have explored till now.

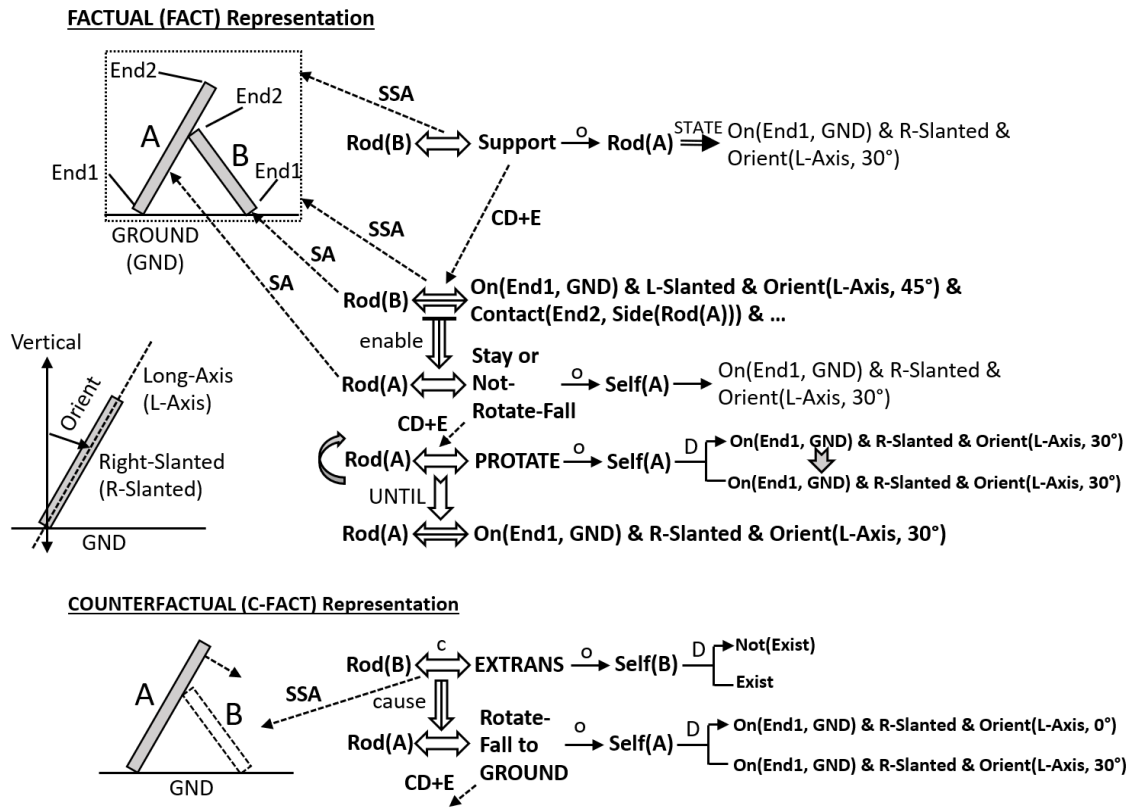

**Fig. 27.** The representation of a variant of Support, in which a rod prevents another one from rotating (or swinging) and falling to the ground.

In Fig. 27 we show both the FACT and C-FACT CD+ expressions for the situation depicted. PROTATE (Physical ROTATE) is a new ground level concept that we introduce here that is the rotating version of PTRANS (Appendix A). Here, we are assuming that when a rod rotates and falls flat onto the ground, the end (say End1 of Rod(A)) that is touching the ground does not slide as there is sufficient friction, so the movement is one of pure rotation. (Refer to Fig. 24(b) for the names of the parts of an elongated object.) Otherwise, we can add another term to characterize it as "rotate and slide." A small diagram on the left side of the figure defines the terms Orientation (Orient), Long-Axis (L-Axis), and Right or Left-Slanted (R- or L-Slanted). Orientation and Long-Axis are listed in Appendix A as near ground level concepts. We have omitted the detailed CD+E of the "Rotate-Fall to GROUND" process of Rod(A) when Rod(B) is dematerialized, but it parallels the processes for PTRANS in Fig. 25.

### 6.2.3. The Support Function of a Box for an Agent – Motivation Concept

In this section, we consider a box functioning as a support for a person who wishes to gain height over the ground level in order to be able to see far, to see over some obstacles, or allow others to see her. Fig. 28 illustrates the CD+ representation of this function of a box.

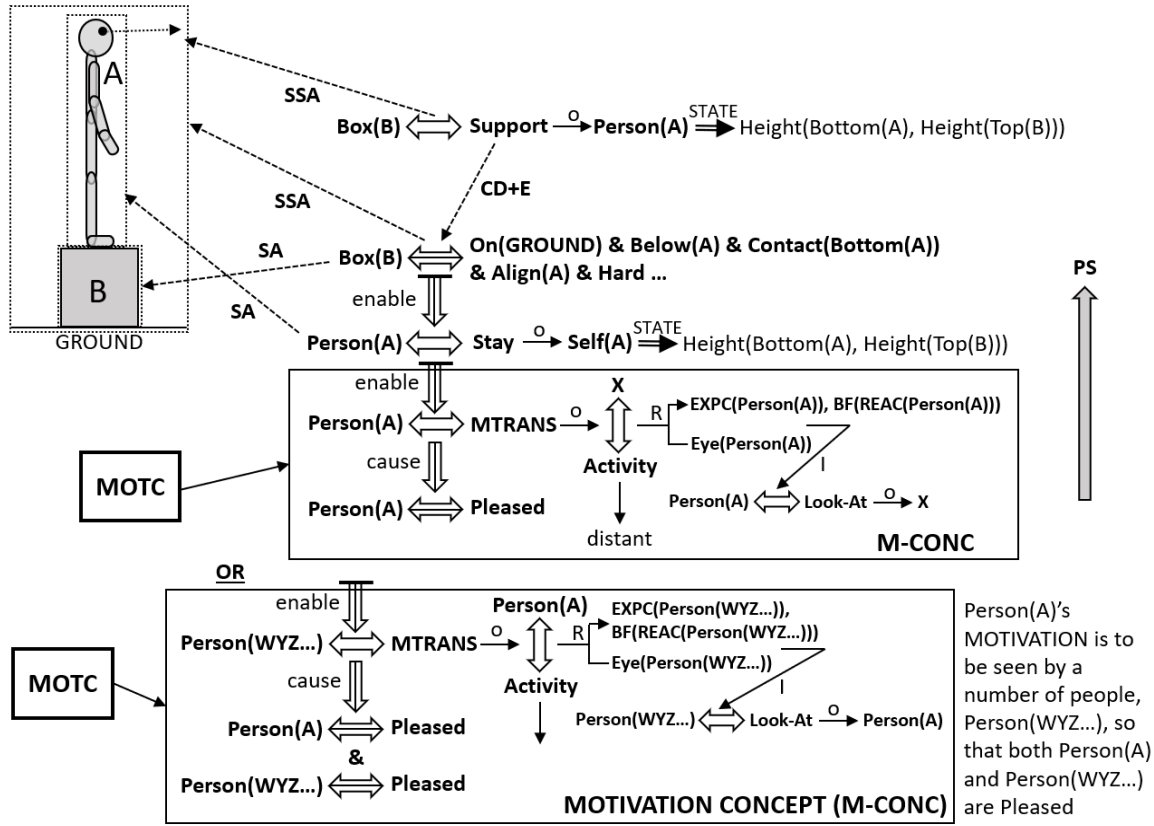

**Fig. 28.** The representation of "Box(B) supports a Person(A)." PS = Problem-Solving.

The core of this representation consists of the CD+ expression of the box supporting the person at a sustained height of the box, much like in Fig. 26 in which a box supports another box. However, because a person is an autonomous agent encapsulating certain internal motivations, the ultimate function of supporting a person, say, in a standing posture at an increased height, other than raising her physical height per se, is usually either to allow the person to be able to see farther than she could standing at the ground level (often for the purpose of seeing over some obstacles), or to allow other people to see her over some obstacles (e.g., because she is blocked by some other people present). Using the idea that "seeing" involves mentally transferring information (MTRANS) about situations in the external world from the eye to certain internal memory of the autonomous agent involved, as discussed in Section 4 (Fig. 13(a)), we add this aspect of the function of supporting a person on the box onto the basic CD+ expression of Support in Fig. 28.

Note that Person(A) must Look-At X in the distance, or Person(WZY…) must Look-At Person(A), before the corresponding Activities can be seen by the corresponding "Eyes" and thereby transmitted to EXPC and REAC. These are shown in Fig. 28. See Fig. 13(a) for a similar situation with the Look-At action. Look-At is listed in Appendix A as a ground level concept.

The need for an agent to see farther or to be seen by others is derived from its motivational center, MOTC. In some situations, it is a primary motivation that drives an agent's behavior. For example, the needs for Food and Safety are primary motivations (Fig. 2), which are "built-in" needs of an agent, often to ensure its better chances of survival. The enjoyment of seeing a beautiful scenery is also a primary motivation (p. 114 of [104]), and this could drive an agent to climb to a higher spot in a situation in which the original height is not conducive for seeing the scenery. Secondary motivations such as seeking money is derived from the process of satisfying the primary motivations – e.g., an agent learns that in order to buy food, money can be used, and hence the need to obtain money arises. Climbing up a box to see farther could also be coming from a secondary motivation – to see if danger is coming in the process of satisfying the primary need of Safety. Secondary needs are often derived from problem-solving processes in the process of either satisfying the primary needs or other secondary needs. Both primary and secondary needs drive problem-solving processes. MOTC thus constitutes an important component of an autonomous system and its integration into the other modules discussed earlier is shown in Fig. 29. Most of the functioning of the modules of Fig. 29 have already been discussed in connection with Figs. 19 and 21. Suffice it here to add that MOTC mainly drives PS and SM in REAC, and an IAS's SAAC output is intimately dependent on the outcome of the processes in PS and SM as driven by MOTC. The MOTC contents could be changed by experience, thus the Learn link from EXPC to MOTC, and the changes and other activities therein are stored as internal experiences in EXPC – thus the Build link from MOTC to EXPC.

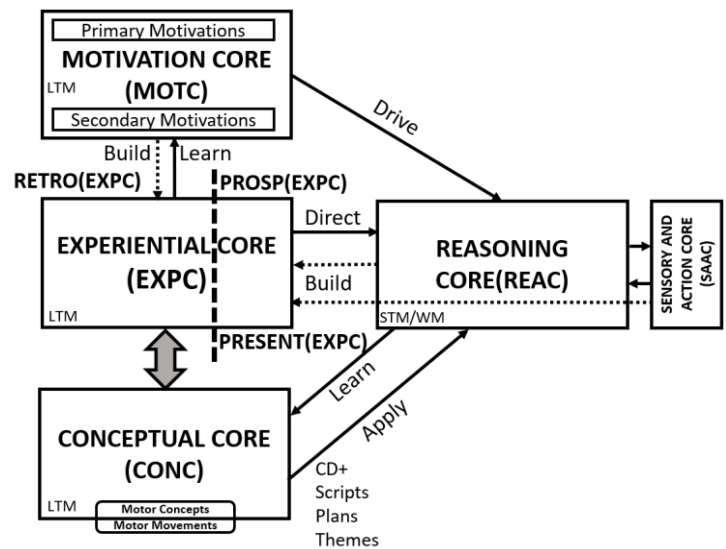

**Fig. 29.** MOTC integrated into the modules discussed earlier in Figs. 19 and 21. MOTC is the main Driver of the reasoning and problem-solving processes in REAC in order to derive solutions that would "Please" the IAS, which may in turn involve actions emitted in the real world.

Hence, in Fig. 28 we indicate the portion of the CD+ expression that is linked with MOTC. MOTC provides what causes an agent to be Pleased and in this case it is the need to see farther, encapsulated in a CD+ representation. This CD+ representation constitutes a "Motivation Concept" (M-CONC), encapsulated in a rectangle in Fig. 28. The M-CONCs involve state that "A motivation of Person(A) is to see the activities happening in the distant and she is Pleased by that," or "The joint motivation of various people, Person(WYZ…) and Person(A), is for Person(A)'s Activities to be seen by the various Person(WYZ…), and Person(A) and Person(WYZ…) are Pleased by this." M-CONCs are themselves functional in nature, thus captured very naturally in our function representational framework. M-CONCs specify what are desired in an IAS's behavior that would allow the IAS to be satisfied, or Pleased, and when this specification is passed to REAC, it works to derive a solution through its reasoning and problem-solving processes (PS and SM), which could involve actions to be carried out in the outside world.

In the above discussion, the "see far" function of a person standing on a box reflects a generic situation. There are of course situations in which the person involved wants to see particular objects or activities over an obstacle or in a distance. Goals such as these can likewise be represented in CD+ accordingly.

As in the cases of Figs. 26 and 27, the functional representation in Fig. 28 is problem-solving actionable, through a process of backward chaining. Depending on the starting point for a need or problem statement, the representation can be activated at different levels to provide the solution. If the need is to raise a person's height to a certain level, say the height of the box, then the representation prescribes that the person must be placed on top of the box, which enables the person to be supported there. If the need is to allow the person to see farther or be seen by others farther, then the representation prescribes that the person must first be supported at a higher height than GROUND, and then that becomes a secondary goal, and the solution to achieve that goal is to place the person on top of the box.

Fig. 30 shows a related concept, Climb. The representation captures the fact that by climbing a box, the person gains height and can be sustained at that height by the box, and that in turn allows her to see farther or be seen by others over some obstacles, which is captured by the representation in Fig. 28 which is now part of the Climb functional concept in Fig. 30 (in this figure only one of the two M-CONCs illustrated in Fig. 28 is shown). There is a causal relation between the Climbing action and the "On(Person(A), Box(B))" consequence. What is of main interest here is the CD+E of the Climbing process.

In Fig. 30, we illustrate the CD+E for Climb in a quasi-CD+ format for clarity. The actions and spatial relations used are ground level or near ground level concepts. PTRANS, Next-to, Above, On, Face, Turn-Body, Lift-Body, and Lift-Right-Thigh (or Lift-Left-Thigh) are listed in Appendix A as ground level concepts. Face is equivalent to Orient but is specifically used for an object that has a "face," like a person. Turn-Body and Lift-Right(Left)-Thigh are ground level actions related to a human body.

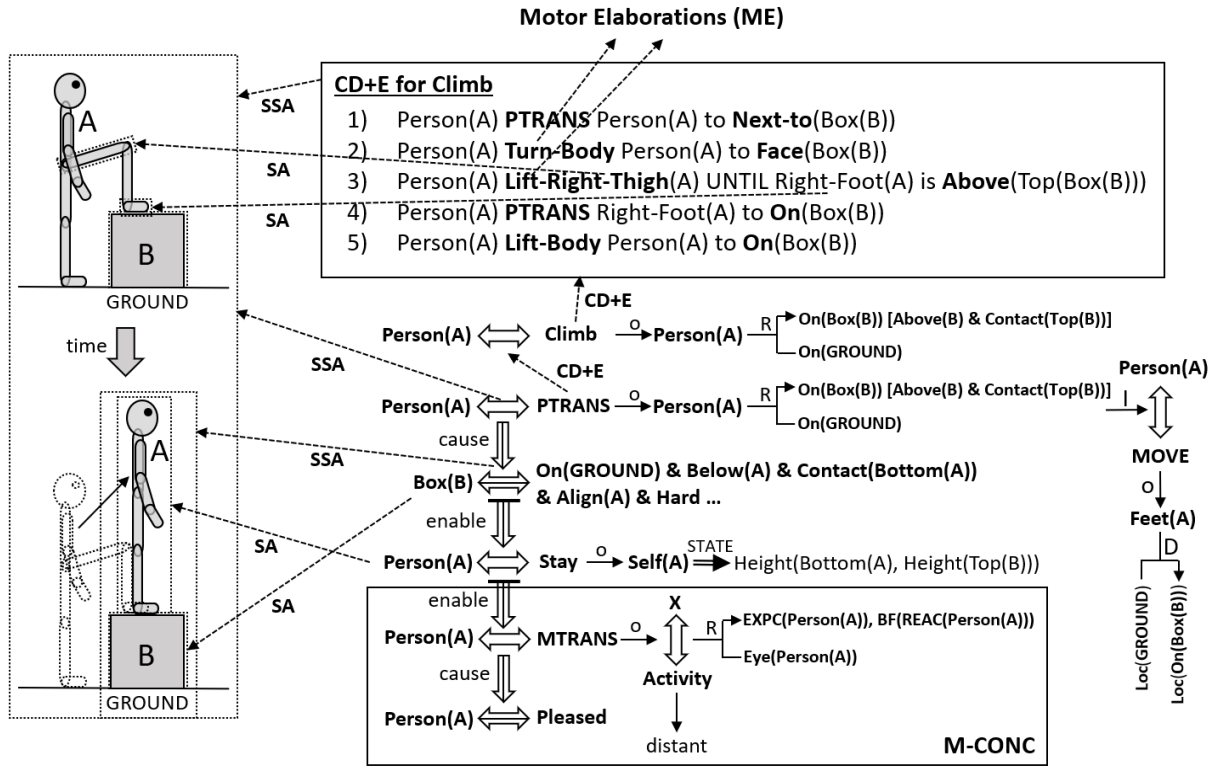

**Fig. 30.** The representation of Climb and its causal consequences.

One important component of the Climb concept is ME (Motor Elaboration). ME consists of two portions. There is a general portion which constitutes certain Motor Concepts, and this is part of CONC as shown in Figs. 19 and 29. These Motor Concepts are explicitly represented in CD+ forms. There is also a specific portion which constitutes detailed Motor Movements to achieve certain states of the body and body parts involved. This portion sits outside CONC as shown in Fig. 29 as it is not "conceptual" but is instead "movement" in nature (also explained in connection with Figs. 5 and 19). It includes, for humans, the skeleto-muscular actions of various parts of the body, and for robots, the motor commands at the body and limbs levels. The details of both the Motor Concepts and Motor Movements modules will be separately addressed in a future paper.

This further extends the problem-solving actionability of the function representation to include *how* to get the person to be placed on top of the box. Of course, Fig. 30 is only one means to do so. There are many other possible processes to achieve On(Person(A), Box(B)) that is not shown in Fig. 30.

In the same spirit as the basic CD representation and the representations depicted in Figs. 17, 18, and 19, the INSTRUMENT (I) Person(A) uses to Climb the box is also depicted. However, this aspect of the representation is not very useful in terms of really elaborating the concepts of Climb to the level that it is actionable for the purposes of detailed understanding (hence supporting "explanation"), problem-solving, and action generation. It is useful perhaps in a language communication process in elaborating on certain aspects of the Climb concept.

There are other uses of the concept of Support in non-physical situations that involve metaphorical transfer of the functional conceptualization involved. E.g., "I support your idea," "I support your bid to become the mayor of the city," etc. The current function representational framework supports the possibility of extending the treatment to such cases in future work.

### 6.2.4. Reasoning with Structure Anchor

Though a full treatment of reasoning with CD+ representations is not the main aim of the current paper, we nevertheless would like to describe a framework that can provide an inkling of how reasoning works with representation to generate inferred consequences and conclusions, given certain situations and configurations, especially in the cases of the physical constructs that we have been dealing with.

In a number of places in the foregoing discussions, we have mentioned how a problem-solving process could use various CD+ representations involved in a backward chained reasoning process (e.g., Figs. 11, 26, 28). This is relatively straightforward as the representations themselves contain forward causal, temporal, and enablement chains of conceptualizations, and a backward chained reasoning and problem-solving process can just capitalize on these chains more or less directly. As for the SA or SSA aspect of the CD+ representation, we have mentioned that we have included it as a grounded description of the structures or configurations involved, for the purpose of "fully defining" the objects, events, configurations, or situations involved (Fig. 5). In this section, we will use a simple example to describe how SA participates in reasoning processes that allows the IAS to forward simulate future consequences of given situations and configurations, and also to backward reason about what needs to be done or what configuration(s) is desired to achieve a certain objective or desired situation.

As mentioned earlier in Section 6.2.1 in the discussion associated with Fig. 26, we separate functional characterization per se from the characterization of the conditions necessary to achieve certain desired functions. For physical constructs such as those we have discussed, there is a need for a Physical Reasoner (PHR) to determine whether a certain configuration can indeed satisfy the desired function as shown in Fig. 31(a). (A Psychological Reasoner – PSR – is needed for reasoning about human behavior.) PHR consults a module that encodes qualitative and quantitative knowledge of physics resident in CONC (not shown explicitly in Fig. 31(a), but described in [4]). The process begins with a Functional Reasoner (FNR) invoking PHR or PSR to determine if a certain desired function can indeed be obtained. FNR accesses the various CD+ representations encoding functional definitions in CONC. In general, the reasoning process will require an internal analogical space to hold the intermediate stages of SA for physical reasoning, and this is the Configuration Array (CFA) (or Situation Array – STA – for cases involving more than individual physical objects) as shown in Fig. 31(a). The CFA or STA is where the SA or SSA type representation may reside in certain stages of the reasoning process.

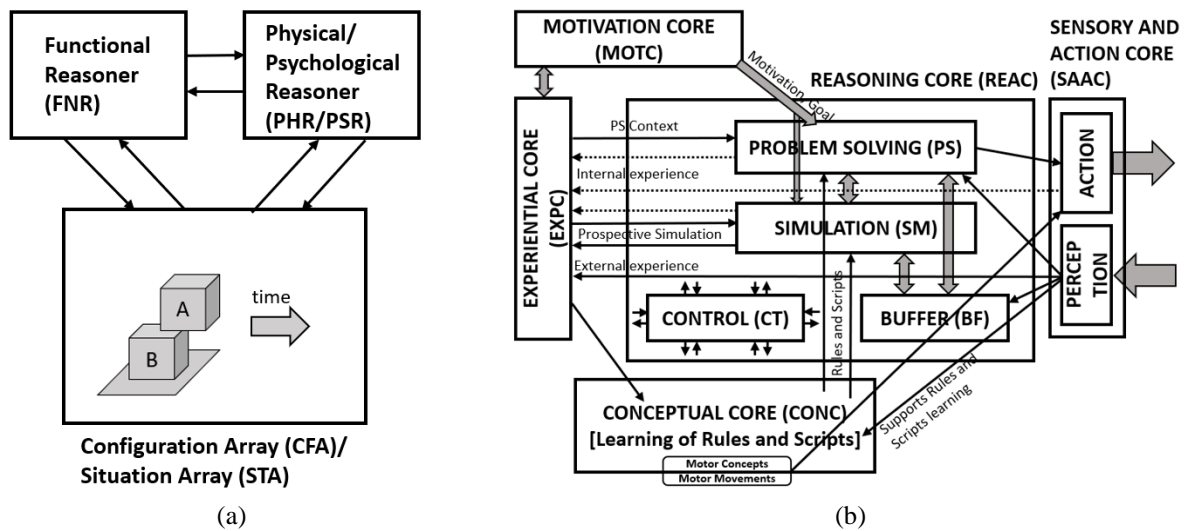

(a)                                                                 (b)

**Fig. 31.** (a) A top level diagram of the interactions between the Functional Reasoner (FNR), the Physical Reasoner (PHR) or Psychological Reasoner (PSR), and the Configuration Array (CFA) or Situation Array (STA) [4]. (b) REAC and other modules from Fig. 21(b), reproduced here for convenience.

The various parts of Fig. 31(b) (same as Fig. 21(b)) correspond to the operations specified in Fig. 31(a). E.g., BF(REAC) corresponds roughly to the CFA, CONC is where the CD+ representations of various functional definitions as well as the physical knowledge are stored, the workings of CONC together with PS and SM modules correspond to the processes in FNR, PHR, and CFA/STA.

In Fig. 32, we depict an *explicitly* stated qualitative physics rule in the CD+ form (stored in CONC) that can be used to reason about a physical configuration similar to that of Fig. 23(c) – A Box(A) is placed off-centered on another Box(B) and it is in a potentially unstable position. CD+ representations are a natural means to encode physical or psychological rules as these rules are functional in nature, consisting of state and causality specifications.

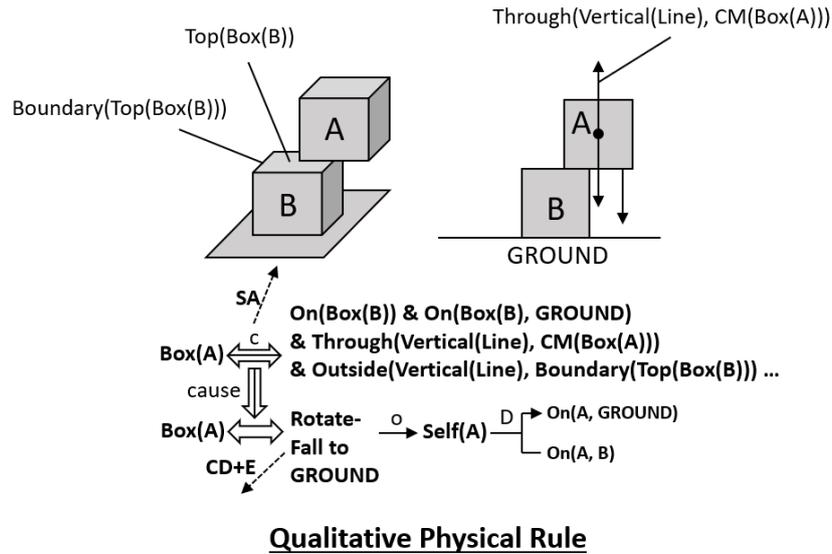

**Qualitative Physical Rule**

**Fig. 32.** An explicitly stated qualitative physical rule for a configuration similar to that of Fig. 23(c), in which Box(A) is placed off-centered on Box(B), and it is assumed that CM of Box(A) is positioned outside the boundary of the top of Box(B), thereby creating an unstable situation in which Box(A) will rotate and fall to the ground. The 2D side view version of Fig. 23(c) is also shown to more clearly illustrate the exact position of CM of Box(A). The description On(A, B) still obtains in this situation as On(A, B) is defined as Above(A, B) & Contact(Bottom(A), Top(B)), so the misalignment does not matter. Contact is listed as a ground level concept in Appendix A.

The qualitative physical rule basically states that "if Box(A) is placed on Box(B) & Box(B) is on GROUND & Vertical(Line) is a line that goes through the CM of Box(A) & Vertical(Line) is Outside the Boundary of the Top part of Box(B), then this will *cause* Box(A) to Rotate-Fall to GROUND." On, Through, Vertical, Line, CM, Outside, Boundary, and Top are ground level concepts listed in Appendix A.

This rule is somewhat specific. Box(A) and Box(B) can actually be replaced by Any Object(A) and Object(B). (Any is listed in Appendix A as a ground level concept or possibly a near ground level concept.) A dual of this rule is that if Vertical(Line) is *Inside* the Boundary of the Top of Box(B), then Box(A) will Stay on Box(B).

In some situations, a physics reasoning engine such as Unity (www.unity.com) can be used to carry out the physics simulation, given a certain configuration. In that situation, the physical rules are not made explicit to the IAS in the form of, say, CD+ representations, but the analogical form of SA is still useful for assisting with the reasoning process.

If the rule is explicit, such as that illustrated in Fig. 32, and if the dual of the rule is also available, it can certainly aid greatly in a problem-solving process in which, say, given an unstable configuration such as that in Fig. 32, a solution can be found to make it stable. The dual of the rule in Fig. 32 prescribes just such a solution – PTRANS Box(A) so that Vertical(Line) is the Boundary of the Top of Box(B).

If the rule is implicit, for example, when they reside in an opaque engine such as Unity, a more involved and "less intelligent" search process can still uncover the solution with the aid of the analogical SA – Box(A) can be moved around on Box(B) until a solution is found.

In this section we have considered two relatively simple objects and a relatively simple configuration to demonstrate the use of SA representations. In Section 6.3, when we consider the function of a more complex concept, Chair, it will be very clear that SA is indispensable for characterizing the functionality of Chair, and for that matter, concepts in general.

As this paper is mainly concerned with the representation of functional concepts, a fuller treatment on the representation of qualitative and quantitative physical knowledge and the accompanying physical reasoning processes are relegated to separate future work.

In Fig. 31, when FNR invokes PHR/PSR to reason about the various consequences of configurations or situations in CFA/STA, FNR can be thought of as carrying out Mental Experiments. "Mental" because the processes take place in the various modules (REAC, CONC, etc.) *within* the IAS. "Experiments" because various configurations, other than what is currently given and perceived in the external world, may be concocted internally to determine the different consequences under physical constraints. CD+ representations can be used to represent these Mental Experimental processes themselves. We will discuss this in more detail in Section 7.2.

### 6.2.5. Other Functions of a Box

In this section we illustrate how some of the other functions of a box listed in the beginning of Section 6.2 can be represented using CD+.

In an event such as "Mary crushes the smaller box with the bigger box by placing the heavier bigger box on top of it and pushing down hard on it," what transpires is the use of a box to reduce the vertical dimension of another box (or another object) to zero or near zero. More likely, in our daily activities, we might use a heavy stone to crush, say a nut, in order to crack it. The functioning of both situations is equivalent. Fig. 33(a) illustrates the idea of using a box to crush another box. Mary's crushing action can be abstractized as Force(F) which is listed as a near ground level concept in Appendix A. Force(F) can be replaced by a detail elaboration of a person using her hand to press down on Box(A). The representation of one of the other functions of a box, "can be pushed to hit another object," could also capitalize on the Force(F) concept here.

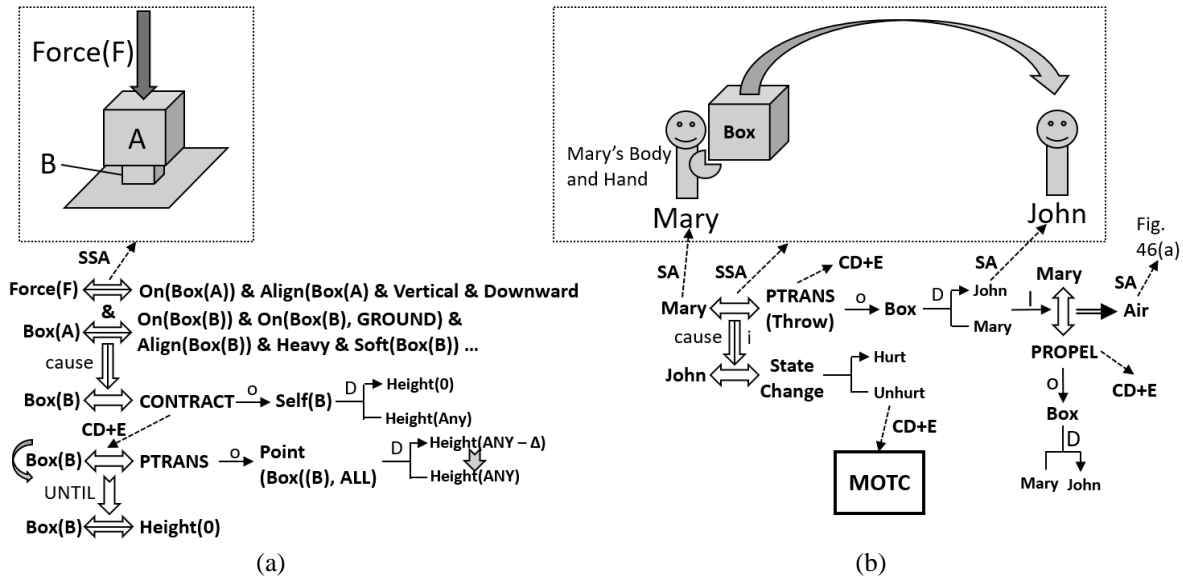

(a)                                                                                (b)

**Fig. 33.** (a) Representation of "using a downward force onto a heavier box to crush a smaller, softer box." Point(Box(B), ALL) refers to ALL the Points on Box(B). Point and ALL are ground level concepts as listed in Appendix A. (b) Representation of "Mary throws the box at John to hurt him." Instead of using an INTRUMENT (I) case to elaborate on a PTRANS as the main conceptualization, an alternative is to use Throw as the main conceptualization and use a CD+E to elaborate on Throw, as discussed in the text. The SA of Air is described in Fig. 46(a).

The action or process of CONTRACT (Appendix B) taking place on Box(B) can be further elaborated to consist of a step-by-step reduction of the Height dimension of Box(B). In Appendix B we also list other kinds of transformation that can take place with regards to the shapes of objects.

Gibson and others [3], [67] discuss the concept of "throwability" as an affordance of some objects. Often, the purpose of throwing something is either to transfer it further from oneself because the item is no longer needed, or to throw it at something to damage it or at someone to hurt her. We consider the event "Mary throws the box at John to hurt him." The CD+ representation of the event is shown in Fig. 33(b).

In the representation of Fig. 33(b), there is an "i" next to the *cause* link to represent the fact that Mary *intends* to cause John hurt ("Mary throws the box at John *to* hurt him"). If the statement is instead "Mary throws a box at John *and* it hurts him," then Mary may not have the intention of hurting John, and the "i" will either be omitted or "i?" is used instead.

In keeping with the convention of Schank [39] who also discusses a similar conceptualization ("John threw a rock at Sam"), we have placed an INSTRUMENT (I) description for the PTRANS involved, and using a primitive PROPEL used in Schank and Abelson [41]. Instead of using PTRANS as the main conceptualization, we could also use Throw, and elaborate it with a CD+E which includes the specification of using Hand(Mary) to effect a PROPEL through Air (i.e., Throw means the use of Hand to PROPEL something through Air). In our framework, PROPEL is not a primitive and instead could be further elaborated using a CD+E. Hurt and Unhurt are listed in Appendix A as near ground level concepts as they can be further elaborated using M-CONCs. These M-CONCs are resident in MOTC. SA for Air is described in Fig. 46(a).

Due to the constraint of space, we will leave the other functions of the box listed in the beginning of this section (Section 6.2) to future work, but suffice it to say that the CD+ representation has sufficient representational power to encode these functionalities.

### 6.3. The Function of a Chair

Chairs are one of the most ubiquitous and important of human artifacts. It serves the function of supporting a human body in a sitting position, so that the person can work comfortably in many daily activities, such as working and dining at various kinds of tables, listening and watching someone on stage for learning or leisure, sitting in a vehicle for transportation, etc. Presumably, before humans invented tools to craft chairs, or when humans found themselves in a situation where chairs are not available, any suitable artifacts such as a stone (say, at the seaside) or a box (say, in a warehouse), can be used for sitting. (In Section 6.2 we therefore list "supporting a human in a sitting position" as one of the functions of a box with a suitable dimension.)

In an early piece of AI work, Ho proposed a computational framework for using a functional definition of a chair for the recognition of chairs along the functional dimension, and demonstrated that the definition can recognize a wide variety of chairs with vastly different visual characteristics [4]. His work was partly motivated by the observation that not only "chair-like objects" that have mainly similar and slightly different characteristics are all labeled "chairs" in normal human language communication, even something like the Balans chair, as shown in Fig. 34, which has totally different visual characteristics, is also called a chair (see also the Internet for the construction of a Balans chair and how it is used). The following are the characteristics of a Balans chair that are very different from that of a normal chair: 1. the seat of a Balans chair is not leveled; 2. there is no back, which normally would convert a chair-like object from the "chair" category to a "stool" category; 3. there is a knee rest that does not exist in normal chairs; and 4. the "leg" support is of a totally different construction than that of a usual chair. Despite this, a Balans chair allows a human to sit comfortably in it to be engaged in various activities. And, despite the lack of a back, partly due to the nature of the human's skeletal anatomy and physiology, and partly due to the slant of the seat, a person can relax her body and yet would not fall backward, thus allowing her to be comfortably supported in the sitting position for prolonged periods of time, unlike in the case of a usual "chair" without a back – i.e., a stool.

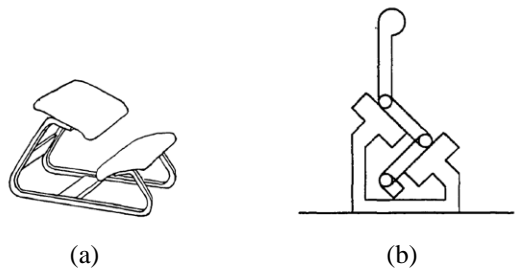

(a)             (b)

**Fig. 34.** (a) A Balans chair. (b) A human sitting on a Balans chair. (From [4]). See the Internet for examples of how a real 3D human body sits in a Balans chair.

Ho [4] thus concluded that other than visual characteristics, sometimes it is the *functional* characteristics that prescribe the category of objects and artifacts. He then proceeded to formulate a cognitive computational process for the recognition of function, specifically for the functional concepts of *support* and *chair* in his work of 1987.

Subsequent to Ho's work, Stark [6] also published work on functional recognition of chairs, though his framework was structured differently than that of Ho. Through the years, there has been some computational AI work on "chair recognition," many pieces of which attempting to use some sort of "functional" characterization [13], [32]. However, all this earlier work on functionality [12], [13], [17], [18], [32]–[38], including that of Ho, suffer from the fact that there are no *explicit* functional definitions, in the form of, say, CD+ representations or in other forms, concocted for performing the recognition of function, and the function recognition framework also cannot be easily extended to be applied in a general way to characterize the almost infinitely large variety of objects or human artifacts. Ho [4], for example, used a non-explicit procedural representation for the functional definition of chair. Procedural representations are generally not amenable to learning, and the framework is also not extendable to other kinds of functional concepts.

The support of a human in a chair is a relatively complex concept. In the case of the support of a human at an elevated level for the purpose of allowing her to see farther or for others to see her in the presence of obstacles, as shown in Fig. 28, the human typically simply adopts a standing (or sitting) posture on the box and lets her eyes collect information from the environment or let others register her image, and her needs are satisfied or motivations catered to. In the case of a chair, firstly the entire body interacts with the chair, and the concept of comfort derived from the support of the chair is intricately linked to the desired states of the various joints and muscles of the human body (Comfort is listed in Appendix A as a near ground level concept). Secondly, a good chair does not only just support the human body at a particular position while the human uses it. The human usually moves about in the chair, continually adjusting her posture in the chair to derive maximal comfort at any given time. Therefore, the function of a chair is much more complex to represent than that of a box support something.

The following lists a number of requirements of a person sitting in a chair (non-exhaustively):

- When a person sits in a chair and works at a table, she would sometimes lean forward (to a certain degree) to write/work on things (with her feet on the ground – or off the ground). Often her leaning forward is supported by her arms on the table's edge. At the same time, the chair needs to be stable under these circumstances.
- Sometimes she needs to, or feels more comfortable to, lean backward to rest with the support of the back of the chair. Her bottom could be at various points on the seat while the chair is still stable.
- Sometimes she would need to lean sideways (to a certain degree) in the process of work, just for the purpose of comfort after having been in other postures for some time, or for the purpose of reaching out for things farther away.
- Sometimes she needs to, or feels more comfortable to, slide her body forward along the seat and lean on the back of the chair that touches a higher point on her back.
- Sometimes she needs to, or feels more comfortable to, bend one or both of her lower legs under the seat to relax, and sometimes she even needs to, or feels more comfortable to, cross her two lower legs together under the seat.
- Sometimes she needs to, or feels more comfortable to, stretch one or both of her lower legs forward beyond the seat to relax, and again the chair has to be stable in supporting this posture.
- The seat of the chair has to be flat, or has a cushion to make sitting comfortable. The seat has to be of a certain size and shape to support the person properly as she moves about in it to achieve the above postures. It should also not be bigger than necessary for her ease of getting in and out of the chair. (Part of the reason for not wanting to be bigger than necessary is due to the economy constraint as discussed in [4].)
- The back of the chair has to be flat and perhaps has a cushion on it to allow the person to lean against it comfortably.
- The legs or other forms of support of the chair have to be at the right length or height to support the seat at the right height for supporting the various activities the person may be engaged in (either for working at a table or other purposes), and they must be arranged in a way that the chair is stable under all the above postures described while the person is in the chair. They must not obstruct the person's legs in going under the seat.
- Whenever a person has adopted a posture for some time, she needs to change posture for comfort.

We submit that the above requirements are all representable within the CD+ framework, even though in subsequent discussions in this paper we will demonstrate only an important subset of them to illustrate the main ideas and relegate the others to future work (e.g., we will not deal with the representation of the idea of cushions on chair to derive comfort, even though that is fully representable within the CD+ framework).

We begin by considering a simplified model of a human body as shown in Fig. 35. We use basically a 2D side view model of a human body instead of a full 3D model for ease of illustration. We will also be using a 2D side view of a chair. For our purpose here, these should be sufficient to bring out the fundamental ideas on the function of a chair in the process of interacting with a human body. The same principles and representational framework is easily extendable to 3D models and 3D situations.

In Fig. 35, it is shown that the human body is divided into four major segments, the head (B1), the torso (B2), the thigh (B3), the lower leg (B4), and the foot (B5). For simplicity, only one thigh, one lower leg, and one foot are shown here. The arm is divided into 3 segments, upper arm (A1), lower arm (A2), and hand (A3). Between the different segments, there are joints: BJ1, BJ2, BJ3, and BJ4 for the body, and AJ1, AJ2, and AJ3 for the arm. The arm will only play a minor role in our subsequent discussion. This simplified sideview model of the human body will not be able to capture the full functionalities of a human body interacting with a chair for sitting comfort, as human moves sideways too in the process of using a chair, and many parts of the human skeletal anatomy and physiology operate in very intricate ways to derive comfort in the body's interaction with the chair, but it should capture the gist of the human-chair functional interaction.

Fig. 35 shows that while the human body is in the standing posture, all four of the joints have to be "tensed" in order for the human body to be stable in that posture. This models why humans prefer not to stand for prolonged periods of time. When a human sits on the floor, BJ1 and BJ2 are still tensed, but BJ3 and BJ4 are *relaxed*, which is a more comfortable posture for the human body. The same goes for the situation in which a human sits on a Box or on a Stool (a "chair" without a back). For the situation in which the human sits in a Chair, all four of the joints can be *relaxed*.

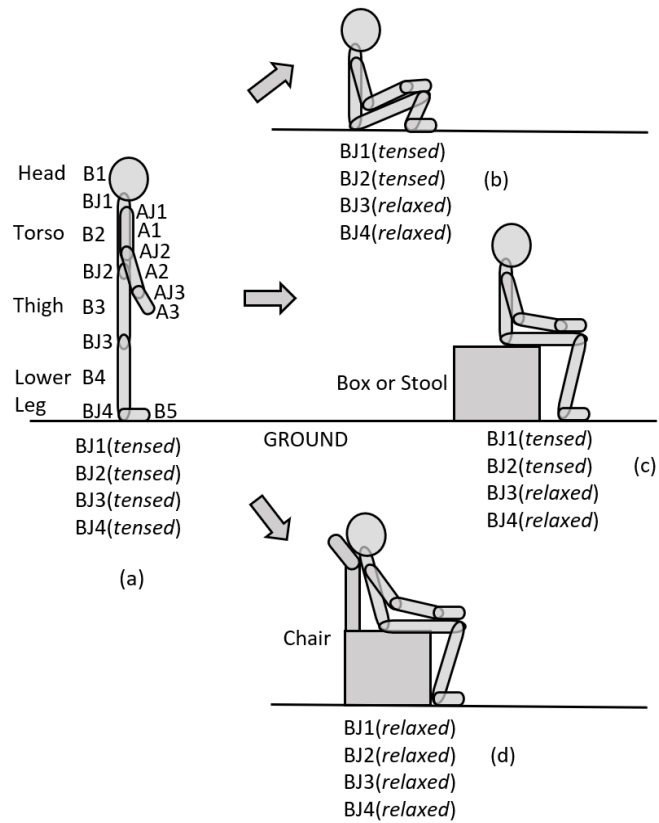

**Fig. 35.** 2D side view of a human body model with various segments and joints, and the corresponding *tensed* or *relaxed* states they are in. Four postures are shown: (a) standing; (b) sitting on the ground; (c) sitting on a Box or Stool; and (d) sitting on a Chair, in a leaned back posture.

In Fig. 35 we also distinguish between a ground/low sitting posture (b) and a "fully raised" sitting posture (c) and (d). Humans could also potentially sit on something like a bar stool, where her feet may not touch the ground, and that would be something higher than the fully raised sitting posture of (c) and (d). The difference between (b) and (c) on the one hand, and (d) on the other, is that there is a back support in (d), including support for the head (B1), and hence all the joints are relaxed, and it would be a preferred resting posture. However, one can also conceive of a ground or lower sitting posture with a back support, such as (b) combined with a wall or some such support for the back of the person to lean on. This way the joint states would be the same as that of (d). Why is it that humans prefer to sit in something like (d), with a fully raised sitting posture, for prolonged periods of time for work or for leisure? There could be several factors at play here: 1. perhaps the ground is considered dirty and humans prefer most parts of her body to be as far away from it as possible; 2. a fully raised sitting posture allows easier transitions between that posture and a standing one for the purpose of moving to and from other locations during work or leisure; and 3. often, a human sits in a chair at a table for work or leisure, and it is often that she would interact with objects on the table in the process. The table, in turn, usually has a fully raised top for perhaps similar reasons – e.g., if a table is used for dining, one would not want the food to be near the ground, and if a table is used for doing other kinds of work where interactions take place with various objects on it, these objects are probably better kept away from the ground for cleanliness and protection (objects placed nearer the ground are more easily displaced or damaged by humans who walk near where the objects are). These are complex reasons why (d) is preferred over (b) with a back support such as a wall, and just by considering a simple model of human body of Fig. 35 does not explain them. In this paper we will take it as a given that a raised sitting posture is preferred. (Note that Ho [4] pointed out that in ancient time in some cultures, even royalties prefer to sit on the floor for work or leisure. There could be other complex reasons which we will not discuss here.)

Note that even though this model of a human body as illustrated in Fig 35 is sufficient for the purpose of bringing out the function of a large variety of chairs, including the kind we are considering in this paper, it is not able to bring out the function of the Balans chair in Fig. 34. The function of the Balans chair requires a fuller model of the human body beyond just the joints and the various major segments. It requires a more detailed model of the muscular anatomy – e.g., to take care of the situation of how the back muscles can be relaxed and yet the torso does not fall backward if the thighs are slanted downward. In Ho [4], an attempt was made to use a simplified human body model similar to that of Fig. 35 to bring out the function of a Balans chair. However, a compromise has to be made – a "back" is added to the standard Balans chair to support the backward leaning torso, creating a "modified Balans chair". However, in that effort of Ho, the purpose of demonstrating that objects that look very different can still serve the function of a chair is still achieved because the modified Balans chair still looks very different from an ordinary chair because of the slanted seat, knee rest, and the totally differently shaped legs. Interestingly, there are more recently designed, commercially available Balans chairs that include a back like the modified Balans chair used in Ho [4] (see the Internet). This kind of Balans chair will be able to interact with the human body model of Fig. 35 adequately to realize the function of a chair, which includes allowing a human to relax joint BJ2 while leaning backward and not falling backward.

In Fig. 36, we illustrate the source of the knowledge from which the function of a chair is derived, and the kind of information that is extracted to characterize the function. As we mentioned above, the function of a chair is basically to support a human in a sitting position. However, unlike the relatively simple cases of the functions of the box discussed in Section 6.2, firstly the human body is more complex and has many parts that articulate independently of each other, and secondly the human body adjusts its posture continuously on the chair to derive comfort in the course of performing certain activities such as eating, working, enjoying a leisurely performance, etc. The SAs of the human body and chair are critical in defining this functionality.

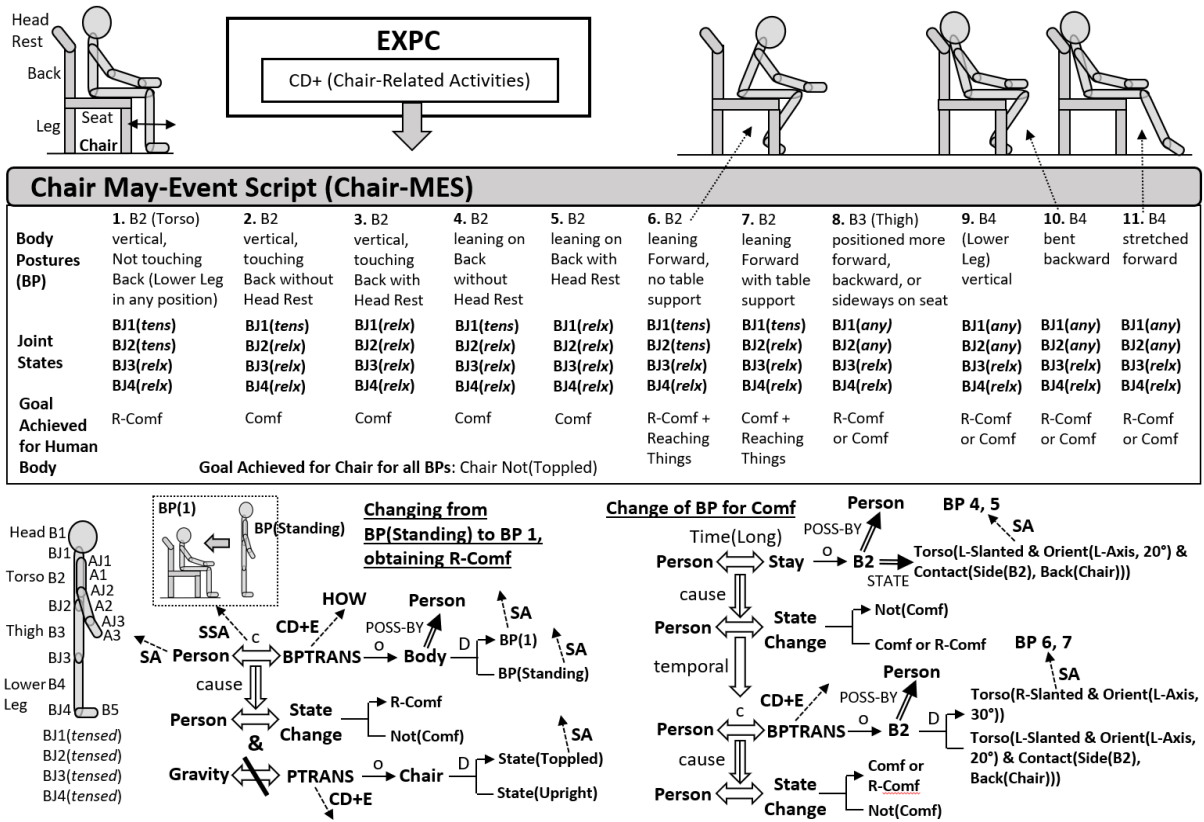

**Fig. 36.** The Chair May-Event Script (Chair-MES) that encodes what *may* happen when a person is interacting with a chair. BPTRANS = Body Posture Transformation. "*tens*" = *tensed*, "*relx*" = *relaxed*. Comf – Comfortable, R-Comf – Reasonably Comfortable. For the CD+ expression in the bottom right of the figure, see Fig. 27 for the definition of L- and R-Slanted, and Orient. STATE is the state of, in this case, B2 (Torso) of Person. Time and Long are listed as ground level and near ground level concepts respectively in Appendix A.

Suppose a person walks into a furniture store and asks herself this question, "Is this chair a good chair to buy?" An AI system could also be posed this question. What would the reasoning processes be like in attempting to make this judgment? Below we lay out the typical steps of consideration for the "goodness" of a chair:

<u>Goodness of Chair Consideration</u>:

1. Consider where and what the chair will be used for. (At home – which room? The study – used for work, or the dining room – used for dining? In the office – where? Etc.)
2. Consider all the instances of how a chair had been used in that context and also project into the future of how the chair may be used in the same context (which could partly be based on past experiences). The *past consideration* is derived from EXPC in which these experiences are stored. The *future consideration* is partly based on these past experiences, and partly based on some consideration of novel uses which requires the activation of PS and SM in REAC.
3. From "2", derive the possible postures of a human body interacting with the chair. Each of these postures presumably provides comfort support for the human body which is why a typical human is either observed to adopt these postures in the chair in the past or is required to adopt these postures in the future simulation that takes comfort into consideration. For each posture, determine if the chair can maintain stability while supporting the human body in those postures, through simulation using qualitative physical knowledge (REAC module – Fig. 31).
4. Decide whether the chair is a good chair to buy based on these considerations.

In Fig. 36, we illustrate what happens from step "3" onward. Suppose there is a means to extract the various typically adopted postures in Chair from EXPC, these postures are listed in a Chair May-Event Script (Chair-MES), encoding what typical postures a person *may* adopt in a chair. We will not go into the detailed process of extraction, but will touch on it cursorily in the discussion in connection with Fig. 37. While the usual CD+ Scripts such as the Restaurant Script (Fig. 15) list events in a sequential order, the May-Event Script (MES) lists a collection of events that *may* happen, not in any particular order. The concept of *may* has been discussed in Section 6.1.

In order to avoid clutter, in Fig. 36 we only illustrate a few of the SAs for the various postures in the Chair-MES. Also, instead of illustrating all the detailed CD+ expressions for all the postures in the chair, we list the various Body Postures (BP) in simple English and only illustrate two examples of the relevant CD+ representations at the bottom of the figure.

In general, a MES would encode both the "may happen" events as well as the states of the events involved. In the top part of Fig. 36, what are shown are the various usual states or BPs of a human body in Chair derived from the events in EXPC.

Fig. 36 shows that there are 11 BPs for the human body in Chair. There are actually many more than the 11 postures possible, but we have collapsed them for the sake of brevity, showing only the main distinct postures. For example, for the very first posture, BP 1, there could be a number of positions for the lower leg, each representing a separate posture. BP 9, 10, 11 list 3 positions of the lower leg, but there could be more positions, and each of these could be combined with other positions for the Torso (B2) and the Thigh (B3). A more general way of representing this would be to use some range specification of the parameters involved along with transformation rules for the various parts, much like how range specification is used in [21] for spatial relations. (The human body model of Fig. 35 is reproduced here in the bottom left of the figure for convenience of reference.)

For each of the BPs, the corresponding Joint States are also listed. The ultimate motivation for a human adopting a certain posture is the comfort she experiences in that posture. Therefore, in the last row, the "Goal Achieved for Human Body" is listed. When no joint or only one joint is tensed, we characterize the human body as Comfortable (COMF). If there are two tensed joints, she is still Reasonably Comfortable (R-COMF). For BP 1 – 5, other than comfort, one of the goals of sitting on a chair is sometimes to interact with objects on a table, for work or leisure, as discussed above in connection with Fig. 35. For BP 6 and 7, besides achieving Comfortable and Reasonably Comfortable states, the human perhaps also achieves Reaching Out for Things.

In the left bottom of Fig. 36, we illustrate the CD+ representation for the event of transforming a human body from a standing posture to a sitting posture, specifically, the BP 1 sitting posture ("Torso Vertical, Not Touching Back of Chair"). The BP 1 sitting posture is quite often the first posture adopted when a person has just sat down from a standing posture. We introduce a new concept named BRTRANS, which stands for Body Posture TRANSformation. There is a CD+E that describes how this particular BPTRANS converts a standing posture (perhaps starting from standing next to the chair) to BP 1 sitting posture in the chair. Similar to the Climb CD+E of Fig. 30, this CD+E provides the "HOW" as indicated in Fig. 36. Furthermore, the very last expression in this representation indicates that Chair must be stable – i.e., it must NOT PTRANS from Upright State to Toppled State. (Strictly speaking, typically there is rotation involved in addition to translation when toppling, so a combination of PTRANS and PROTATE – Fig. 27.) This is the same as the Prevent concept in Fig. 11(b).

In the right bottom of Fig. 36, we illustrate the CD+ representation for one of the events of BP change: "Lean Forward after a Prolonged Stay in BP 4 or 5 – i.e., B2 (Torso) leaning on Back of Chair with (BP 4) or without (BP 5) headrest." As mentioned above, the human body has the property that after a prolonged stay in a certain adopted body posture, which could be Comfortable initially, the person would begin to feel Uncomfortable and has to change her posture to regain Comfort. Here, the conceptualization of having the BP Staying in a long time in a certain posture (BP 4 or 5) is tagged with the predicate description Time(Long). Time and Long are listed in Appendix A as ground level and near ground level concepts respectively. The CD+ representation states that having Stayed in the BP4 or 5 posture for a long time, the Person converted from Comfortable to Not Comfortable state, and if she were to change her posture to, say, BP 6 or 7, she would feel Comfortable again.

This BP change specification can be stated in a more general manner – i.e., a prolonged adoption of ANY BP will lead to Not Comfortable state for the Person involved, and the moment she changes to ANY other BP, Comfortable is restored.

The next step the reasoning system would carry out is the last condition of Step 3 of Goodness of Chair Consideration above – to determine whether given all the usual states and activities a human body would adopt or execute in a chair as specified by Chair-MES of Fig. 36 (including the changes of BPs), the entire configuration/event of human BP/BP change *plus* Chair is stable. This process is carried out in a similar manner as described in connection with Figs. 31 and 32 with regards to the simulation of a certain physical configuration, invoking FNR and PHR in REAC. FNR would invoke Chair-MES from CONC, and carry out the simulation of various specified BP and BP changes in the CFA to see if the entire Chair *plus* BP of Person configuration is stable. The CD+ representation for the stable condition is depicted in Fig. 36 as the last statement in the left group of CD+ expressions – that Gravity does Not Topple Chair.

When determining if a certain artifact can satisfactorily function as Chair, as stipulated in Chair-MES in Fig. 36, Mental Experiments, as mentioned in Section 6.2.4, and which will be further discussed in more detail in Section 7.2, are carried out by FNR, invoking the services of the other modules. We think of the processes invoked to check through all the *may* situations in Fig. 36 as a kind of "experiment" because different configurations are "tested" and the consequences determined. The processes involved in Mental Experiments themselves can also be represented in CD+ representations (see Section 7.2), giving the reasoning processes various important properties: flexibility, adaptability, and learnability. The issue of learnability of the reasoning processes are delegated to future work.

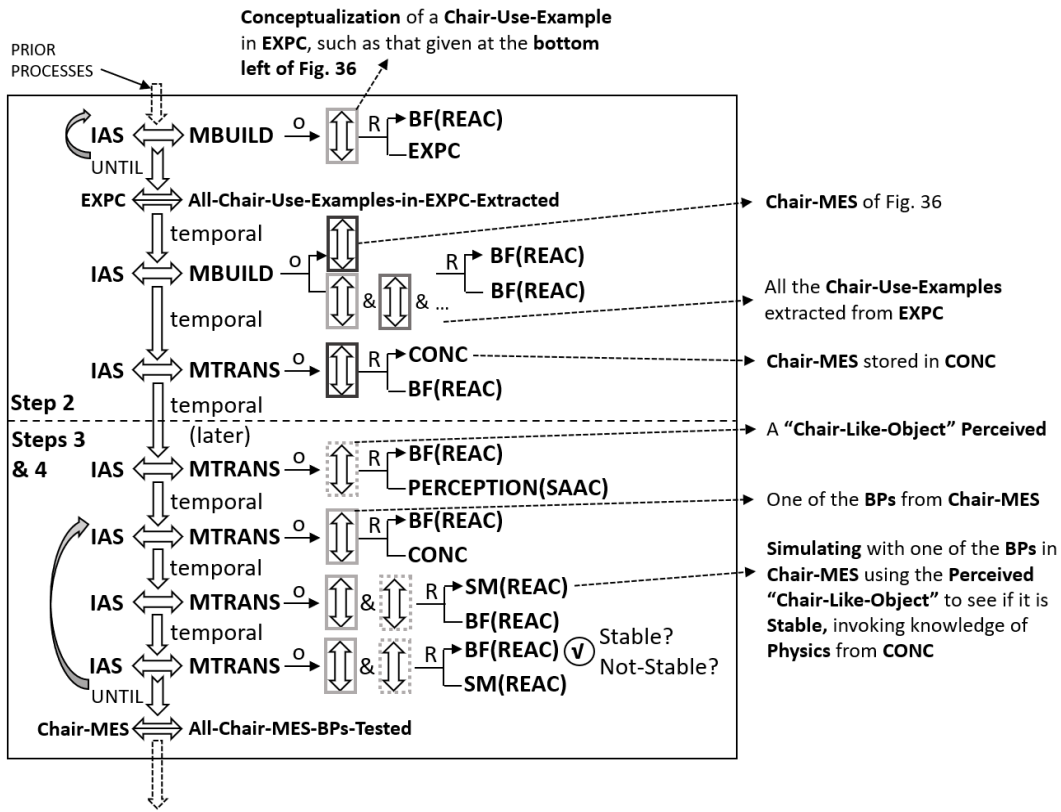

**Fig. 37.** Top level CD+ characterizations of the major parts of Steps 2-4 of the Goodness of Chair Consideration. Step 2 begins with scanning EXPC to extract conceptualizations of all use examples of chairs (*past consideration*). What is not included in Step 2 here is the *future consideration* described under Goodness of Chair Consideration. At the end of Step 2, CONC stores Chair-MES of Fig. 36. In Steps 3 and 4, an observed "Chair-Like-Object" to undergo Goodness of Chair Consideration is tested against all the requirements stipulation by Chair-MES to see if the required Comfortable conditions lead to the stability of the "Chair-Like-Object" or not.

Fig. 37 encodes the major parts of the top-level processes of Steps 2 – 4 of the above Goodness of Chair Consideration. These mental processes are stored in and executed from CT of REAC as shown in Fig. 31(b). There are finer mental operations that operate within BF(REAC), SM(REAC), etc. that are not shown in the figure. Further finessing of mental operations is relegated to future work.

A complete concept of Chair would consist of the representations in Figs 36 and 37, the representation for Step 1 of the Goodness of Chair Consideration (not depicted in this paper), as well as the representations for the structure causals of the various parts of Chair as will be depicted in Fig. 40.

The SA portion of the CD+ representation was first defined in Fig. 5, and the issue of its utility was discussed in Section 6.2.4. As can be seen in this section, SA representations are indispensable in characterizing the functionality of Chair in its totality (in this case, in supporting the physical simulation/reasoning to determine the stability of the Chair-Person configuration involved), hence providing the necessary information for a human to assess the "goodness" of a chair, or for that matter, any artifact or natural objects, for satisfying her needs.

Fig. 38(a) illustrates BP 7 of Chair-MES of Fig. 36 – leaning forward with table support. This state is more Comfortable than leaning forward without the table support, as BJ2 can be relaxed. Also, with a table support, there is a possibility that even a Chair that is potentially unstable – with a single leg and a small base area as shown in Fig. 38(a) – could be stable even when subject to BP 8 – moving the Person's Thigh or the whole body around in the Seat.

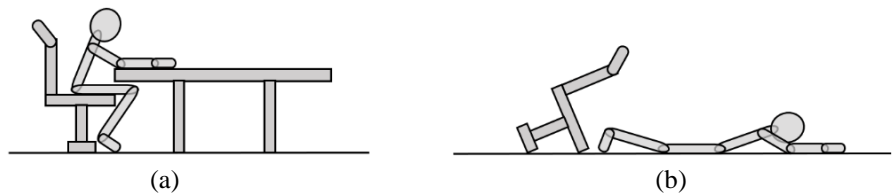

(a)                                                                    (b)

**Fig. 38.** (a) A BP in which there is table support (BP 7). The potentially unstable chair could be relatively stable in this situation. (b) Without the table support, the potentially unstable Chair Topples and Person falls flat onto GROUND when BP 8 is adopted – moving the body around in Seat of Chair.

However, without the table support, the output of the physical simulation of a configuration in which a Person adopting, say, BP 8 (with the Thigh or the whole body being moved along Seat of Chair to different positions) would be as shown in Fig. 38(b) – the Chair Topples and the Person Falls flat onto GROUND.

Fig. 39 illustrates a Not Comfortable Chair and Table. As specified in Chair-MES of Fig. 36, BP 10 dictates that one of the postures of the lower leg that the human body prefers to adopt, after the lower leg has been in some other positions for a prolonged period of time, is to bend it backward. A usual four-legged chair shown in Fig. 36 (only two legs are shown in Fig. 36 because of the 2D side-view depiction) allows the lower leg to bend backward and go "below" Seat. However, the chair depicted in Fig. 39, which has a bottom portion that is totally enclosed, does not allow this repositioning of the lower leg. Furthermore, if this is used in combination with a table that has its bottom also totally enclosed as shown, then BP 11 of Fig. 36, with the lower leg extended forward for comfort, also becomes impossible. The table in Fig. 39 is then also not a good table in supporting a human working in a chair next to it. (We will not discuss the function of Table in this paper, which is relegated to future work.)

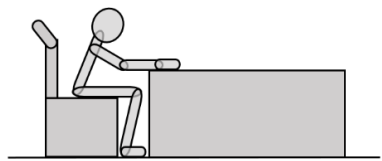

**Fig. 39.** An uncomfortable Chair and Table – BP 10 and 11  cannot be realized.

With the CD+ representational framework put to use in representing functional concepts, the earlier-mentioned deficiencies with regards to earlier efforts in functionality, and in particular in representing the functional concept of Chair, is addressed.

## 7. Structure Causals, Mental Experiments, and Functional Segmentation

In this section, we discuss a number of important concepts – structure causals[6], mental experiments, and functional segmentation – in connection with the characterizations of functionalities. We will also discuss further applications of the CD+ function representational framework to two functional concepts – a tool used for piercing and a jet engine.

### 7.1. Structure Causals of Structure and Parts

In the foregoing discussions, we considered the causal objects involved in certain functionalities as an undifferentiated item, whether it is a simple object such as a box used to support something, or a complex object with distinct parts such as a chair used to support another relatively complex object, a human body. As long as the object involved is not just a dimensionless point, it will have parts that participate in certain functions, or certain aspect of the overall function. For a chair, obviously the three major parts - Back, Seat, and Legs – all participate in different aspects of the overall function of supporting a human body in a comfortable manner. Even for something as simple as a box, there are different sides and corners, which are its "parts," though they are not as distinctively separated parts as Back, Seat, and Legs of a chair. When a box participates in the function of support, perhaps we can identify the top surface as providing the contact to the object supported in order to hold up the object, and the bottom surface provides the contact with the ground to hold up the entire box and the supported object. The middle portion provides the bulk of the material that supports the top surface. The function of these "parts" may not be very distinctive and is hard to characterize, but if a box is hurled at someone and one of its sharp corners hurts that person (Fig. 33(b)), then the function of that particular "part" becomes very distinct.

In general, when an object is selected or built for certain functions, its structure as well as its various parts are meant to engender certain causalities, i.e., the entire structure or its parts are used to effect or enable some events, leading to certain desired states. We term these causalities brought about by certain structural properties *structure causals*.

We begin by using a structure discussed above, Chair, to illustrate this concept. The various CD+ expressions in Figs. 40(a), (b), and (c) taken together describe the functions of the various parts of Chair. "Chair" typically has very distinctive parts – Legs, Seat, and Back. These parts can be segmented relatively easily visually. However, another way to segment an object is through *functional segmentation*. This will be discussed later in Section 7.4 after our discussion on the structure causals of a chair.

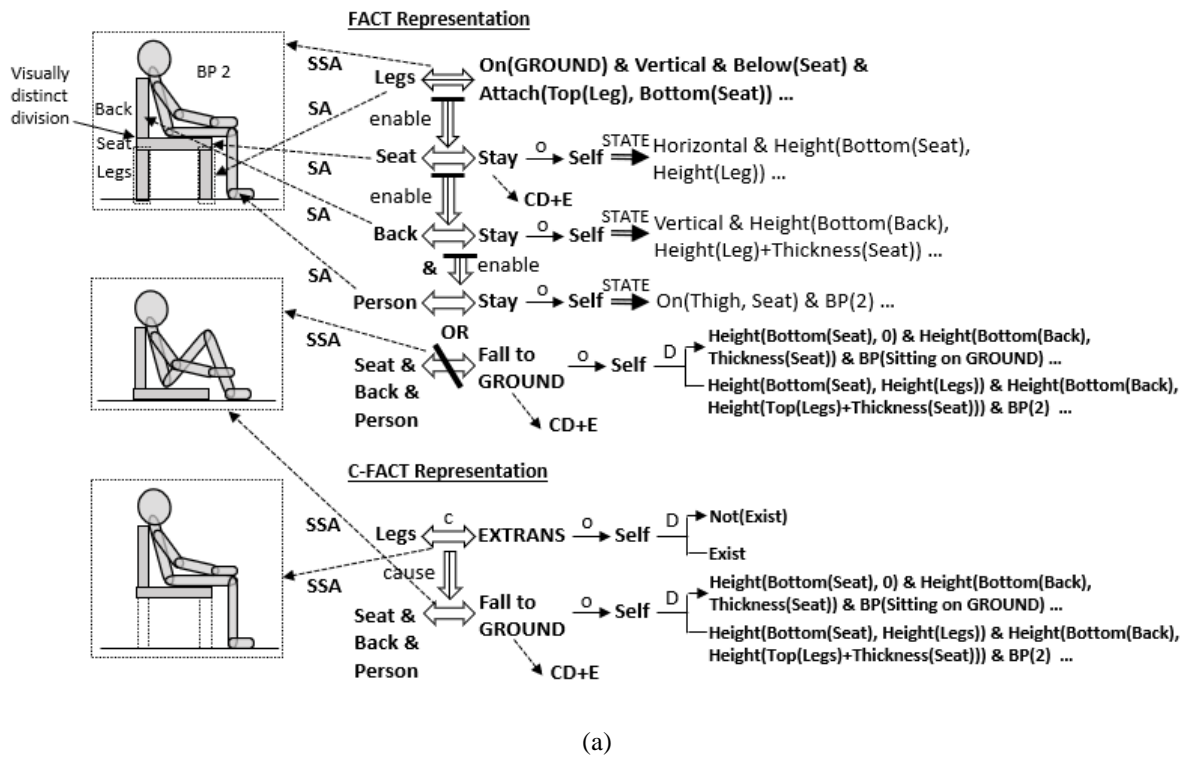

(a)

---

[6] "Causal," as currently used in the noun form, refers to a certain word expressing a cause, such as *since*, *therefore*, *for* (collinsdictionary.com). We extend the usage here.

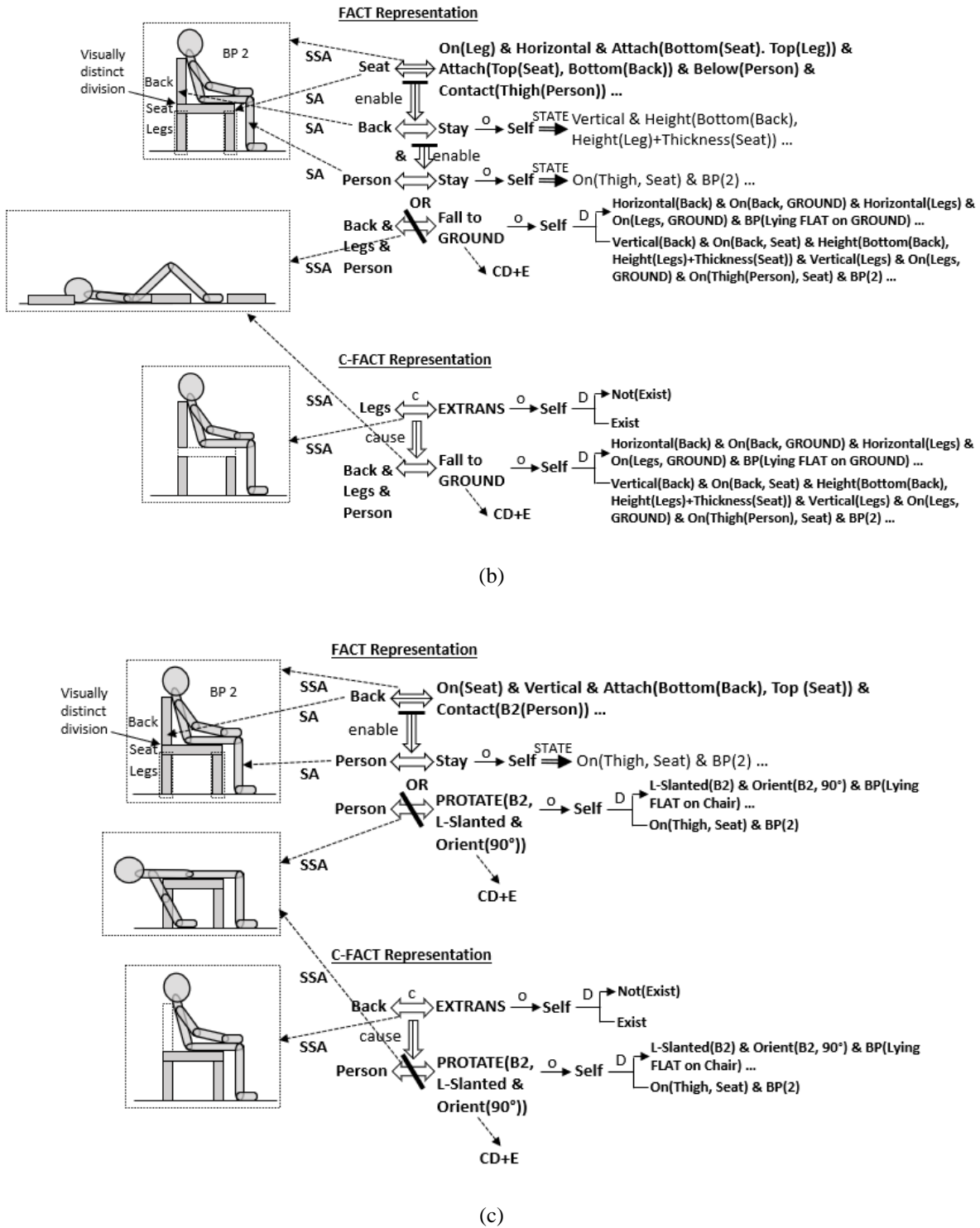

**Fig. 40.** The representation of the functions of the various parts of Chair. (a) The function of Legs. (b) The function of Seat. (c) The function of Back. See Fig. 27 for the parameters associated with PROTATE. FACT = Factual, C-FACT = Counterfactual.

So, let us assume that the three major parts of a chair – Legs, Seat, and Back – can be identified and isolated (say, visually, based on visual heuristics and perhaps visually distinct divisions shown in the figure). We group the various legs of a chair together as a whole – Legs (in 3D there will be 4 legs). Fig 40(a) shows the function of Legs in a FACT form and a C-FACT form. An important concept, Attach, is used as part of the description of the state of Legs, which is listed in Appendix A as a near ground level concept. (Attach has important and interesting representational issues associated with it which will be further investigated in future work. See [10] for some earlier work on Attach.) What the FACT CD+ representation in Fig. 40(a) describes is that Legs *enables* Seat to stay at a certain height, and in turn Seat *enables* Back to Stay at a certain height *and* Person to stay in BP 2 on the chair. Note that Seat *enables* both Back to Stay at a certain height *and* Person to Stay on Seat and in the posture BP 2, but Back also has a role in *enabling* Person to Stay in BP 2. So, the attainment of BP 2 on the part of Person is due to *both* Seat and Back. Strictly speaking, the last Stay, the Stay associated with maintaining Person's BP, should be "StayP," which is a slight variant of the other Stays in that as distinct from the usual Stay defined in Fig. 26 which dictates that the *location* of the object involved does not change, StayP instead dictates that the *BP* of the person does not change. A variation of the FACT description is that the Legs *enables* the Seat, Back, and Person NOT to Fall to GROUND, indicated as the last CD+ statement in the FACT representation as a disjunction of the statement above. Note that in this last statement, in the predicate logic specification of the DIRECTION case of the before and after states of the Seat & Back & Person assembly, only the predicates that change as a result of the Fall to GROUND action are indicated. As mentioned earlier, the SAs involved encode the rest of the details, and the predicate description is incomplete, indicated with a "…" We also introduce a new BP, BP(Sitting on GROUND) as the final state of the Person's body after Fall to GROUND.

The C-FACT description of Fig. 40(a) states that if Legs are not there, Seat, Back, and Person will Fall to GROUND. Similar to the description in the FACT representation, Seat & Back & Person will adopt a certain final configuration as described by the SSA and the process of Falling to GROUND is described by a CD+E. A physical reasoning process such as that described in Fig. 31 can be used to derive the final state of the Seat & Back & Person assembly when the Legs are removed.

Figs. 40(b) and 40(c) show, for Seat and Back respectively, the corresponding FACT and C-FACT CD+ representations. Note that in Fig. 40(b), a new BP is used, namely BP(Lying FLAT on GROUND). As in Fig. 40(a), the predicates in the DIRECTION cases do not describe all the details of the SA involved, only those that change. In Fig. 40(c), Torso (B2) of Person falls "backward" when Back of Chair is removed, and the resultant BP is BP(Lying FLAT on Chair). The falling behavior of B2 is similar to the PROTATE of Rod in Fig. 27, therefore PROTATE is used to describe the action involved.

Thus, taken together, the CD+ representations in Fig. 40 constitute the structure causals of Chair and its attendant parts. And as mentioned in the discussion in connection with Fig. 37 in Section 6.3, these structure causals of Chair are parts of the total concept of Chair, as they encode the functioning of the parts of Chair - the other parts of the total concept being the representations depicted in Figs. 36 and 37, as well as the representation for Step 1 of Goodness of Chair Consideration in Section 6.3 not depicted in this paper.

## 7.2 Understanding Functionality through Mental Experiments

In Section 6.1, we demonstrated how the CD+ representation can be used to represent certain mental constructs and processes which are used to capture a variety of Meta-Functional concepts such as *will, may, if, want, can,* and *enable.* CD+ representations can also be used to represent the "mental" and reasoning processes that take place in attempting to understand the function of objects and its parts such as those discussed so far. As we have seen, one way to characterize the function of certain parts of a chair (or for that matter, the function of the chair itself) is to consider the counterfactual situations – i.e., if those parts of the chair are not there, what would happen. The consequences of these counterfactual situations can be determined through performing Mental Experiments – i.e., considering what would happen, based on known laws of physics, say, if certain configurations are present or are changed. These Mental Experimental processes are in turn representable using CD+ representations and are stored and executed from CT in REAC, as shown in Fig. 21(b).

Mental Experiments have been mentioned in Section 6.2.4 in connection with the discussion on physical reasoning carried out with SA, and in Section 6.3 in connection with the reasoning associated with Chair function. Unlike in earlier sections, in which the Mental Experiments involved were in testing the behaviors and physical consequences of various configurations, in this section, the experiments involve *dissecting* the given artifacts to understand the functions of the parts. This section illustrates specifically how these Mental Experimental processes are represented in CD+ to serve the *function of counterfactual reasoning* to uncover the functions of the various parts of an object. This further enhances the *understanding* of the functions involved.

Fig. 41 illustrates that the process begins when some PRIOR PROCESSES arrive at a point at which the system MBUILDs a concept in which Legs of Chair are to be removed to determine the consequences. This concept of removing Legs is first MBUILDed in BF(REAC). The concept is then MTRANSed to the simulation module – SM(REAC). The results of the simulation, which is the SA in the top right corner of the figure, is MTRANSed back to BF(REAC). Taking the configuration sent for the Mental Experiment and the results of the experiment, the system then MBUILDs the C-FACT representing the structure causal of the legs of the chair, which in this case is the C-FACT representation of Fig. 40(a).

Regarding the PRIOR PROCESSES that take place before the starting point, they could involve another higher level of mental processes that direct the entire Mental Experiment which, at this point, decides to determine the consequence of what happens when Legs are removed. Having completed this analysis, depending on what the current task at hand is, the system may go on to use the results of this simulation and the attendant C-FACT rule to participate in other reasoning processes, or it may continue to explore the structure causals of the other parts of the chair, etc. Suffice it to note that all these mental processes at various levels are representable within the CD+ representational framework.

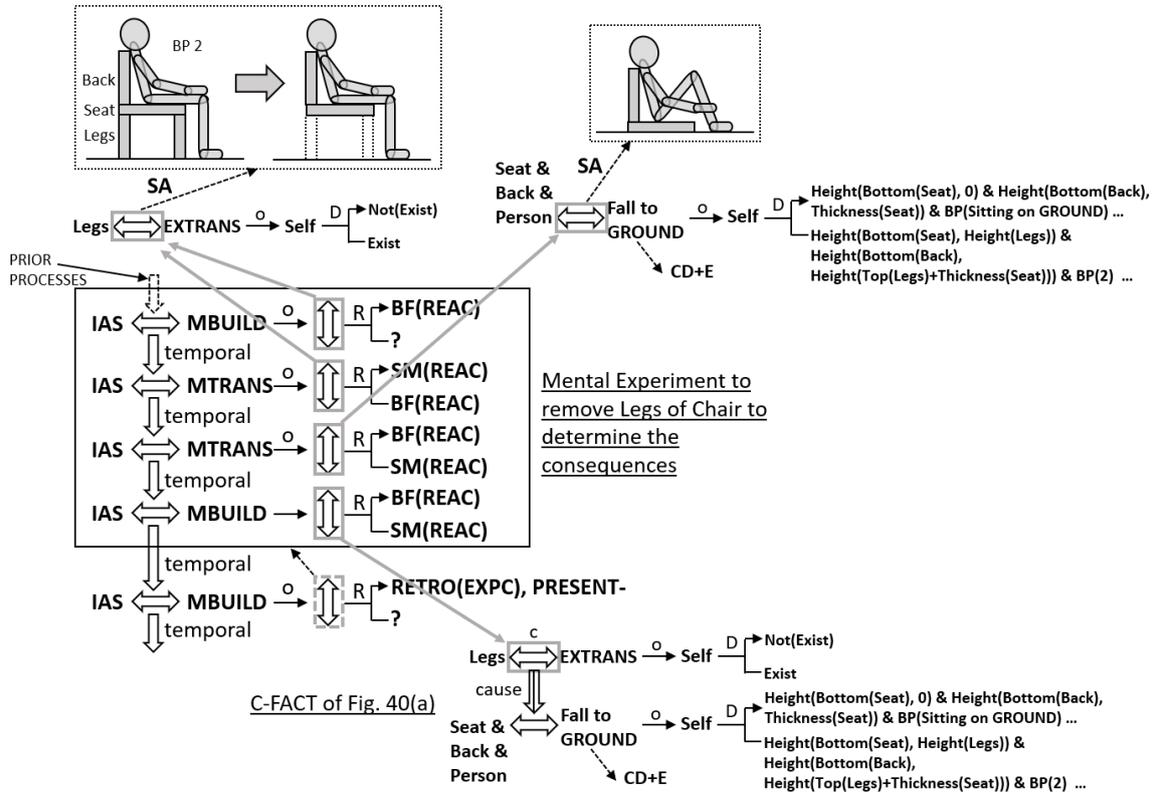

**Fig. 41.** Mental Experiment to remove Legs of Chair to determine the consequences, and then to build the C-FACT representing the structure causal.

Other than using a counterfactual process such as this, there are of course other ways to reason about the structure causals and functions of the various parts of an object or the object itself, perhaps through reasoning by analogy, reasoning through category inheritance, etc. Again, suffice it to note that the CD+ constructs also have the power to represent the reasoning (mental) processes involved in these other various kinds of reasoning. Applying CD+ to represent these other kinds of reasoning processes is relegated to future work.

As mentioned in Section 6.3, by representing Mental Experiments in CD+ representations, we provide the reasoning processes with various important properties: flexibility, adaptability, and learnability. The issues of learnability of the reasoning processes are delegated to future work.

### 7.3 Other Possible Considerations on the Structure Causals of Chair Parts

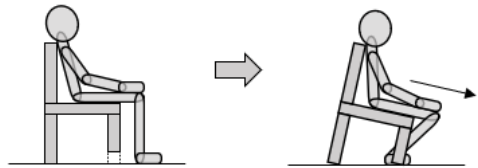

**Fig. 42.** Consequences of different subparts being removed from Chair.

As mentioned above, we are assuming that we can employ certain visual heuristics to identify the three major parts of the chair – Legs, Seat, and Back. What if a reasoning system would like to consider the function of other parts that may obtain from a different way of dividing up the chair? Fig. 42 illustrates just such a situation in which two smaller subparts of two of the "front" legs are conceptually carved out for consideration. In this case, a similar CD+ representation and reasoning process as those described in Figs. 40 and 41 can be devised to represent the function of those subparts – without the subparts, i.e., if those subparts of the legs are removed, the chair will become slanted, and the person may have the danger of sliding off to GROUND. This functional consequence can be represented in the same manner as that in Figs 40 and 41 and other earlier figures.

### 7.4 Functional Segmentation and Reasoning: What Really Supports the Torso of a Person Sitting on a Chair?

The CD+ representational framework also provides for the constructs that enable a system to continue to consider how the support of a person's Torso in a leaned back posture is effected. Earlier we have been considering the functions of the parts of the chair, assuming that these parts have already been identified, possibly through a visually-based segmentation process. In the examples above before Fig. 42, we considered that Back of Chair extends from above Seat, presumably based on some visually distinct partitions, which is found in some real chairs that are produced by some manufacturers, and which can be discerned in our various pictorial depictions. In Fig. 43(a) we show that Back of Chair could also extend below Seat, if the manufacturer so decides to build the Chair by attaching a Back portion in that manner. And, if there is a visually distinct division, the reasoning process we described in Fig. 40(c) would remove that Back according to that visual division and determine the function of that part accordingly.

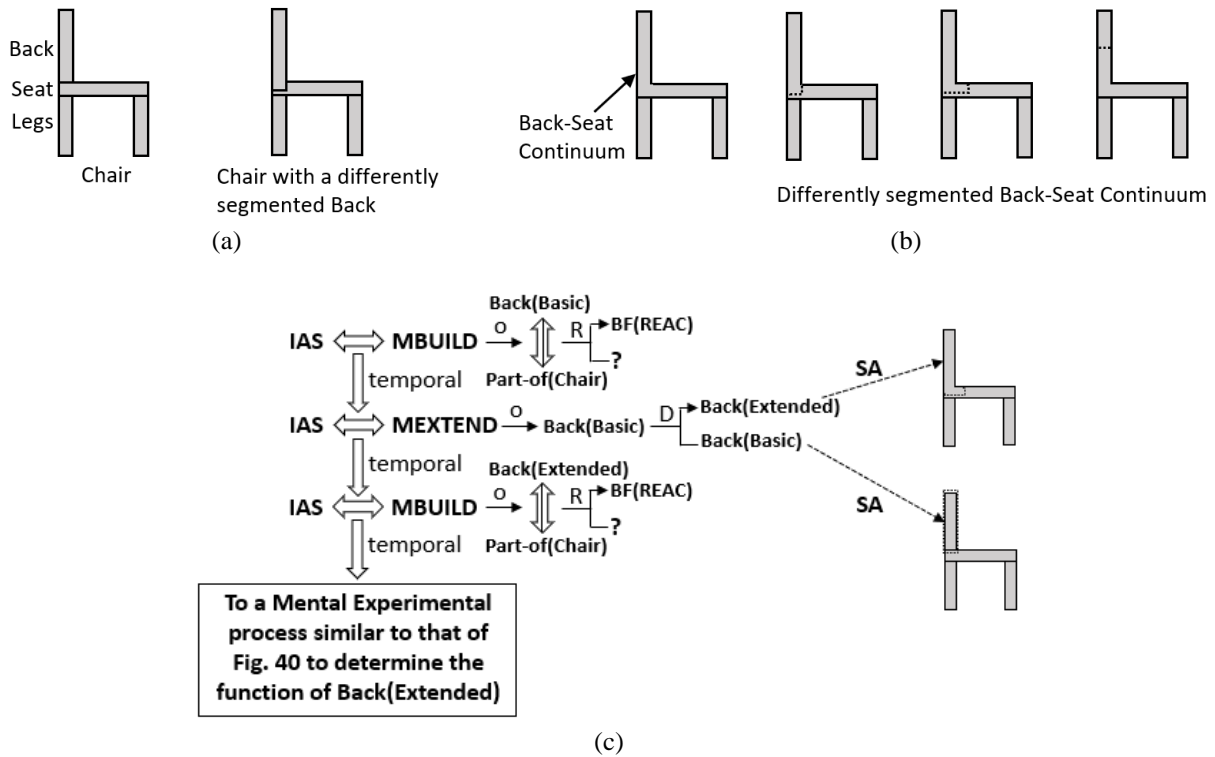

**Fig. 43.** (a) A "basic" Chair and Chair with a differently segmented Back. (b) Differently segmented Back-Seat Continuum in imagination. (c) Mental Experiment to determine the function of an extended Back – Back(Extended).

However, notwithstanding the visual division, one can consider that Back and Seat are a continuous piece of material and *imagine* carving out different regions of the entire continuum to be considered for the "Torso support function when it leans backward." A few different ways of segmenting the Back-Seat assembly are shown in Fig. 43(b). These segmented-out Backs function to support the leaned-back Torso in the same way as the visually segmented Back of Fig. 40(c). We term this *functional segmentation*. This concept is applicable to Seat and Legs as well.

Fig. 43(c) illustrates the Mental Experiments involved in the above functional segmentation. First, beginning from perhaps a visually segmented Back(Basic) (the earlier one identified in Fig. 40(c)), the IAS involved MBUILDs a concept which states that Back(Basic) is a part of Chair ("Part-of" is listed in Appendix A as a ground level concept). Then, the IAS MEXTENDs ("mentally extends") Back(Basic) to include more material along the Back-Seat continuum, MBUILDs a Back(Extended), and activates the functional reasoning process (Fig. 41) to determine its "function to support a leaned-back Torso" as in Fig. 40(c). MEXTEND is a *mental* extension process carried out in the REAC module that parallels the *physical* extension and other transformation processes normally obtain in physical objects as described in Appendix B. The MEXTEND process is elaborated by a CD+E, the details of which are not shown in Fig. 43(c) but they are similar to the details of the CD+E shown in Appendix B in connection with the transformation of some physical shapes.

We assume that there are other processes directing the MEXTEND carried out in Fig. 43(c) – i.e., how much to extend and in what way – perhaps in response to the query from other problem-solving and reasoning processes, and perhaps also in connection with consideration on some functional segmentation issues.

What is possible with the CD+ representational scheme is the representation of the *narrative* of the mental considerations involved. In fact, all the CD+ representations that we have put forth to represent functional concepts so far are "mini" causal narrative of the various functional concepts involved. The Restaurant Script shown in Fig. 15 is certainly a more lengthy, complex, and involved narrative. For the consideration on the support function of Back, the causal narrative is yet another instance of CD+ representations we have been using for the various functional concepts so far, except that it is used for the representation of mental considerations.

If the consideration of the Torso-support function of Back of Chair is continued further, the continued extension of the vertical portion (originally known as *Back*) into the horizontal portion (originally known as *Seat*) can mean that perhaps part of the *Seat* region can be functionally segmented as serving the "Back function" – namely the function of supporting a leaned-back Torso (the last picture of Fig. 43(b) shows some inclusion of the *Seat* region as part of the region to be considered for Back function). In fact, using the Mental Experiment method described above, Torso of Person in the simulation could fall backward too when the entire Back-and-Seat part is removed, and the entire Person may also fall to GROUND, without the support of *Seat*. Should the system then identify the entire visually distinct *Seat* region as functionally serving the Back function as well together with the visually distinct *Back* region? The answer to this question is that perhaps if the *Seat* region does not serve other functions better, it might be considered to be serving the Back function as well. In this case, however, because *Seat* is serving primarily to prevent the person from falling vertically to the floor, as the simulation would reveal that such is the case if *Seat* is removed, the system should probably not over extend the vertical *Back* region into the horizontal *Seat* region and include too much of the horizontal region as serving the Back function. We could perhaps define a functional segmentation rule as such: "If a part of an object is indispensable for serving a certain function, even if it could also participate in serving another function, we should identify its primary function as the function it is indispensable in serving."

The foregoing discussion constitutes the reasoning process involved in functional segmentation and the detailed reasoning process is representable by the CD+ representational scheme. Fundamentally, the CD+ representation, as well as the CD representation as originally conceived by Schank [39] are concocted to represent any language-like construct at the deep and grounded level, which is the level at which functional reasoning operates.

### 7.5 The Function and Structure Causals of a Piercing Tool or Spear

In this section and the following section, we consider the function and functioning of two tools – one, a very primitive tool used by early humans – a piercing tool usually fashioned from stones, and another, a very advanced tool constructed by humans in the modern age – a jet engine.

Fig. 44(a) illustrates a simple stone tool used for piecing of objects or skins of living things. Often, tools like this are also used for shaping of other materials by chipping off on their surfaces, but if the end of the tool is sharp, the primary intent would be to use it for piercing. The PierceTool can be thought off as consisting of two parts – one part (the Blunt end) affords Grasping - GraspPart, and the other one (the Sharp end) affords Piercing – PiercePart. The purpose of GraspPart is for the human hand to handle the tool appropriately to direct PiercePart to create the intended indentation – the associated action is Stab, which is PTRANSing of PierceTool into Target-Object Rapidly and with Force (both Rapid and Force are listed in Appendix A as near ground level concepts). On the pierced Target-Object illustrated, the final indentation created is shown, more or less mirroring the shape of PiercePart.

An important point about the two portions of the tool (GraspPart and PiercePart) is that from the point of view of visual segmentation, there may or may not be a distinct boundary between the two portions. (Contrast this with a similar discussion in Section 7.4 with regards to Chair. In the illustration in Fig. 44(a), there may appear to be a boundary, but when a tool such as this is fashioned from a stone, that boundary does not usually exist.) There is typically a gradual transition between the two portions. From the point of view of functional segmentation, the same gradual transition happens – the human hand may grasp further away or further into the piercing portion, depending on how firm a grasp the person is trying to exert. Similarly, the piercing depth and effect intended and created will determine what is characterized functionally as the Piercing Portion. We therefore show that there is an overlap between the two portions in Fig. 44(a), with a double arrow pointing to parts that more affords Grasping or more affords Piercing, with a gradual transition between them.

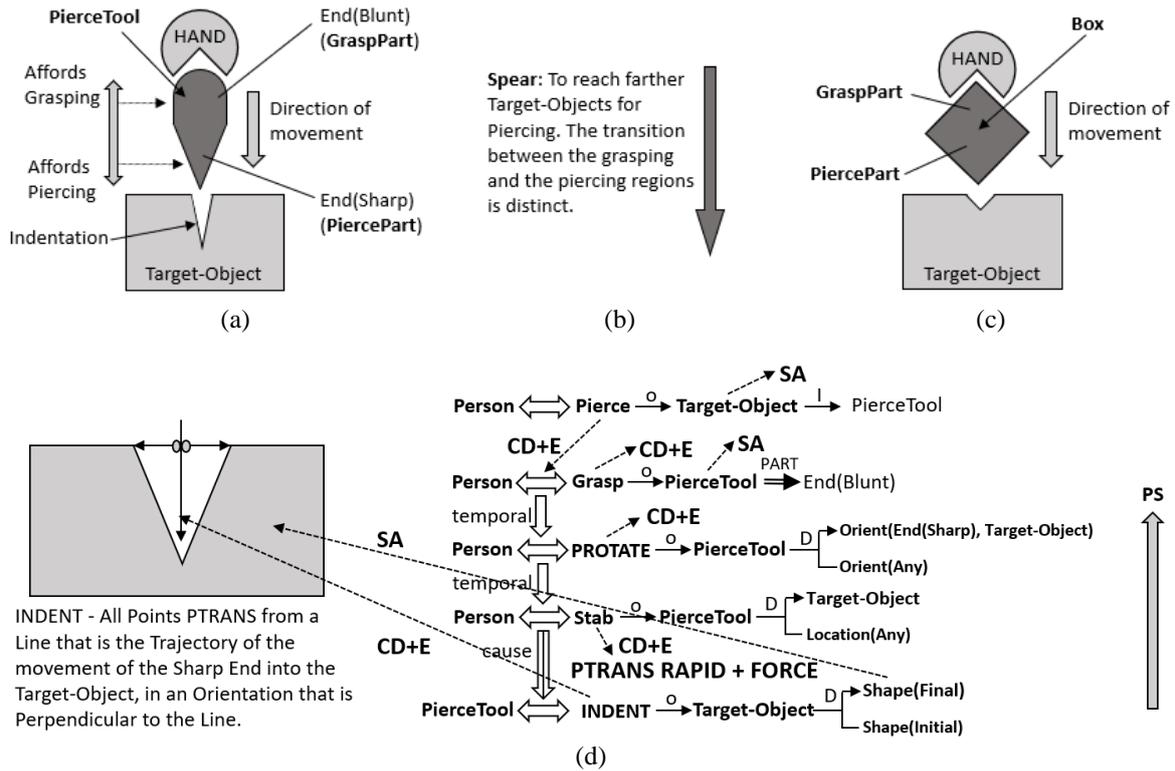

**Fig. 44.** (a) A PierceTool used to create an indentation on a Target-Object. (b) A Spear which is a PierceTool with an extended GraspPart to allow the user to reach farther. (c) Using a Box as a PierceTool. (d) The CD+ representation of the functioning of the PierceTool. The portion of the CD+ representation from the Grasping action to the creation of the indentation can be thought of as the CD+E for the action Pierce – Person uses the INSTRUMENT (I) PierceTool to Pierce a Target-Object. PS = Problem-Solving.

Fig. 44(b) shows a related tool, a Spear, which has a long handle instead of just a rounded blunt portion for grasping. Visually, the GraspPart and PiercePart are more distinctly separated in this case. The long handle allows the user to extend her reach for the piercing function.

In Fig. 44(c) we illustrate how a box could also be used as a piercing tool, albeit not a very good one. One corner of the box could be used for grasping, though the grasp may not be very good, and could be hurtful to the hand. Another corner could possibly fulfil some degree of piercing function, but perhaps not as effectively as what the specialized PierceTool of Fig. 44(a) could do.

Fig. 44(d) shows the CD+ representation of the functioning of PierceTool. The process begins with Grasping the tool at the Blunt end, Orienting it appropriately, and PTRANSing it Rapidly and Forcefully (Stab) toward Target-Object, creating an indentation. On the right side of Fig. 44(d) we indicate with an arrow showing how a Problem-Solving process (PS) would employ the CD+ representation "backward" – if there is a need to create an indentation, the preceding actions could be taken.

The entire CD+ representation from the Grasping action to the creation of the indentation can be thought of as the CD+E for the action Pierce – Person uses the INSTRUMENT (I) PierceTool to Pierce Target-Object – represented as the first CD+ statement in Fig. 44(d).

The description of the indentation is stated in English in Fig. 44(d) for ease of illustration but it can likewise be stated in CD+. The SA shown is specific, arising from the specific form of PiercePart of PierceTool, but the corresponding English description has some generality, covering a wide range of description of indentations arising from various forms of piercing end of the tool involved.

In Fig. 44, we have used the concepts of Blunt and Sharp to define the two ends of PierceTool, in order to state the correct orientation to adopt for the tool to fulfil its function. In our framework, Blunt and Sharp can be defined visually as shapes that have certain properties, which can in turn be defined in terms of the trajectories of the points along the boundaries, a point being the most elemental, grounded visual concept (Appendix A). However, Blunt and Sharp can also be defined functionally – if the Blunt end of PierceTool is used to impact on Target-Object (or for that matter, any object), it will not make much of an indentation, but if something that is Sharp is used to impact on Target-Object, it will make an indent as stipulated in Fig. 44(d). These are the functional definitions of Blunt and Sharp.

The conceptual representation of the functioning of PierceTool in Fig. 44(d) possesses the RUUI property as defined in Section 2. The representation specifies the conditions to *identify/recognize* (R) an instance of a piercing action, and it provides for the explanation of how the piercing action is achieved because all causal-temporal steps are explicitly stated, reflecting the *understanding* (U) of the functionality involved. It certainly prescribes how to achieve a piercing action through a backward chained Problem-Solving (PS) process, hence the *use* (U) of the functional concept of piercing. It could also certainly facilitate the *invention* (I) of new instances of piecing function – i.e., the piecing action can be applied in other situations and use scenarios, even though, as mentioned in Section 2, we relegate the further exploration of this aspect to future work.

Fig. 45 illustrates the structure causals of the two parts of PierceTool. In Fig. 45(a), the CD+ expression states that the GraspPart of the tool *enables* the grasping action for the Person involved, which in turn *enables* the orientating of the tool in the appropriate direction (pointing of End(Sharp) of the tool at Target-Object). At the same time, the grasping *enables* two kinds of PTRANS. One kind, which is a Rapid and Forceful one, in turn *causes* PierceTool to create an indentation on Target-Object. Another kind, which is a "normal" one, supports the transportation of PierceTool from one location to another.

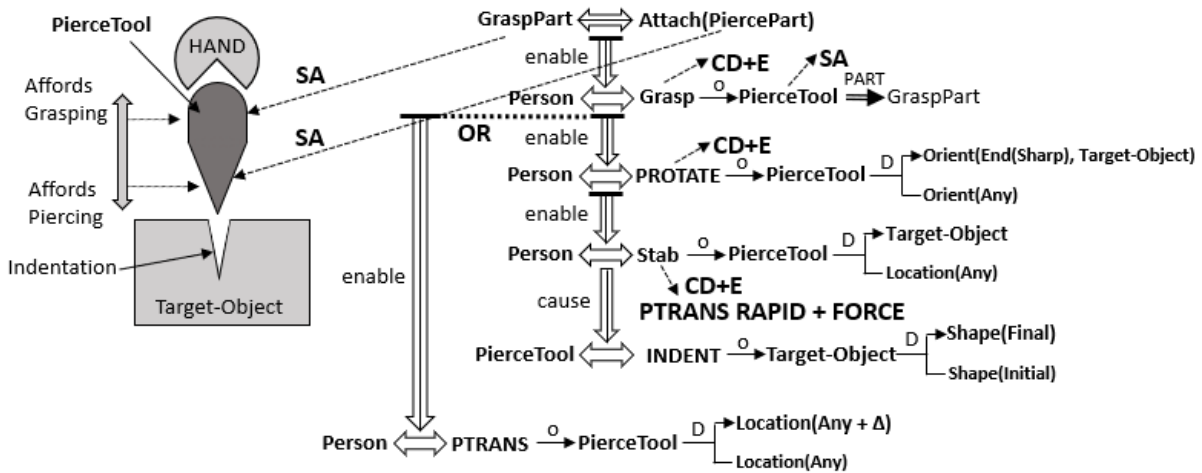

(a)

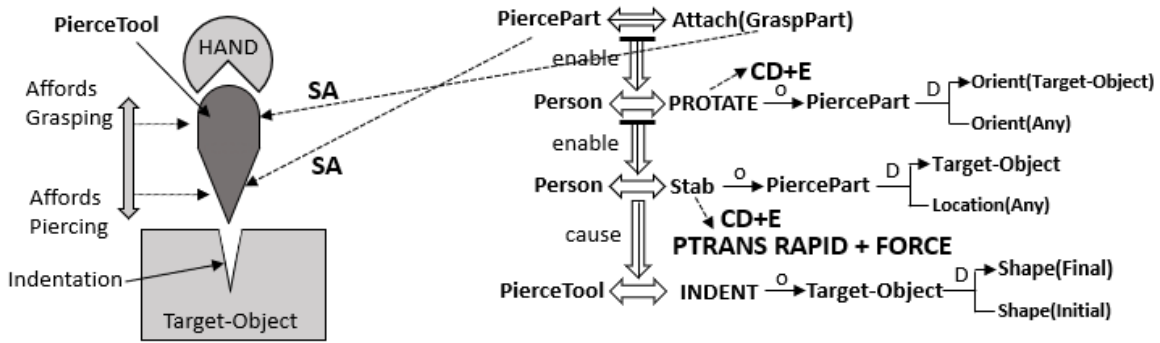

(b)

**Fig. 45.** (a) The structure causals of GraspPart of PierceTool. (b) The structure causals of PiercePart of PierceTool.

In Fig. 45(b), the CD+ expression states that PiercePart of PierceTool, being in a state in which it is Attached to the GraspPart, *enables* it being oriented in an appropriate direction (pointing of End(Sharp) of the tool at Target-Object), which in turn *enables* the entire PiercePart to be PTRANS Rapidly and Forcefully toward Target-Object from any location, which in turns *causes* an indentation in Target-Object.

Similar to what we have shown in Fig 40, the corresponding C-FACT versions of these structure causals could also be stated in CD+ representations.

### 7.6 The Function, Functioning, and Structure Causals of a Jet Engine

In this section, we will apply the CD+ representational framework to represent the function and functioning of a jet engine (JetEngine).

The function of JetEngine at the highest level can be characterized as a machine that is able to PTRANS itself forward by ejecting air, much like a human is able to PTRANS herself by walking. This is the topmost level characterization of JetEngine and it is shown in Fig. 46(a). The CD+ expression basically states that JetEngine PTRANs itself in a direction that is toward the Front part of the engine, and the means of doing so is by Ejecting Air in a direction that is Opposite to the direction of the PTRANS (Front and Opposite are listed in Appendix A as near ground level objects). Ejection of Air, in turn, is defined as a Rapid and Forceful PTRANS of air out of the JetEngine, in this case out through the Exhaust region of the engine at the back of the engine. The CD+E of Eject given is specific to JetEngine, but it is a general concept applicable to movement with respect to any object.

The interesting part of the representation of Fig. 46(a) is the concept of Air. Since Air does not have a visible presence and is amorphous, it is strictly not possible to have an SA in the form that we have been using associated with Air. However, Air can be characterized functionally, using precisely the same representational devices that we have been using – CD+. Therefore, instead of an SA, Air can be elaborated with a CD+E. We will not delve into the details of the CD+E but instead describe in English the properties and functionalities of Air in the right bottom corner of Fig. 46(a). (The representation of CAN in Fig. 21(a) will be useful in fleshing out part of the CD+ representation of Air.) The detailed exposition of this is relegated to future work, but suffice it to say that this is consonant with the characterization of objects in general with CD+E, as we have been doing all along.

The regions like Front or Back of an object are defined in [21], and listed in Appendix as near ground level concepts. In the case of JetEngine, these also take on specific names like Intake and Exhaust respectively, as shown in Fig 46(a). Intake and Exhaust of course have functional connotations, which can be seen in the detailed representation of the functioning of JetEngine in Fig. 46(b). (The concepts of Front and Back are not merely visual – they are usually tied in with the potential activity of the object involved. E.g., the fact that the labels exist for a certain object implies that the object at some point may move, and the part of it that corresponds to its usual direction of movement is Front and the opposite of that is Back. In the case of a house, say, there is no movement of the house involved, but Front is where visitors or strangers are expected to approach or enter. Of course, it is often possible to distinguish Front and Back based on visual or physical characteristics because these visual/physical features are designed to serve the associated functionalities.)

Having defined the overall function of JetEngine at the top level, to further understand the *functioning* of a jet engine, it would be instructive to drill down one level to look at the basic functional components of JetEngine and see how they function together to produce the ejected air, or what is also known as "thrust" to PTRANS the engine. In our parlance, this would be looking at the structure causals of the parts of JetEngine.

In Fig. 46(a), three major parts of JetEngine are identified – Compressor, Combustion Chamber, and Turbine. There are also two minor parts – Intake and Exhaust. There is no overlap shown between the different parts partly because a jet engine is a precisely engineered human artifact, and the parts are often distinctly defined, and partly because the functions of these parts, in the form that we will be defining them, do not overlap, unlike say in the case of PierceTool (Fig. 44(a)). This SA of the engine is shown as a 2D cross section. Ideally, the SA should be a full 3D analogical model. However, sometimes 2D cross sections can illustrate the machine's internal structures and their attendant operations better, and that is why many a technical document illustrates with cross sections as aids to the explanation of the operational principles of various machines involved. Hence we adopt a 2D cross-sectional version of the SA for ease of illustration.

The complex looking CD+ representation of Fig. 46(b) basically describes the following process. First, Air ENTERs Compressor, which *causes* it to be compressed (Volume reduced and Pressure increased). (Compressor is normally started by a separate mechanism – a "starter motor" – but we ignore that here.) Then, the compressed Air, having entered Compressor, also *enables* Compressor to PTRANS it into Combustion Chamber. This *enables* Combustion Chamber to heat up Air, through the burning of fuel, which in turn *causes* the Air's Volume and Temperature to increase (the Pressure may or may not increase, depending on whether or how quickly the expanded Air could escape to another part of the engine). Partly due to the continued momentum imparted on Air by Compressor, partly due to its Volume expansion, the heated Air gets PTRANSed into Turbine, *causing* Turbine-Blades to turn faster. This in turn *causes* Compressor-Blades to turn faster through Shaft connecting Turbine-Blades and Compressor-Blades, and at the same time it *causes* Air to EXIT Exhaust region Rapidly and Forcefully. Both Turbine and Compressor-Blades may not turn faster at steady state, and simply sustain they turning speed. Finally, it is this Ejected Air that *causes* JetEngine to PTRANS forward, in the direction of the Front of the engine.

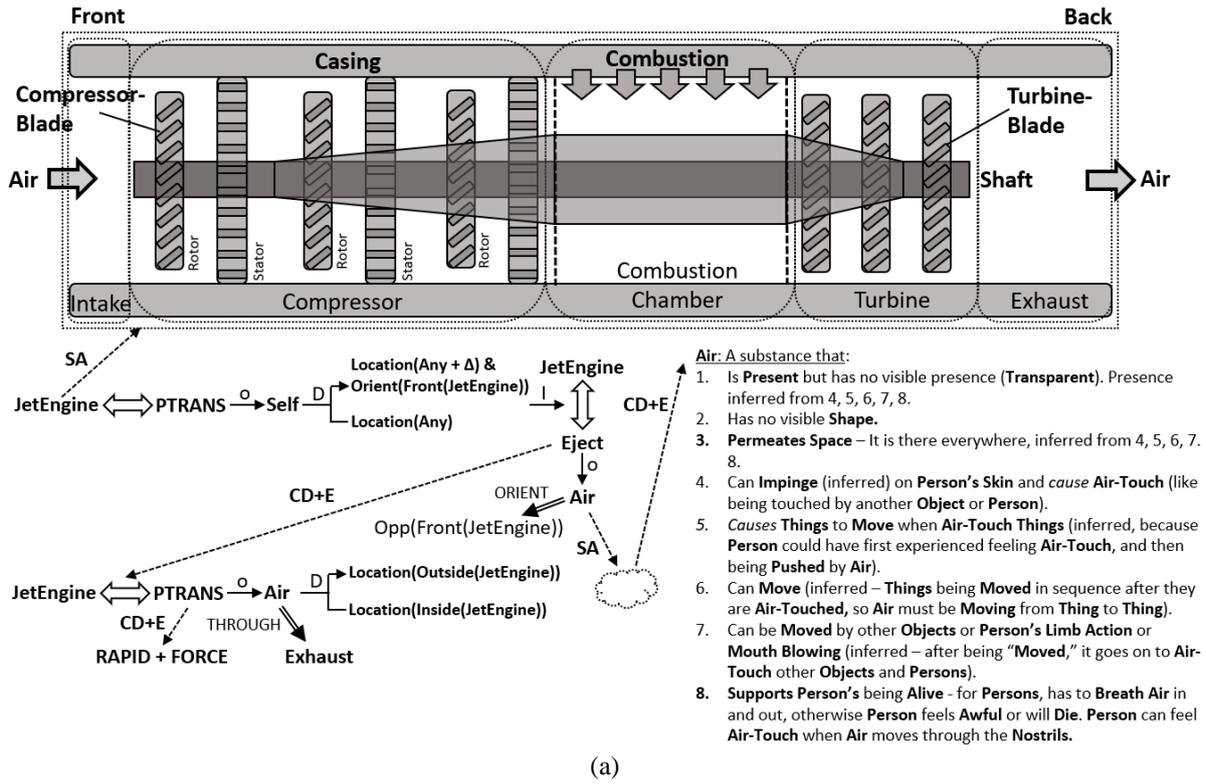

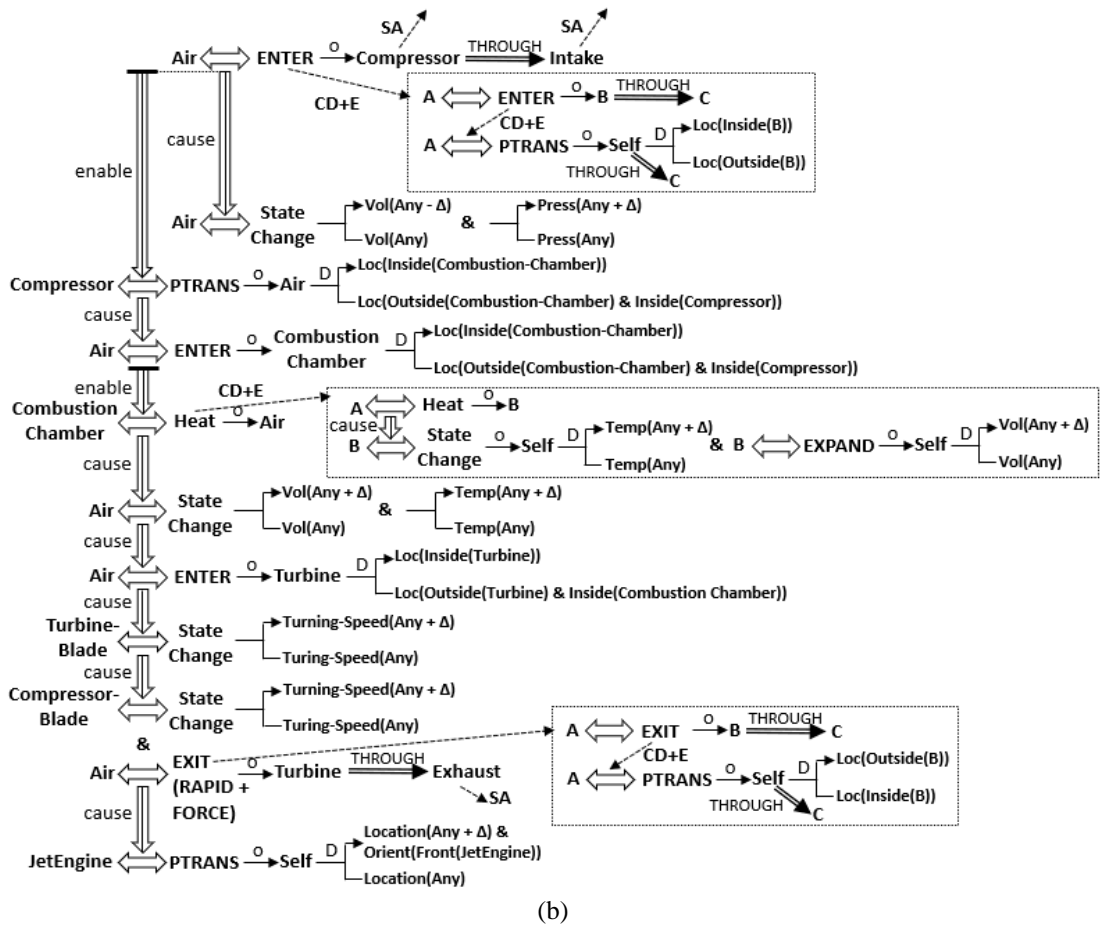

**Fig. 46.** (a) SA of JetEngine and its top level function. (b) The next level detailed functioning and structure causals of JetEngine. Some general CD+ expressions are shown, for concepts ENTER, EXIT, and Heat. Vol = Volume, Temp = Temperature, Press = Pressure.

The representation of Fig. 46(b) supports the reasoning about the functioning or malfunctioning of JetEngine. For example, if a system equipped with the representation is asked the following question: "What happens if Air, after passing through Compressor, is blocked from Entering Combustion Chamber?" Following the causal links, the system would answer that there would ultimately be no Ejected Air and the engine would not be able to PTRANS itself forward. It would also be able to explain *why* that is the case by tracing and explaining the step-by-step processes that is now not taking place instead because no Air is now Entering Combustion Chamber.

We can of course continue to drill down to further details of the functioning of JetEngine, such as going down to the level of describing how air molecules striking the Compressor -Blades or Turbine-Blades behave, plus all the other detailed operations. The real JetEngine is a highly complex machine. The simplified 2D JetEngine SA model of Fig. 46(a) only captures the highest-level essential constructs, and the CD+ descriptions in Fig. 46 only capture some, though the critical, aspects of the functioning of JetEngine. There are also other aspects of JetEngine in the simplified model of Fig. 46(a) that is not described here in CD+, and needless to say, JetEngine's SA can go on to be further elaborated down to the last screw. Suffice it though to say here that CD+ representations have the power to elaborate and describe these detailed functionalities, level by level, down to the smallest aspects of the operation of JetEngine.

## 8. A Unified and General Autonomous and Linguistic Reasoning System (UGALRS)

Since the CD framework was born of natural language understanding, the CD+ framework and the attendant architecture articulated in Fig. 29 is not merely suited for the representation and processing of functional concepts, but they should also be able to support natural language understanding and generation, if suitably extended. Fig. 47 shows how a language processing module can be added to the architectural framework of Fig. 29 to form a complete "Unified and General Autonomous and Language Reasoning System" (UGALRS).

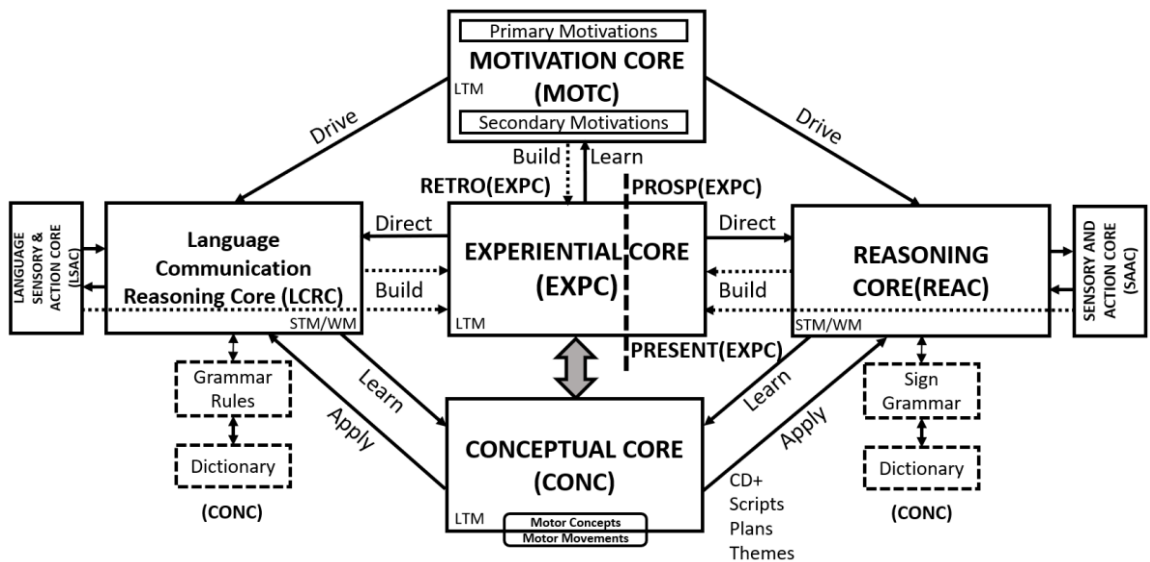

**Fig. 47.** Unified and General Autonomous and Language Reasoning System (UGALRS) for an IAS. Modules in dash-lined boxes are part of CONC. See Fig. 21(b) for the inside details of REAC.

There are things unique to language processing separate from the kind of constructs usually needed for other kinds of perception (visual, auditory, etc.) and action (the SAAC module). Language *understanding* is a kind of "perception," an input to the system involved, akin to scene understanding in the visual domain. Language *generation* is a kind of "action emission," an output from the system, akin to what we usually expect a human or robot to do with their limbs. For one, language communication relies on some Grammar Rules, and this would be stored in CONC, even though shown outside it in Fig. 47 for clarity of illustration. There is also a need for a Dictionary of sort that contains the definitions of the lexical items used, that is also part of CONC. The equivalent of these modules over on SAAC side would be the modules supporting sign language communication. Within the Language Communication and Reasoning Core (LCRC), there are the equivalents of those submodules in REAC, such as PS (Problem-Solving - e.g., given that I would like to express this content to my listener, constrained by my motivational and affective signals from MOTC, do I want to agree with him or retort him? How do I concoct an appropriate sentence?) and SM (Simulation - e.g., if I were to say this, what would the listener likely to say in return based on the known communication rules?). Understanding, say, of a sentence spoken by an utterer would correspond to *scene understanding* on SAAC side, in which signals from the Perception portion of SAAC gets interpreted by CONC, in the context of the current state in EXPC.

If there is a need to communicate in natural language about functionalities, such as explaining the intricate causalities and other aspects of the CD+ representations to a human enquirer, or receive input from a human to modify the functional or other contents of knowledge in CONC or elsewhere in UGALRS, LCRC will kick into operation. Of course, there is a host of other things that UGALRS could communicate about, and that could form the basis of learning through the use of natural language (symbolic learning) which can vastly speed up the acquisition of knowledge for an IAS. And needless to say, learning through language is also what sets humans apart from the other animals that has enabled them to dominate the world. Of course, this is strongly contingent upon the fact that true language understanding is achieved through firmly grounded CD+ representations operating within the UGALRS framework.

In any case, to achieve maximal generality, a framework addressing the characterization and representation of functionality must involve the various modules of processing in something like UGALRS. For example, we have demonstrated that issues on motivation such as embodied in MOTC, often treated in AI research as separate issues related to some narrow purposes, are also involved in the characterization of even the most basic of functionalities (e.g., Fig. 28, 33, and 36) and contribute a critical part to the over intelligent functioning of an IAS [7]–[10]. As pointed out and explored in [10], Artificial General Intelligence is built upon an all-encompassing architecture such as UGALRS because in general situations and general tasks, many components within an over-arching architecture such as UGALRS are involved and engaged. Such an idea of engagement of many components is also echoed in Embodied Artificial Intelligence, along with the idea of grounding [105].

## 9. Conclusions and Future Work

As mentioned in Section 1, there was a position paper published by the UCLA group (Center for Vision, Cognition, Learning and Autonomy – http://vcla.ucla.edu) recently in 2020 that called for a paradigm shift in AI, and it lays out very clearly what lies ahead in the fundamental challenges facing AI [14]. It is not merely the classification of objects (though that is a part of it), but the task-based, functional understanding of objects that constitutes the vast majority of the critical issues that will endow an artificial system with intelligence. The five key issues identified are: Functionality, Physics, Intent, Causality, and Utility (FPICU).

As can be seen, the function representational framework promulgated in this paper addresses all of the FPICU issues, setting up a framework to embed all of them inside, and exploring in detail some specific examples. Among the five items in FPICU, Functionality is the key aspect that ties everything together. As we have demonstrated, the knowledge of Physics supports the determination of whether something can serve certain functionality. Causality lies at the center of the characterization of functionality, as reflected in numerous CD+ representations. The Intent aspect is the "why" aspect of functionality – an object's functional aspects are to serve certain deep motivations and needs, and how well that is achieved is Utility. Therefore, Intent and Utility are integral parts of the characterization of functionality.

Unlike classification, which is basically an input (pattern) – output (category) task, the representation and understanding of functionality involves a descriptive representation language – describing the functionality involved, and achieving/providing true understanding so that the representation is actionable by, say, a robotic system. We have developed a CD+ representational framework that is general enough to capture, as we have demonstrated, a wide range of functionality.

As has been described and demonstrated, the generality of CD+ is derived from the generality of the base CD framework, enhanced with two key representational devices - SA and CD+E – together with a defined set of grounded level and near ground level concepts.

As was mentioned in Sections 1 and 2, though there have been sporadic efforts devoted to the computational study of functionality, including the author's own early work, there has not been any attempt to devise a unified and general framework to encompass the functionality of most if not all objects and constructs. We believe the CD+ representational framework has addressed the shortcomings of various earlier works, and has the capability of representing the functionalities of not only the more tangible kinds of objects and constructs, but also the more ephemeral and amorphous kinds of substances and phenomena such as air, liquid, electricity, and magnetism (for air, we have described how it can be done in Fig. 46), or for that matter, functionalities at the molecular, atomic, and subatomic levels of operations associated with various constructs at those level, whether they are classical or quantum mechanisms. The key to the power of CD+ is that it has identified a general set of primary, or "atomic" conceptual constructs from which all other conceptual constructs could be built, following the footsteps of and extending on the spirit of identifying atomic conceptual constructs promulgated in [10], [103].

Future work would extend the current work in a number of key directions. First, as many of the objects used in this paper for demonstrating the idea of applying CD+ to characterize functionality are relatively simple toy objects (even the JetEngine example is just a high-level skeletal depiction of a real engine), real world objects with full visual and physical constructional details should be used. Second, for many of the objects that we have discussed and with which we have demonstrated the application of CD+, we have not always drilled down all the CD+Es to their ground level details, even though it is quite obvious that the CD+ framework is capable of characterizing these details. Also, even though the CD+ representations are quite obviously capturing the core essence of the functionality involved, we have not quite demonstrated in detail, except in some simple cases (Section 6), the full power of these function representations in problem-solving and robotic actions, reflecting a kind of "true understanding" of the functionalities involved. These are certainly directions of future explorations.

The focus of this paper is in the representational constructs needed for functionality, and in a number of places we have demonstrated how some reasoning processes can work with the representation to make inferences for certain tasks (such as determining if a certain object can serve the function of Chair – Section 6.3), but we have not presented a more complete and thorough treatment on the reasoning processes working in conjunction with the representations. This should be the one of the primary focuses of future work.

As mentioned in Section 2, *invention* is one area of the RUUI criteria which is not touched on in this paper. In principle, if an AI system has a good and full representation and understanding of the functionalities of various objects and physical constructs, this representation and understanding should be able to greatly reduce the search space needed to concoct/invent new objects and constructions to serve certain purposes. It should obviate largely the need to do a relatively blind and exhaustive kind of search to find new solutions which typically consumes large amounts of computational resources, in addition to the fact that blind search methods are definitely not deemed "intelligent."

One other key area for future work is the learning issues of CD+ representations. As mentioned earlier in Section 5 and as historical lessons would strongly suggest, any representational scheme that is not amenable to learning is bound for failure, no matter how powerful it seems to be in capturing complex concepts. As was discussed in Section 5, there exists already some initial work in the learning of causality and CST graphs that can be further developed toward the learning of complex CD+ representations and scripts. This will be an important direction of future work.

The details of the representations and operations of MOTC certainly constitute an important aspect of the complete characterization of functionality, because of its contribution to the Intent and Utility aspects of FPICU. Human motivations are complex, and one of the most recent psychological studies identifies up to 161 basic human drives [69]. Unlike other computational studies of motivation which usually provides characterizations of relatively simple motivations, such as the hunger drive, avoidance of danger, etc. [7]–[9], the devices of CD+ could be used to represent relatively complex M-CONC (Motivational Concepts – Fig. 29), and can potentially handle and encode the 161 complex human motivations identified by psychologists [69]. This will be an important component to develop in future work.

We have mentioned, in connection with the discussions associated with Fig. 26, that future work will investigate representations for more general conditions associated with objects participating in functional relationships such as Support. Also, in connection with the discussions associated with Figs. 19 and 29, the Motor Concepts and Motor Movements modules have only been cursorily dealt with in this paper. These modules are important for directing a robot to execute the necessary actions as a result of complex reasoning processes in the IAS, and they also form an integral part of the characterization of the functionalities involved. Therefore, future work should also focus on investigating the issues involved further along this direction. In connection with the discussions associated with Figs. 31 and 32, it is clear that the knowledge of qualitative and quantitative physics constitutes an integral part of functional reasoning, as this knowledge works hand-in-hand with FHR (Fig. 31) to allow the IAS involved to determine the Utility of any object under consideration for its satisfaction of certain functionality.

An interesting feature of our CD+ representational framework is its provision for the representation of mental processes, say in the form of Mental Experiments, as discussed in connection with Fig. 41. This allows the mental processes to not only be transparent and explainable, but also amenable to learning. Hence, the meta-level reasoning processes themselves are adaptive and can be adjusted as a result of experience. So, for example, if a certain manner of carrying out Mental Experiments is found to be ineffective or defective, it can be changed and improved – i.e., the particular steps involved in a Mental Experiment such as that shown in Fig. 41 is subject to modification. Other than Mental Experiments, all other kinds of reasoning processes in an IAS (such as, say, analogical reasoning process, reasoning processes arising from interpreting sensory input such as vision, etc.) are also representable in CD+ and subject to learning. This allows the IAS involved to acquire another layer of adaptability. And, as discussed in connection with Fig. 37, there are much finer mental operations than have been discussed in this paper (i.e., MTRANS, MBUILD, MEXTEND, etc.) that are needed for characterizing and representing much more complex internal mental operations at a finer level that takes place within an IAS. This will be a worthwhile direction for future work.

Lastly, the UGALRS architecture of Fig. 47 is general and all-encompassing, and will be further refined and applied to a vast array of intelligent tasks a typical IAS is challenged with.

Unlike the current main thrust of AI research which is centered around learning, namely the various techniques of machine learning, the current paper focuses instead on representational issues related to functionality. This is because we believe it is of paramount importance to firstly understand what representations are needed for functionality, then the representations become the targets of the learning mechanisms to be investigated. As has been demonstrated in this paper, the constructs required for the representation of functionality are non-trivial and intricate, especially when what we are after is a general framework for handling functionality in various domains and applications. As has been mentioned in the foregoing discussions, the representational framework we propose is amenable to learning, and one major focus of future work will be learning, along with the enhancements of the various aspects of the framework as discussed above.

As mentioned, the characterization and representation of function and affordance is an important part of an intelligent system or an IAS's operations, hence addressing it adequately and in a general manner is an important step toward Artificial General Intelligence. This paper has attempted to take just such a step in this direction.

**Appendix A: Ground Level Concepts**

Ground level concepts are something that cannot be learned and that are built into an IAS (artificial or natural such as a human). These concepts are also primitive in the sense that they cannot be defined in terms of other concepts. Dictionaries generally tries to define all concepts in the human lexicon, but when it comes to ground level concepts, they are always circular – e.g., the Merriam-Webster (www.m-w.com) defines "location" in terms of "position," and "position" in terms of "location." However, higher/non-ground level concepts need not be circular as they can be defined in terms of ground level concepts. E.g., for a concept such as "move," dictionaries tend to define it in a circular fashion, defining "move" in terms of "go," and "go" in terms of "move" (www.m-w.com), but in our framework "move" is defined in terms of ground level concepts of "location" and "change". See [10], [20], [21], [99], [103] for discussions on ground level concepts and symbolic circularity.

Consider the example of location. Suppose a visual system does not exist for an IAS, or it exists but does not return the locations of objects in the environment, and suppose the IAS would like to learn that a certain location is associated with, say, food [10], [106], and it wishes to use this information to return to the same location time and again to consume the food. Using this location information and having learned the association or causality between food and a certain location, the IAS can save a lot of time in looking for food, improving its chances of survival [10], [106]. Suppose location is not a piece of information available to the IAS, the IAS then has to blindly search the environment until it chances upon food, and it has to do this repeatedly. Of course, food may not always be available at a certain location, but in the event that it does, the location information is of paramount importance to guide the IAS to the food. Therefore, to aid the IAS in survival, nature has evolved built-in location information in natural IASs (e.g., the brain structure hippocampus has been established to be the neural structure that encodes spatial locations [107]). Roboticists have also typically built in the ability to detect and represent the location information in the robot or artificial IASs. Therefore, unlike the correlational or causal information between location and say, food, which can be learned, location itself has to be built in first for this learning to be possible.

Likewise, there is a host of parameters that have to be built in that cannot be learned. These are all known as *ground level concepts*. Concepts such as the various colors and sensations such as saltiness, sweetness, etc. are also ground level concepts. Scientifically, we can define colors in terms of electromagnetic wavelengths and tastes in terms of the structures of the molecules associated with them, but an IAS, natural or artificial, does not need to rely on these to function properly on basic tasks that require the ability to identify or distinguish different colors or tastes. The deeper scientific understanding can be learned later for other purposes.

That is, in an IAS, natural or artificial, there is definitely the need to have built-in processes or procedures to supply these ground level concepts such as locations, colors, tastes, etc. to the system, but there is no need to *explicitly* define these processes or procedures, in scientific terms or otherwise, such as in the form of CD+, nor make available their attendant definitions to higher reasoning modules of the IAS. The IAS can basically begin to build its understanding of the world and carry out its general learning processes based on these supplied parameters.

Sometimes, near ground level concepts are built-in as well to the IAS to increase the efficiency of intelligent operations. One example is the *distance* between objects or points which is usually made available by a visual system to an IAS. But if the IAS has encoded *location* and has the ability to compute the *difference* of the locations between two objects or points, in principle it can obtain the *distance* between them. The concept of *difference* can be applied to all kinds of parameters including *location,* and is a ground level concept itself.

Therefore, we list the following as ground level or near ground level concepts (non-exhaustive):

GROUND LEVEL CONCEPTS

- **Time**, **Present**
- **Point** (physical or conceptual)
- **Location** (of Object or Point in space)
- **Orientation** (of Object), Direction, Face (for human body, as it the direction in which the face faces)
- **Volume** (of Object)
- **CM** (Center of Mass) of Object (certain Point on Object)
- **Part-of Object** (a subset of the collection of Points on Object)
- Labels of different parts and dimensions of Object – **Top**, **Bottom**, **Side**, **Length**, **Width**, etc. (see Fig. 24)

- Spatial Relations – **At**, **Next-to**, **Near**, **Far**, **Above**, **Below**, **On**, **Parallel**, **Align**, **Inside**, **Outside**, **Beyond**, etc. (see [21], [99])
- **Gravity** – tendency for Objects to fall toward ground
- Locations, directions, and measurements in a gravitational field – **Ground**, **Sky**, **Up**, **Down**, **Horizontal**, **Vertical**, **Height**, etc. (see Fig. 24)
- Various fundamental needs and motivations – e.g., **Food** (natural IAS) and **Energy Supply** (artificial IAS), **Safety**, **Companionship**, **Competence**, **Beauty Appreciation**, **Neatness**, … (see [7], [8], [10], [20], [69], [71], [102], [104])
- **Pleased** – satisfaction of fundamental needs and motivations.
- Various emotion types and values – **Happy**, **Sad**, **Anger**, etc (see [7], [8], [10], [20], [72])
- Various visual, auditory, gustatory, and somatosensory sensations: **Colors**, **Tones**, **Tastes** (**Sweetness**, **Saltiness**, etc.), **Temperature**, **Pain**, **Itch**, … (see [10], [20], [102])
- **State** – Any of the above parameters such as Location, Volume, Happy, Sad, Temperature, Pain, etc. adopting certain values
- **Same**, **Different**, **Greater** (**More**), **Smaller** (**Less**)
- **Difference** - between parameter values
- **Any** – exact value irrelevant (this could actually be a near ground concept instead, defined in terms of other more fundamental concepts, but we will leave it at this level for this paper)
- **All** – inclusive of everything satisfying certain properties (this could actually be a near ground concept instead, defined in terms of other more fundamental concepts, but we will leave it at this level for this paper)
- **Appearance** (**Materialization**) & **Disappearance** (**De-Materialization**) – of material points or light (see [10], [103])
- **Contact** (between material points)
- **Look-At** (the fixation of the human eye on certain points in the visual field)

NEAR GROUND LEVEL CONCEPTS

- **After, Before** (**Greater** than or **Smaller** than **Present** along the **Time** axis)
- **Line**, **Trajectory** – collection of Points (see [10], [103])
- **Object** (certain collection of Points)
- **Boundary** of Object (certain subset of Points on Object)
- **Bulk** of Object – refer to the "inside" of the Object, other than the Boundary (certain subset of Points on Object)
- **Distance** – difference in location (near ground level - built from ground level concepts of location and difference)
- **Distance between Locations** on Object – e.g., length of object
- **Shape** of Object – including "Blunt" or "Sharp" – which can be defined in terms of the boundary points' trajectory.
- **Long-Axis** and **Short-Axis** of Objects – computable from Points on Boundary of Object (see [21])
- **Front/Back** – see
- **Long**, **Short**, **Hard**, **Soft**, **Heavy**, **Light** (scalar quality of Objects, can be defined in terms of certain functionalities). Long and Short can also be applied to Time
- **Alive**, **Energetic**, **Comfortable**, **Weak**, **Hurt/Injured** (physical or psychological), **Sick**, **Dead** (M-CONCs that represent bodily states to be desired or avoided. E.g., a certain temperature range, or a certain body posture = Comfortable)
- **Change** (and **No-Change**) - of parameter values (defined in terms of Same or Different over Time)
- **Activity** – any changes in State
- **PTRANS, Move** – change of location – combining both ground level concepts "change" and "location"

- **Rapid/Fast, Slow** – the amount of change per unit time. See [10], [103] for representations
- **PROTATE** – physical rotation
- Dynamic Spatial Relations – **Through**, **Over**, etc. (can be defined in terms of the grounded Spatial Relations and Time – e.g., Through is Outside, then Inside, then Outside).
- **Force** – an abstract entity that can cause Movements or other Activities
- **Move-Toward**, **Move-Away**, **Orient-Toward**, **Orient-Away** (Move - defined in terms of Smaller or Greater Distance after Time Change. Orient – defined in terms of Smaller or Greater Angle after Time Change, between a line connecting the current object/agent location to a target location and the direction the object or agent is currently "facing")
- **Opposite** (Move-Toward = Opposite(Move-Away) and vice versa)
- **Attach** - defined as a state that engenders concomitant movements of two objects/parts that are attached to each other when one of them is moved (see [10]).
- **EXPANSION**, **CONTRACTION**, and **EXTENSION** of, and **EXTRUSION** and **INDENTATION** on Objects – built on movements (PTRANS) of points on object – see Appendix B
- **Cause** (see p. 405 of [10] for a CD representation of the learning of Cause)
- **UNTIL** (change of activities when certain conditions are met)
- **MBUILD** (see Section 4 and [39])
- **MTRANS** (see Section 4 and [39])
- **BPTRANS** (see Section 6.3 and Fig. 36)
- **EXTRANS** (see Section 6.2.1 and Fig. 26)
- **MEXPAND**, **MCONTRACT**, **MEXTEND**, **MEXTRUDE**, **MINDENT** – mental operational versions of the shape transformation actions in Appendix B.
- Human body and limbs near ground level actions - **Turn-body**, **Lift-Left(Right)-Thigh**, **Swing-Left(Right)-Leg**, **Lift-Left(Right)-Arm**, **Grasp**, **Spread-Fingers**, **Point-(Which)-Finger**, …

## Appendix B: Representation of Shape Change Concepts

CD+ representations can be used to represent general activities, and shape changes are basically activities that take place on the points within the shape. The following therefore illustrates the idea with examples of shape changes such as EXPANSION, CONTRACTION, EXTENSION, EXTRUSION, and INDENTATION.

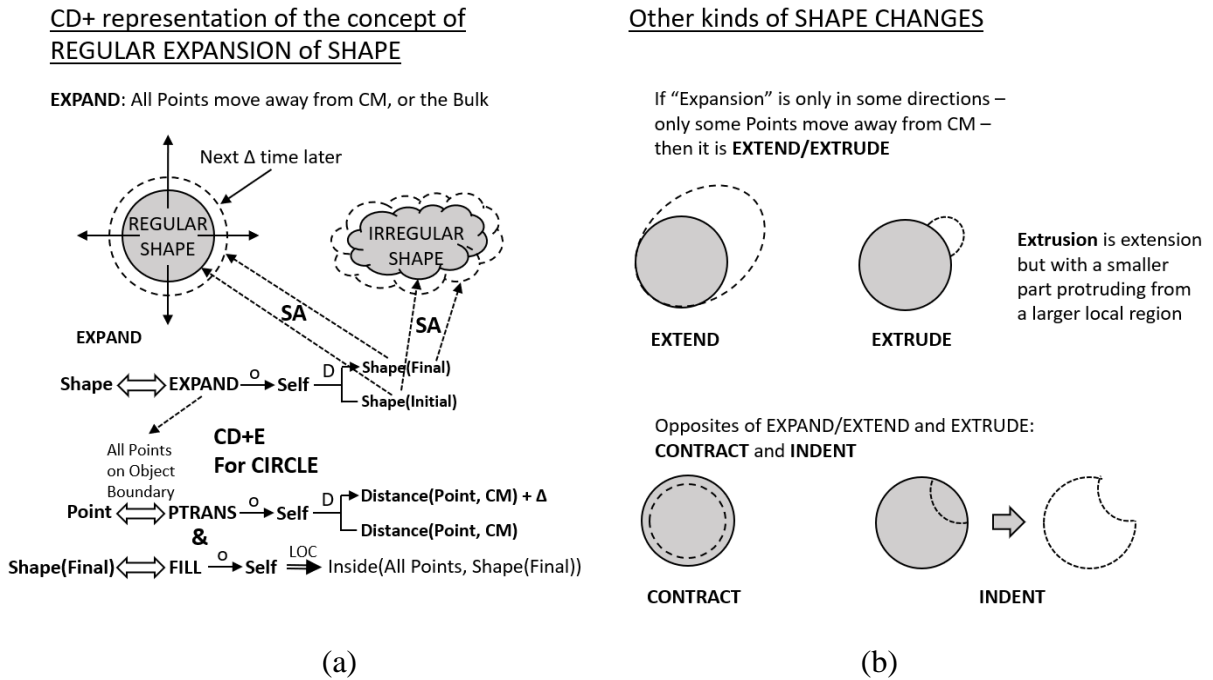

(a)                                                          (b)

**Fig. B.1.** (a) CD+ representation of regular expansion of regular *and* irregular shapes. (b) Shape changes other than regular expansion.

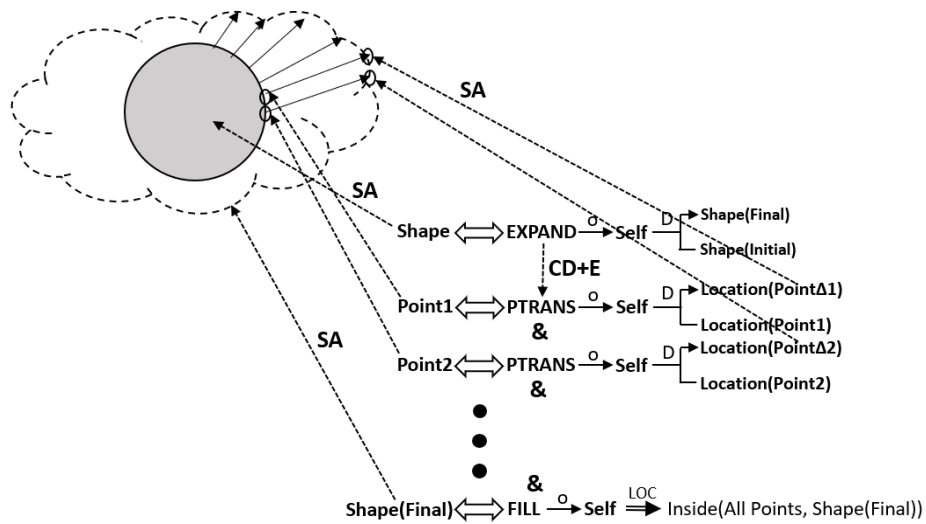

**Fig. B.2.** CD+ representation of irregular expansion of shape.